\documentclass[10pt,journal,compsoc]{IEEEtran}
\usepackage{amsmath, graphicx}
\usepackage{amsfonts, amsthm, amssymb, mathrsfs, mathtools}
\usepackage{bm}
\usepackage{setspace}
\usepackage{cite}
\usepackage{subfig}
\usepackage{array}
\usepackage{booktabs}
\usepackage{multirow}
\usepackage{cases}
\usepackage{makecell}
\usepackage{bbding}
\usepackage{footmisc}
\usepackage{rotating}
\usepackage{caption}
\usepackage{ulem}
\usepackage{diagbox}

\usepackage[english]{babel}
\usepackage{dblfloatfix}
\usepackage{url}
\usepackage{hyperref}
\hypersetup{colorlinks,linkcolor=magenta,anchorcolor=magenta,citecolor=magenta}
\usepackage{algorithm}
\usepackage{algpseudocode}
\usepackage{algorithmicx}

\newcolumntype{M}[1]{>{\centering\arraybackslash}m{#1}}

\usepackage{xcolor}%
\usepackage{textcomp}%
\usepackage{manyfoot}%
\usepackage{listings}%
\usepackage{bbm}
\usepackage{pifont}
\usepackage[accsupp]{axessibility}

\newcommand{\dcircle}[1]{\ding{\numexpr181 + #1}}
\newcommand{\cmark}{\ding{51}}%
\newcommand{\xmark}{\ding{55}}%
\newcommand{\ie}{\textit{i.e.}}
\newcommand{\eg}{\textit{e.g.}}
\newcommand{\hazel}[1]{\textcolor{red}{#1}}

\newcommand{\grey}[1]{\textcolor[RGB]{160,160,160}{#1}}
\definecolor{pblue}{RGB}{76, 175, 234}

\begin{document}

% \title{Refreshing Similarity-based Feature Upsampling}

% \title{A {Re}freshed {S}imilarity-based Framework for Direct High-Ratio {F}eature {U}psampling}

\title{A Refreshed Similarity-based Upsampler for Direct High-Ratio Feature Upsampling}

\author{Minghao Zhou, Hong Wang, Yefeng Zheng, \IEEEmembership{Fellow, IEEE}, and Deyu Meng, \IEEEmembership{Member, IEEE}
\IEEEcompsocitemizethanks{\IEEEcompsocthanksitem
M. Zhou and D. Meng are with School of Mathematics and Statistics and Ministry of Education Key Lab of Intelligent Networks and Network Security, Xi’an Jiaotong University, Xi’an 710049, China. Work was done when M. Zhou interned in Tencent YouTu Lab.\protect\\
(E-mail: woshizhouminghao@stu.xjtu.edu.cn, dymeng@mail.xjtu.edu.cn)

\IEEEcompsocthanksitem H. Wang and Y. Zheng are with Tencent YouTu Lab, Shenzhen, China. Y. Zheng is also with the Medical Artificial Intelligence Lab, Westlake University, Hangzhou, China (E-mail: hongwang9209@hotmail.com,  zhengyefeng@westlake.edu.cn)}

}% <-this % stops an unwanted space

% \markboth{Journal of \LaTeX\ Class Files,~Vol.~14, No.~8, August~2023}%
% {Shell \MakeLowercase{\textit{et al.}}: Bare Demo of IEEEtran.cls for Computer Society Journals}

\IEEEtitleabstractindextext{%
\begin{abstract}
Feature upsampling is a fundamental and indispensable ingredient of almost all current network structures for {dense prediciton} tasks. Very recently, a popular similarity-based feature upsampling pipeline has been proposed, which utilizes a high-resolution (HR) feature as guidance to help upsample the low-resolution (LR) deep feature based on their local similarity. Albeit achieving promising performance, this pipeline has specific limitations in methodological designs: 1) HR query and LR key features are not well aligned in a controllable manner; 2) the similarity between query-key features is computed based on the fixed inner product form, lacking flexibility; 3) neighbor selection is coarsely operated on LR features, resulting in mosaic artifacts. 
These shortcomings make the existing methods along this pipeline primarily applicable to hierarchical network architectures with iterative features as guidance and they are not readily extended to a broader range of structures, especially for a direct high-ratio upsampling. 
Against these issues, we {thoroughly} refresh this pipeline and meticulously optimize every methodological design. Specifically, we firstly propose an explicitly controllable query-key feature alignment from both semantic-aware and detail-aware perspectives, and then construct a parameterized paired central difference convolution block for flexibly calculating the similarity between the well-aligned query-key features. Besides, we develop a fine-grained neighbor selection strategy on HR features, which is simple yet effective for alleviating mosaic artifacts. Based on these careful designs, we systematically construct a refreshed similarity-based feature upsampling framework named \textbf{ReSFU}. Comprehensive experiments substantiate that our proposed ReSFU is finely applicable to various types of architectures in a direct high-ratio upsampling manner, and consistently achieves satisfactory performance on different applications, including semantic segmentation, medical image segmentation, instance segmentation, panoptic segmentation, {object detection, and monocular depth estimation}, showing superior generality and ease of deployment beyond the existing upsamplers. {Codes are available at \url{https://github.com/zmhhmz/ReSFU}. }
\end{abstract}

}

\maketitle

\IEEEdisplaynontitleabstractindextext

\IEEEpeerreviewmaketitle

\section{Introduction}
\label{sec:intro}
\begin{figure}[t]
  \centering
\includegraphics[width=0.99\linewidth]{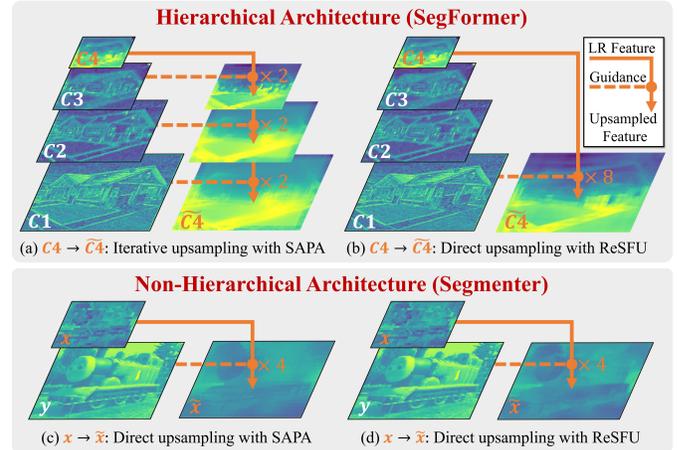}
  \vspace{-0.2cm}
\caption{Upper: For the $\times 8$ upsampling from $C4$ to $\widetilde{C4}$ in the hierarchical SegFormer~\cite{xie2021segformer}, (a) SAPA \cite{lu2022sapa} iteratively performs three $\times 2$ upsampling processes with intermediate features as guidance; (b) our ReSFU obtains clearer object boundaries in a direct one-step $\times 8$ upsampling. Lower: For the $\times 4$ upsampling from $\bm{x}$ to $\tilde{\bm{x}}$ {under the guidance of the shallow HR feature $\bm{y}$ in the non-hierarchical Segmenter \cite{strudel2021segmenter}, SAPA with (c) the default iterative upsampling incurs mosaic artifacts, which is worse in the (d) direct upsampling, while (e) ReSFU can largely eliminate the artifacts with one-step direct upsampling. }
\textit{All the images are best viewed by zooming in on screen, especially to observe mosaic artifacts.}} 
  \label{fig:first}
  \vspace{-0.3cm}
\end{figure}

As a fundamental ingredient in {deep network architectures}, feature upsampling aims to restore the spatial resolution of low-resolution (LR) features, {widely used in dense prediction} tasks, such as {semantic/instance/panoptic segmentation \cite{long2015fully,hafiz2020survey,kirillov2019panoptic}, object detection \cite{ren2016faster}, and depth estimation \cite{li2023depthformer}.}
% semantic segmentation~\cite{long2015fully}, instance segmentation~\cite{hafiz2020survey}, and panoptic segmentation~\cite{kirillov2019panoptic}. 

% During the feature upsampling process, each high-resolution (HR) feature element is typically estimated by weighting its neighboring LR feature elements.
During the feature upsampling process, each high-resolution (HR) feature element is typically estimated by weighting its neighboring elements {from the input LR feature}.
To generate the weights (also known as the upsampling kernels \cite{wang2019carafe,lu2022sapa}), 
in recent years, various upsampling research lines have been proposed. Specifically, bilinear and nearest-neighbor interpolation are the most widely-adopted feature upsampling methods, which are implemented based on hand-crafted weighting rules and often cause blurry effects. To enhance the flexibility of feature upsampling, some studies have developed different content-aware dynamic upsamplers~\cite{shi2016real,gao2019pixel,wojna2019devil,tian2019decoders,wang2019carafe}. For example, in \cite{wang2019carafe}, the authors devised a dynamic network to generate the upsampling kernel from LR features. Despite promising performance, they usually struggle to restore clear object boundaries due to the loss of fine-grained details in LR features.

{
Against this issue, a novel similarity-based feature upsampling pipeline SAPA~\cite{lu2022sapa,liu2023point} has been proposed, which utilizes an HR feature $\bm{y}$ as guidance to help accomplish the upsampling from the LR feature $\bm{x}$ to the HR one $\tilde{\bm{x}}$, which can be generally formulated as (see Sec.~\ref{sec:revisit} for more details):
\begin{equation} \label{eqn:intro}
    \tilde{\bm{x}}_i = \underbracket[0.4pt][2pt]{\text{Softmax}\left(\text{sim}(\bm{q}_i,\bm{k}^T_{\mathcal{N}(i)})\right)}_{\text{upsampling kernel}}\bm{x}_{\mathcal{N}(i)},
\end{equation}
where the HR query $\bm{q}$ and the LR key $\bm{k}$ are obtained from $\bm{y}$ and $\bm{x}$, respectively; $\mathcal{N}(i)$ is the neighborhood of the pixel $i$; $\text{sim}(\cdot,\cdot)$ denotes the similarity measure. From Eq. (\ref{eqn:intro}), we can easily see that the similarity-based feature upsampling framework mainly consists of three design components: \hypertarget{R1}{\dcircle{1}} the acquisition of query-key features $\bm{q}$ and $\bm{k}$; \hypertarget{R2}{\dcircle{2}} the similarity measure $\text{sim}(\cdot,\cdot)$ between the query-key pair; \hypertarget{R3}{\dcircle{3}} the neighbor selection $\mathcal{N}(i)$. Clearly, for the pipeline, the core essence of generating the high-quality $\tilde{\bm{x}}$ is to systematically optimize these three parts.}

% Albeit obtaining impressive success, we carefully delve into this pipeline and find that each of the aforementioned components in {the existing method SAPA \cite{lu2022sapa}} still suffers from unresolved issues. 

% {As a representative instantiation of the similarity-based feature upsampling pipeline Eq. (\ref{eqn:intro}), albeit obtaining impressive success, SAPA\footnote{{Without loss of generality, we discuss the base version of SAPA.}}  \cite{lu2022sapa} still encounters unresolved issues from all three aspects. }
{By carefully delving into the representative base version of SAPA, we find that each of the aforementioned components in SAPA suffers from unresolved issues.
Specifically, \dcircle{1} SAPA directly uses the linear projection of the HR guidance feature $\bm{y}$ and the LR deep feature $\bm{x}$ to obtain the query $\bm{q}$ and the key $\bm{k}$, respectively. Considering that $\bm{y}$ is usually shallower than $\bm{x}$}, only with a content-agnostic linear projection operation, the generated $\bm{q}$ and $\bm{k}$ are generally not well semantically aligned in a controllable manner. 
This would adversely interfere with the subsequent similarity calculation; 
\dcircle{2} {For $\text{sim}(\cdot,\cdot)$, SAPA simply uses the inner product form, which lacks flexibility and often results in blurry effects in the upsampled features (see $\widetilde{C4}$ in Fig. \ref{fig:first} (a))}; 
\dcircle{3} {For $\mathcal{N}(i)$, SAPA coarsely selects the neighbors} on the LR features $\bm{k}$ and $\bm{x}$, which would incur mosaic artifacts, especially in a direct high-ratio upsampling (see $\tilde{\bm{x}}$ in {Figs.~\ref{fig:first} (c)(d)}). 

These three methodological design limitations weaken the potential application of SAPA, making it usually only applicable to iterative $\times 2$ upsampling structures with hierarchical guidances by default but less suitable for direct high-ratio upsampling. For instance, for the hierarchical architecture SegFormer \cite{xie2021segformer} with four levels of features as shown in Fig. \ref{fig:first} (a), to accomplish the $\times 8$ upsampling from $C4$ to $\widetilde{C4}$, SAPA, as well as other existing feature upsampling methods \cite{wang2019carafe,lu2022fade,liu2023learning}, performs three consecutive $\times 2$ upsampling processes with $C3, C2$, and $C1$ as guidance features, respectively. Such an iterative pattern definitely leads to the laborious deployment of more upsampling modules. {If incorporating them in a direct high-ratio upsampling manner like Fig. \ref{fig:first} (b), the performance generally drops (see Sec.~\ref{sec:discuss}).}
{Besides, we pioneer the application of existing upsampling methods on non-hierarchical structures, such as Segmenter~\cite{strudel2021segmenter}. As shown in Figs. \ref{fig:first} (c) and (d), for the $\times$4 high-ratio upsampling,  whether inserting SAPA in the iterative upsampling manner by default or in the direct high-ratio upsampling manner causes different degrees of mosaic artifacts.} These issues generally exist in most of the current feature upsampling methods (see Sec.~\ref{sec:sem_seg}).
% Besides, for the non-hierarchical structures, such as Segmenter \cite{strudel2021segmenter} in Fig.~\ref{fig:first} (c), the deep LR feature $\bm{x}$ requires a direct $\times 4$ upsampling under the guidance of a shallow HR feature $\bm{y}$. Without {intermediate} guidance features, the upsampled feature $\tilde{\bm{x}}$ exhibits adverse visual artifacts. These issues generally exist in most of the current feature upsampling methods (see Sec.~\ref{sec:sem_seg}). 
%Faced with this tricky scenario, is it possible to build an \textit{architecture-agnostic} feature upsampling framework that is applicable to different types of network architectures and can always achieve satisfactory effects in \textit{direct high-ratio upsampling} without complicated iterative processes? 

{Faced with this tricky situation, a crucial question inevitably arises: is it possible to build an \textit{architecture-agnostic} feature upsampling framework that is applicable to different types of network architectures and can always achieve satisfactory effects in \textit{direct high-ratio upsampling} without complicated iterative processes? 
}

To answer this question, in this paper, we meticulously refresh the pipeline in Eq. \eqref{eqn:intro} and carefully optimize the involved three main components one by one. 
{
\dcircle{1} Firstly, we propose an explicitly controllable algorithm for the effective query-key alignment of $\bm{q}$ and $\bm{k}$. 
% to finely align query-key features from both semantic-aware and detail-aware perspectives.
Concretely, we utilize the guided filter (GF)~\cite{he2012guided} and Gaussian smoothing to accomplish semantic-aware mutual-alignment and detail-aware self-alignment, respectively. 
% Concretely, we utilize guided filter (GF)~\cite{he2012guided} to explicitly optimize the original linearly-projected HR query $\bm{q}$ and execute a controllable transformation on it to promote semantic alignment with the key. Meanwhile, we further investigate the HR query $\bm{q}$ itself and adopt an explicit Gaussian smoothing to help accomplish the HR feature self-alignment for better detail preservation. 
Such careful query-key alignment designs facilitate more accurate similarity computation and make it possible to directly execute the one-step high-ratio upsampling without complicated iterative procedures as in SAPA (see Figs.~\ref{fig:first} (b)(e)).
\dcircle{2} Secondly, for $\text{sim}(\cdot,\cdot)$, we propose a paired central difference convolution (PCDC) to inherently capture the local relevance between the query-key features, and then correspondingly construct a parameterized PCDC-Block to flexibly calculate the similarity between every well-aligned pair. Compared with the non-parametric inner product, the proposed PCDC-Block can help better ameliorate blurry artifacts, as experimentally validated by comparing ${\widetilde{C4}}$ in Figs. \ref{fig:first} (a) and (b).
\dcircle{3} Thirdly, we propose a fine-grained neighbor selection strategy by selecting the neighbors $\mathcal{N}(i)$ of every HR pixel $i$ based on the bilinearly-upsampled HR version of the LR features $\bm{k}$ and $\bm{x}$. Such a simple yet effective design introduces no extra parameters, which evidently reduces mosaic artifacts even in case of the direct high-ratio upsampling, as presented in {Fig.~\ref{fig:first} (e)}. Based on the proposed formulation in Eq.~\eqref{eqn:intro} and these three aforementioned methodological designs, we systematically construct a \textbf{Re}freshed \textbf{S}imilarity-based \textbf{F}eature \textbf{U}psampling framework, called ReSFU. In summary, our main contributions are three-fold: 
}

% {
% Collectively, we systematically construct a \textbf{Re}freshed \textbf{S}imilarity-based \textbf{F}eature \textbf{U}psampling framework called ReSFU. Our contributions are mainly three-fold:
% }

\vspace{1mm}
{
1) To the best of our knowledge, we are the first to provide the mathematical formulation of the similarity-based feature upsampling framework and reveal the underlying optimization essence (see Eq.~\eqref{eqn:intro}). Motivated by the essential understanding, we comprehensively optimize the three components. Correspondingly, we propose the explicitly controllable query-key alignment strategy to acquire the query-key features, PCDC-Block for similarity calculation $\text{sim}\left(\cdot,\cdot \right)$, and fine-grained neighbor selection $\mathcal{N}(i)$. These three essential methodological designs harmoniously form an inseparable whole as our proposed ReSFU.
% This should be the first attempt to carefully design a new similarity-based upsampling scheme based on the mathematically essential understanding of feature upsampling.
}

\vspace{1mm}
{
2) Attributed to our meticulous methodological designs, the proposed ReSFU is equipped with multiple application advantages.
% It can be inserted multiple steps iteratively like Figs.~\ref{fig:first} (a)(c), or inserted one step directly like Figs.~\ref{fig:first} (b)(e).
% Even using the more difficult one-step high-ratio upsampling procedure, our proposed ReSFU can still obtain superior performance while producing visually satisfactory effects, which facilitates the deployment (see Secs.~\ref{sec:sem_seg} and \ref{sec:discuss}).
Not only can it be inserted multiple steps iteratively like Figs.~\ref{fig:first} (a)(c) adopted by previous upsampling methods, but more meaningfully, even using the more difficult one-step high-ratio upsampling manner like Figs.~\ref{fig:first} (b)(e), our proposed ReSFU can still obtain superior performance while producing visually satisfactory effects (see Secs.~\ref{sec:sem_seg} and \ref{sec:discuss}).
% In addition to the iterative upsampling manner as in Figs.~\ref{fig:first} (a)(c), which is commonly adopted by previous upsampling methods, ReSFU can revolutionarily obtain superior performance while producing satisfactory visual results with the more difficult one-step high-ratio upsampling procedure as in Figs.~\ref{fig:first} (b)(e) (see Secs.~\ref{sec:sem_seg} and \ref{sec:discuss}).
Moreover, with the default configuration of direct high-ratio upsampling, our ReSFU can be easily and widely deployed to various network types, including hierarchical and non-hierarchical architectures, without requiring iterative intermediate features as guidance, which shows excellent generality and universality (see Fig.~\ref{fig:heads}).}
%in such a direct high-ratio upsampling manner which is the default configuration

 % By correspondingly replacing the three main components in Eq.~(\ref{eqn:intro}) with our proposed aforementioned three designs, we systematically build ReSFU.

 \vspace{1mm}
{3) Extensive experiments comprehensively demonstrate that compared to baselines, the proposed ReSFU consistently (i) achieves better results in a direct high-ratio upsampling on various types of architectures, and (ii) shows superior generality across different tasks, such as, semantic/instance/panoptic segmentation, object detection, and monocular depth estimation, and across various types of datasets, including natural and medical images. Moreover, we provide detailed model visualizations and ablation studies to verify the working mechanism of every component in the proposed ReSFU.}

The rest of the paper is organized as follows: Sec.~\ref{sec:related} introduces the related work. Sec.~\ref{sec:revisit} reformulates and analyzes the current similarity-based feature upsampling framework based on experimental visualization. Sec.~\ref{sec:method} presents the specific designs of our proposed ReSFU. Sec.~\ref{sec:exp} substantiates the effectiveness of our method through comprehensive experiments. Finally, Sec.~\ref{sec:con} concludes the paper.
  %4) Finally, by correspondingly replacing the three main components as analyzed in Eq.~(\ref{eqn:intro}) with our proposed specific designs, we can systematically build ReSFU, which is applicable to various types of network architectures with different guidance manners, including direct high-ratio upsampling, hierarchical upsampling, and progressive upsampling (see Fig.~\ref{fig:head}). Extensive experiments demonstrate the superiority of the proposed ReSFU as well as its fine generality on different segmentation tasks. Moreover, we provide detailed model visualizations to clearly verify the working mechanism underlying ReSFU.

\section{{Related Work on Feature Upsampling}}\label{sec:related}

% \noindent \textbf{Dense Prediction Tasks.}

% \paragraph{Feature Upsampling.} Feature upsampling is an indispensable component in modern deep-learning architectures, particularly for segmentation tasks \cite{long2015fully,he2017mask,kirillov2019panoptic}, which require accurate delineation of object boundaries. The two commonly used feature upsampling methods are NN and bilinear interpolation. However, these techniques lack awareness of the content and can result in over-smoothing of the features. Deconvolution \cite{noh2015learning} and PixelShuffle \cite{shi2016real} are alternative techniques for increasing the resolution of features, which may involve changing the contents of the original features. These methods, along with their subsequent improvements \cite{gao2019pixel,wojna2019devil,tian2019decoders}, provide certain flexibility but still lack the ability to dynamically adjust the upsampling parameters based on the feature contents.

% Feature upsampling is an indispensable component in modern deep-learning architectures, particularly for segmentation tasks \cite{long2015fully,he2017mask,kirillov2019panoptic} which require accurate delineation of object boundaries. 
% These traditional techniques lack awareness of feature contents and may lead to over-smoothing. 
% \noindent \textbf{Feature Upsampling.} 
% \subsection{Feature Upsampling}
 % \noindent \textbf{\hippo{Content-agnostic upsampling methods.}} 
 
\noindent\textbf{Guidance-Free Upsampling Methods.}
Nearest neighbor and bilinear interpolation are two widely adopted feature upsampling methods that are implemented based on pre-defined distance-aware rules. 
% For higher upsampling flexibility,  Deconvolution \cite{noh2015learning} and PixelShuffle \cite{shi2016real} 
For higher flexibility, several techniques have been proposed to learn the upsampling kernels in an end-to-end manner, such as Deconvolution~\cite{noh2015learning}. 
{PixelShuffle \cite{shi2016real} is also commonly adopted to perform efficient sub-pixel convolution for upsampling, which rearranges the elements within the channel dimension to increase the spatial resolution.} 
% are used as alternative techniques to upsample LR deep features. 
%Nevertheless involve changing the original feature contents. %Compared with traditional hand-crafted upsampling strategies, 
Although these approaches and their subsequent improvements \cite{gao2019pixel,wojna2019devil,tian2019decoders} achieve certain performance gains, they cannot dynamically adjust the upsampling kernels based on feature contents. 
% However, these approaches and their subsequent improved versions \cite{gao2019pixel,wojna2019devil,tian2019decoders}

% CARAFE \cite{wang2019carafe,wang2021carafe++} proposes to dynamically learn the upsampling kernels from the decoder features. Inspired by this spirit, subsequent methods such as IndexNet \cite{lu2019indices}, A2U \cite{dai2021learning}, and FADE \cite{lu2022fade} are developed to generate the dynamic kernels from one or both of the encoder and decoder features. Dysample \cite{liu2023learning} takes an alternative perspective by generating the point sampling offsets of the upsampled pixels from the decoder features. To leverage the information contained in the shallow features to guide the upsampling process, Max unpooling \cite{badrinarayanan2017segnet} utilizes the stored positional indices from the previous max-pooling downsampling process. GUN \cite{mazzini2018guided} proposes to integrate information from features of a higher-resolution image into the final upsampling operation. SAPA \cite{lu2022sapa,liu2023point} obtains the upsampling kernels by calculating the local similarity between the shallow and deep features. Our work inherits and improves this research line by meticulously designing a module to enhance the upsampling results under the guidance of shallow features.

% \noindent \textbf{\hippo{Dynamic upsampling methods without guidance.}} 
Recently, CARAFE \cite{wang2019carafe,wang2021carafe++} firstly proposed to dynamically learn the upsampling kernels from the to-be-upsampled LR features. Alternatively, DySample \cite{liu2023learning} generated the point sampling offsets of the upsampled pixels from the deep LR features. In these methods, the upsampling procedure can be dynamically adjusted based on deep LR features, further improving the upsampling flexibility. However, due to the loss of fine-grained details in deep LR features and the lack of HR guidance features, the upsampled features achieved by these guidance-free methods usually have blurry effects. 

%\noindent \textbf{\hippo{Dynamic upsampling methods with guidance.}}
\vspace{1mm} 
\noindent\textbf{Guidance-based Upsampling Methods.}
% Inspired by this research line, several models have been developed to generate the dynamic kernels from encoder and decoder features, such as IndexNet \cite{lu2019indices}, A2U \cite{dai2021learning}, and FADE \cite{lu2022fade}.
% To better recover details in the upsampled features, a research line leverages the information in shallow features to guide the upsampling process 
To enhance the detail recovery in upsampled features, a research line has been proposed to utilize the information from shallow HR features to guide the upsampling process \cite{badrinarayanan2017segnet,mazzini2018guided}. For example, IndexNet \cite{lu2019indices}, A2U \cite{dai2021learning}, and FADE \cite{lu2022fade} were developed in succession to generate the dynamic upsampling kernels from both the HR guidance and LR deep features. 
% \hippo{These methods depend on the HR guidance features that provide high-resolution details that would benefit the upsampling procedure \cite{badrinarayanan2017segnet,mazzini2018guided}.} 
Along this line, SAPA \cite{lu2022sapa,liu2023point} proposed to generate the upsampling kernels by modelling the local similarity between HR guidance and LR deep features. FeatUp \cite{fu2024featup} introduced a multi-view consistency learning framework based on the stacked parameterized joint bilateral upsampler (JBU) \cite{kopf2007joint} and an implicit multilayer perceptron (MLP) upsampler, respectively. The implicit version requires inference-time training, which is extremely time-consuming. In this paper, driven by the competitive performance of the latest representative SAPA, we follow its similarity-based feature upsampling framework, thoroughly analyze the inherent characteristics, and then specifically propose optimization designs for better upsampling effects.

{Please refer to the supplementary material (SM) for more extensive related works, such as the convolution operators and image segmentation methods. }

\section{Revisiting Similarity-based Feature Upsampling Pipeline} \label{sec:revisit}

\begin{figure*}[t]
  \centering
  % \vspace{-0.2cm}
\includegraphics[width=0.99\linewidth]{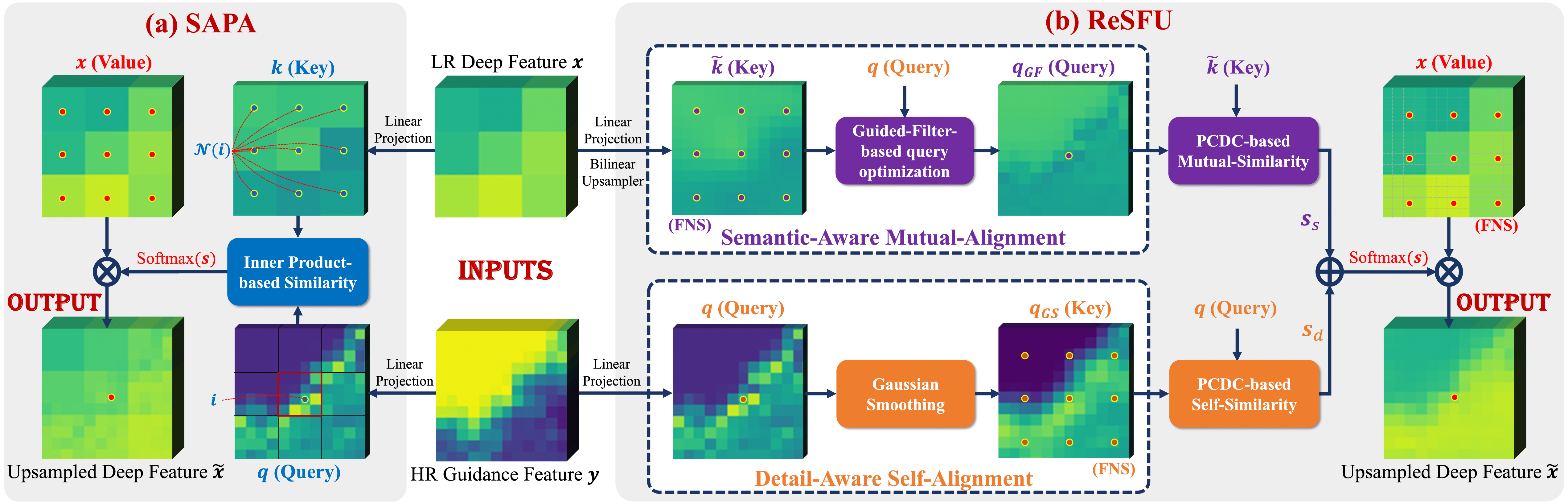}
  \vspace{-0.1cm}
% \caption{Illustration of the workflows of (a) SAPA-B and (b) our proposed ReSFU, mainly consisting of three key parts, \emph{i.e.}, query-key feature alignment, PCDC-based similarity calculation, and smooth neighbor selection (SNS). Here each visualization component is experimentally obtained based on Segmenter-S~\cite{strudel2021segmenter} for the $\times 4$ upsampling of the LR deep feature $\bm{x}$ under the guidance of the HR feature $\bm{y}$.}
\caption{Targeting every part in (a) SAPA, (b) our ReSFU proposes specific optimization designs, \ie, controllable query-key feature alignment from both semantic-aware and detail-aware perspectives, paired central difference convolution (PCDC)-based flexible similarity calculation between aligned query-key pairs, and fine-grained neighbor selection (FNS). Here the visualization is experimentally based on Segmenter-S for a $\times 4$ feature upsampling from $\bm{x}$ to $\tilde{\bm{x}}$. \textit{Best viewed with zoom-in.}}
 % Here each visualization component is experimentally obtained based on Segmenter-S~\cite{strudel2021segmenter} for the $\times 4$ upsampling of the LR deep feature $\bm{x}$ under the guidance of the HR feature $\bm{y}$.
% \caption{The pipeline of (a) SAPA-B mainly consists of three key parts, \emph{i.e.}, quey-key feature design, similarity calculation, and neighbor selection. Targeting every part, (b) our proposed ReSFU proposes specific optimization, \emph{i.e.}, query-key feature alignment from both semantic-aware and detail-aware perspectives, PCDC-based similarity calculation, and smooth neighbor selection (SNS). Here each visualization component is experimentally obtained based on Segmenter-S~\cite{strudel2021segmenter} for the $\times 4$ upsampling of the LR deep feature $\bm{x}$ under the guidance of the HR feature $\bm{y}$.}
  \label{fig:main}
  \vspace{-0.2cm}
\end{figure*}

In this section, we carefully reformulate and delve into the current similarity-based feature upsampling pipeline.

Given an LR deep feature $\bm{x}\in \mathbb{R}^{hw \times C}$ and an HR guidance feature $\bm{y}\in \mathbb{R}^{HW \times c}$,\footnote{For simplicity, we aggregate the spatial dimensions $H \times W$ as $HW$.} {the similarity-based feature upsampling framework SAPA~\cite{lu2022sapa,liu2023point} }aims to upsample $\bm{x}$ to an HR deep feature $\tilde{\bm{x}}\in \mathbb{R}^{HW \times C}$ under the guidance of $\bm{y}$. Specifically, for each HR pixel $i\in\{1,\ldots, HW\}$, the HR deep feature element $\tilde{\bm{x}}_{i}\in \mathbb{R}^{C}$ is generally estimated by weighting its neighboring LR deep feature elements based on an upsampling kernel $\text{Softmax}(\bm{s}_{i})$ as:
\begin{equation} \label{eqn:attn}
\tilde{\bm{x}}_{i}=\text{Softmax}(\bm{s}_{i})\bm{x}_{\mathcal{N}(i)},
\end{equation}
where $\bm{x}_{\mathcal{N}(i)} \in \mathbb{R}^{K^2\times C}$ denotes the neighboring LR feature elements of pixel $i$; $|{\mathcal{N}(i)}| = K^2$ is the number of neighboring LR pixels with a pre-defined kernel size $K$; $\bm{s}_{i}\in \mathbb{R}^{K^2}$ is the similarity scores assigned to the $K^{2}$ neighbors of pixel $i$, {which is generally computed as:
% In the representative base version of SAPA \cite{lu2022sapa}, $\bm{s}_{i}$ is computed as:
\begin{equation}\label{eqn:sapa}
\bm{s}_{i}=\text{sim}(\bm{q}_{i},\bm{k}_{\mathcal{N}(i)}),
\end{equation}}
%\triangleq  \bm{q}_{i} \bm{k}^{T}_{\mathcal{N}(i)}
where $\bm{q}$ and $\bm{k}$ are generated from $\bm{y}$ and $\bm{x}$, respectively, with $D$ channels; $\bm{q}_{i} \in \mathbb{R}^{D}$ is the $i$-th feature element of $\bm{q}\in \mathbb{R}^{HW \times D}$; $\bm{k}_{\mathcal{N}(i)} \in \mathbb{R}^{K^2 \times D}$ represents the $K^{2}$ neighboring feature elements of pixel $i$ in $\bm{k}\in \mathbb{R}^{hw \times D}$; $\text{sim}\left(\cdot,\cdot \right)$ is a general similarity function. We regard $\bm{q}$, $\bm{k}$, and $\bm{x}$ as the HR query, LR key, and LR value features, respectively.
% designed as the inner product between $\bm{q}_{i}$ and $\bm{k}_{\mathcal{N}(i)}$ in SAPA.

% The core of the upsampling procedure as indicated by \cref{eqn:attn} is the local mutual similarity between the deep and shallow features. Specifically, according to the inner-product similarity between the embeddings of the shallow and deep features, a position is more likely to be assigned to the semantic cluster where the obtained similarity is higher. 

%\sout{features $\bm{q}$ and $\bm{k}$ learning} 
As seen, the entire similarity-based feature upsampling pipeline is strongly associated with the design 
of three main parts: {\dcircle{1}} {query-key feature acquisition for facilitating the computation of similarity $\bm{s}_i$}; {\dcircle{2}} the similarity function $\text{sim}\left(\cdot,\cdot \right)$; {\dcircle{3}} neighbor selection $\mathcal{N}\left(i\right)$. 

{For the representative base version of SAPA~\cite{lu2022sapa}, the query $\bm{q}$ and the key $\bm{k}$ are obtained by taking linear projections of $\bm{y}$ and $\bm{x}$, denoted as
$\text{Proj}_{\bm{\theta}_{\bm{q}}}(\bm{y})$ and 
$\text{Proj}_{\bm{\theta}_{\bm{k}}}(\bm{x})$, respectively, where $\bm{\theta}_{\bm{q}}$ and $\bm{\theta}_{\bm{k}}$ are network parameters. The similarity measure is implemented via a simple inner product operation, \ie, $\text{sim}(\bm{q}_{i},\bm{k}_{\mathcal{N}(i)})=\bm{q}_{i} \bm{k}^{T}_{\mathcal{N}(i)}$. The neighbors $\mathcal{N}(i)$ of pixel $i$ are selected coarsely on the LR features.} For better understanding, based on the backbone Segmenter-S with a direct $\times 4$ upsampling for the last-layer feature under the guidance of a shallow feature (see {Fig.~\ref{fig:heads} (b)}),
% without hierarchical processes like Fig.~\ref{fig:first}(a)
we experimentally visualize SAPA in Fig.~\ref{fig:main} (a) and identify the following issues: 

\noindent 1) \textbf{Uncontrollable Query-Key Alignment.}
In such a direct high-ratio upsampling case, the HR guidance feature $\bm{y}$ is relatively shallow and it generally contains low-level information, such as texture patterns, while the LR deep feature $\bm{x}$ usually contains high-level semantics \cite{zeiler2014visualizing,dosovitskiy2021an}.
% Since the linear projection operation is content-agnostic, which means it is unaware of the content of either $\bm{y}$ or $\bm{x}$,
Since the linear projection operation with fixed weights is not sensitive to the content of either $\bm{y}$ or $\bm{x}$, 
{the linearly-projected query and key cannot be well-aligned in the feature space without extra control, which would seriously mislead the subsequent similarity computation.}

\noindent 2) \textbf{Inflexible Similarity Computation.} The conventional inner product operation is non-parametric and {lacks flexibility in capturing the relations between the query and key}. Directly adopting this inner product form would cause that within a small local region $\mathcal{N}(i)$, the computed $K^{2}$ scores $\bm{q}_{i} \bm{k}^{T}_{\mathcal{N}(i)}$ are close to each other (see Fig.~\ref{fig:sim} (a)). This would consequently generate an overly smooth upsampling kernel, leading to blurry artifacts in the upsampled $\tilde{\bm{x}}$.

% \noindent 3) \textbf{Non-smooth Neighbor Selection.} For the $\times 4$ upsampling example shown in \cref{fig:main} (a), all the feature elements in the $4\times4$ red box marked in $\bm{q}$ share identical neighboring feature elements marked by points in $\bm{k}$. On the other hand, when considering two adjacent feature elements in $\bm{q}$ from different $4\times4$ boxes, the selected neighbors for them in $\bm{k}$ would differ. Such a grid-wise neighbor selection strategy often results in mosaic artifacts in $\tilde{\bm{x}}$, especially for high-ratio upsampling.%没有重点指出LR

\noindent {3) \textbf{Coarse Neighbor Selection.} 
{From Eqs.~\eqref{eqn:attn} and \eqref{eqn:sapa}, the neighbors $\mathcal{N}(i)$ of pixel $i$ are correspondingly selected on the LR key $\bm{k}$ and the LR value $\bm{x}$, respectively. This manner is coarse. Specifically,} taking the neighbor selection on $\bm{k}$ as an example, as shown in Fig. \ref{fig:main} (a), for the $\times 4$ upsampling, all the sixteen pixels in the $4\times4$ red box marked in $\bm{q}$ share identical neighbors, \ie, the nine pixels marked in $\bm{k}$. On the other hand, for two adjacent pixels $i$ and $j$ in $\bm{q}$ from two different $4\times4$ boxes, the selected neighbors $\mathcal{N}(i)$ and $\mathcal{N}(j)$ in $\bm{k}$ would differ abruptly. Such a grid-wise neighbor selection strategy on LR features often results in mosaic artifacts in $\tilde{\bm{x}}$, especially for a direct high-ratio upsampling.}

{These limitations in methodological designs constrain that most of the existing feature upsampling methods along this similarity-based pipeline are always injected into network backbones in a step-by-step $\times$ 2 upsampling manner. They are generally suitable for hierarchical architectures with the iterative upsampling process, like SegFormer in Fig.~\ref{fig:first} (a), and are difficult to be extended to more types of backbones, such as non-hierarchical architectures with a direct high-ratio upsampling, like Segmenter in Fig.~\ref{fig:first} (c).}

\section{Method}\label{sec:method}
% {To alleviate the issues existing in the current similarity-based unsampling methods as analyzed in Eq.~\ref{sec:revisit}, in this section, we carefully optimize every part in Eq.~\eqref{eqn:attn} and specifically propose detailed designs, including guided filter based query learning for better feature matching, Paired Central Difference Convolution for similarity computation $\text{sim}\left(\cdot\right)$, and a smoothed neighbor selection mechanism $\mathcal{N}(i)$. Then we correspondingly construct a more flexible unsampling framework which obtains an unsampled deep feature $\tilde{\bm{x}}$ with higher fidelity and better semantic consistency as shown in Fig.~\ref{fig:main} (b). The details are given as below.}

% In this section, following the current similarity-based feature upsampling pipeline,  we carefully optimize the involved three key components in Eqs.~(\ref{eqn:attn})(\ref{eqn:sapa}) and

{Motivated by the analysis in Sec.~\ref{sec:revisit}, in this section, we aim to refresh the current similarity-based feature upsampling framework by carefully optimizing every involved component in Eqs. (\ref{eqn:attn}) and (\ref{eqn:sapa}), including query-key alignment, similarity computation, and neighbor selection. Then we correspondingly construct a more flexible and universal upsampling framework, called ReSFU, which has the potential to adapt to various network structures.}

% To alleviate the aforementioned issues as analyzed in Sec.~\ref{sec:revisit}, in this section, {focusing on} the pipeline of SAPA-B, we carefully optimize every part in Eqs. (\ref{eqn:attn})(\ref{eqn:sapa}) and construct a more flexible similarity-based feature upsampling framework, called ReSFU, as shown in Fig.~\ref{fig:main} (b).

Specifically, as shown in Fig.~\ref{fig:main} (b), we first propose a controllable query-key feature alignment method from both semantic-aware and detail-aware perspectives. Then, we design a specific paired central difference convolution (PCDC) block for flexibly calculating the similarity between the aligned query-key pairs. {Finally, we devise a fine-grained neighbor selection (FNS) strategy to alleviate mosaic artifacts.} % and design a smooth neighbor selection strategy. Finally, based on Eq.~\eqref{eqn:sapa}, we naturally obtain the upsampled feature $\tilde{\bm{x}}$ with richer details and fewer mosaic artifacts. 
For each part, the detailed designs are given below. 

% .~\footnote{From Sec.~\ref{sec:revisit}, query and key are derived from HR features and LR features, respectively. For simplicity, we also call query and key as features.} 

% To alleviate the issues in the current similarity-based feature upsampling methods as identified in \cref{sec:revisit}, in this section, we propose Enhanced Guided Feature Upsampling (EGFU) by redesigning the three key components of \cref{eqn:attn}, \ie, feature pair selection, similarity calculation, and neighbor selection. The framework of our proposed method is shown in \cref{fig:main} (b) and the detailed designs are given below. 

%\subsection{Selecting Comparable Feature Pairs}
\subsection{Explicitly Controllable {Query-Key} Alignment}\label{sec:align}
% \paragraph{Semantic perspective.} Relying solely on the detail-aware similarity calculated on $\bm{q}$ can lead to inaccuracies. Considering the presence of semantic-irrelevant noises in the shallow features, especially at object boundaries, it is possible for elements with substantial differences within a local region of the shallow features to have similar semantics, while the detail-aware similarity between them would be very low. Hence, the similarity scores measured at the detail level may not adequately capture the underlying similarity. 

\begin{figure}[t]
  \centering
  % \vspace{-0.2cm}
% \includegraphics[width=0.98\linewidth]{figs/detail_align.pdf}
\includegraphics[width=0.98\linewidth]{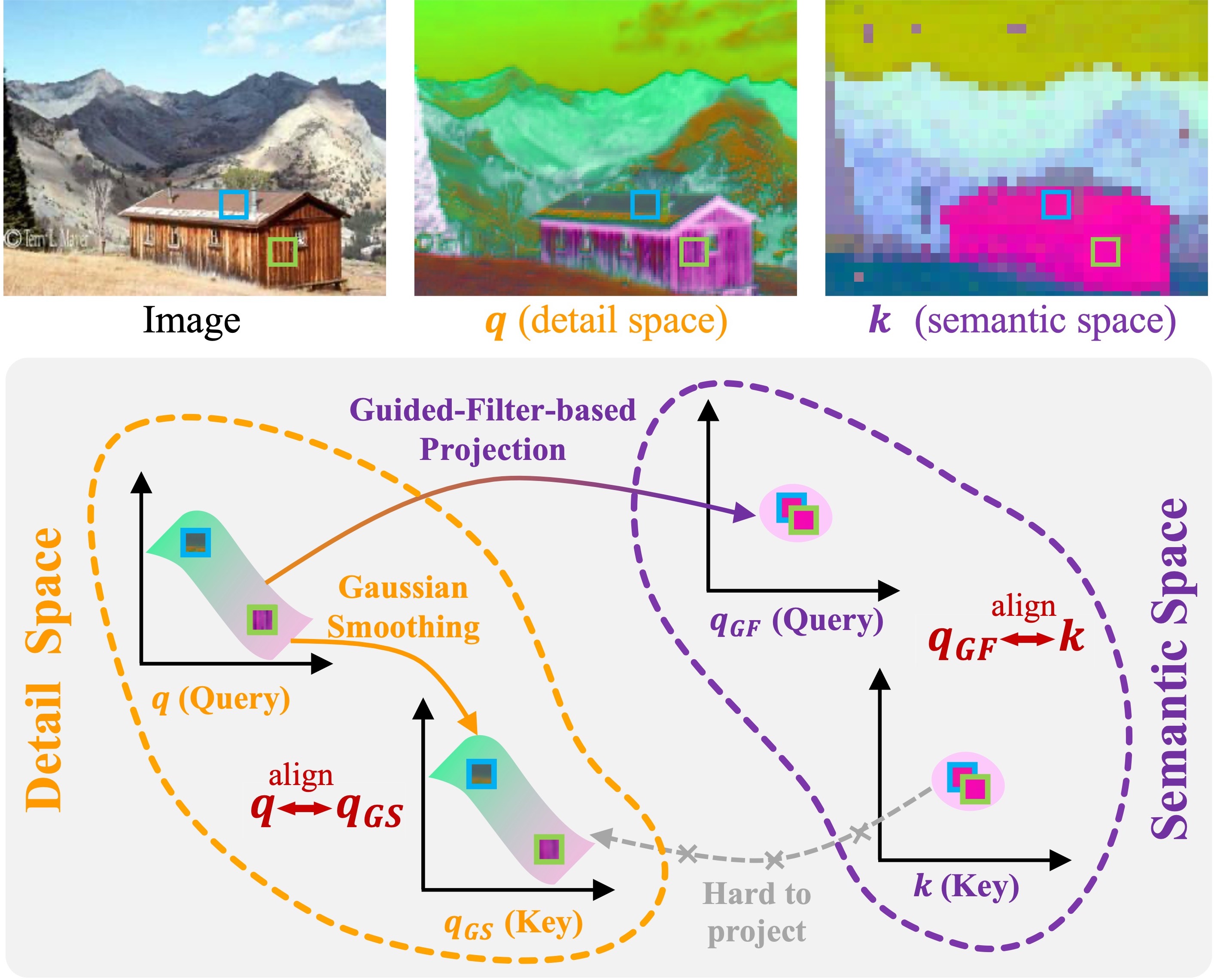}
 \vspace{-0.2cm}
  \caption{{Upper: For the two regions marked by blue and green boxes, \ie, the roof and the wall of the house, they have similar features in the semantic space but significantly differ in the detail space. Lower: The overall concept of our two-pronged approach to construct the aligned query-key pair in the semantic space and detail space, respectively.}} %Here the visualization of $\bm{q}$ and $\bm{k}$ is experimentally based on Segmenter-S.
  \label{fig:detail}
  \vspace{-0.2cm}
\end{figure}

% \hippo{As pointed out by \cite{zeiler2014visualizing,dosovitskiy2021an}, an HR guidance feature $\bm{y}$ from a shallower layer of the network usually captures more detail-level information while the LR deep feature $\bm{x}$ encodes more semantic-related information. We thus informally refer to $\bm{y}$ (and its linear projection, the query feature $\bm{q}$) and $\bm{x}$ (and its linear projection, the key feature $\bm{k}$) as residing in the \textit{detail space} and \textit{semantic space}, respectively. To facilitate the controllable alignment of the query-key pair, we adopt a two-pronged approach: 1) aligning them in the semantic space, and 2) aligning them in the detail space. The two aligned pairs are jointly utilized for the final similarity calculation.}
% {As in SAPA, we can obtain the query $\bm{q}$ and the key $\bm{k}$ by performing linear projections on $\bm{y}$ and $\bm{x}$, parameterized by $\bm{\theta}_{\bm{q}}$ and $\bm{\theta}_{\bm{k}}$, respectively, for the purpose of channel reduction.} 
From \cite{zeiler2014visualizing,dosovitskiy2021an}, the HR guidance feature $\bm{y}$ from a shallow layer usually captures more detail-related information while the LR deep feature $\bm{x}$ encodes more semantic-related information. Thus, we informally refer to $\bm{y}$ (and its linear projection, the query feature $\bm{q}$) 
and $\bm{x}$ (and its linear projection, the key feature $\bm{k}$) 
as residing in the \textit{detail space} and the \textit{semantic space}, respectively, as presented in Fig.~\ref{fig:detail}.

Confronted with the representational discrepancy between the detail and semantic spaces, it is inaccurate to directly calculate the similarity score between the original query-key feature pair, \ie, $\bm{q}$ and $\bm{k}$, across different spaces. To fully capture and exploit the relations among different neighboring pixels that share similar information both at
the detail and semantic levels for guiding the upsampling process, we propose a two-pronged approach to explicitly transform the original features and generate two query-key pairs aligned in detail space and semantic space, respectively, to facilitate accurate similarity computation. 
Please refer to the lower row of Fig.~\ref{fig:detail} for the overall design.

\subsubsection{Semantic-Aware Mutual-Alignment} \label{sec:mutual}
 % \hippo{Firstly, we aim to optimize the query $\bm{q}$ and transform it into the semantic space. Inspired by guided filter (GF) \cite{he2012guided,he2015fast}, we propose to linearly transform $\bm{q}$ in every local window while minimizing the MSE between the transformed result and the key $\bm{k}$ within the window. As a result, the detail-level structural information in $\bm{q}$ can be well-preserved thanks to the local linear transformation operation, and the transformed output can indeed approach the semantic space by minimizing its difference with $\bm{k}$.}

 % {Inspired by guided filter (GF) \cite{he2012guided,he2015fast} that can efficiently integrate the structural information of a guidance image with the semantic information of an input image, for the specific feature upsampling task, here we propose to utilize GF to explicitly transform the original query $\bm{q}$ to better align with the key feature in the semantic space.}

{Firstly, our goal is to project the original query $\bm{q}$ into the semantic space so that it can better align with the key feature $\bm{k}$ in the semantic space, while also preserving the structural details in the detail space. To this end, inspired by the guided filter (GF) \cite{he2012guided,he2015fast}, we propose to linearly transform $\bm{q}$ in every local window for detail preservation while minimizing the distance between the transformed query and the key $\bm{k}$ for the mutual-alignment in the semantic space. In this manner, the structural information of $\bm{q}$ can be efficiently integrated with the semantic information of $\bm{k}$ into the transformed query~\cite{he2012guided}.}

{Mathematically, let ${\bm{q}_{GF}}\in\mathbb{R}^{HW\times D}$ be the transformed query and each of its element $({\bm{q}_{GF}})_{id}$ can be estimated via solving the following optimization problem~\cite{he2012guided}:}
% Inspired by guided filter (GF) \cite{he2012guided,he2015fast}, we propose to linearly transform $\bm{q}$ in every local window while minimizing the distance between the transformed query and the key $\bm{k}$ within the window. Mathematically, let ${\bm{q}_{GF}}\in\mathbb{R}^{HW\times D}$ be the transformed query and its each element $({\bm{q}_{GF}})_{id}$ can be optimized via solving the following problem~\cite{he2012guided}:}
% {Specifically, we first regard $\bm{q}$ and {$\tilde{\bm{k}}$ $\in\mathbb{R}^{HW \times D}$ (bilinearly upsampled result of $\bm{k}$)} as the guidance image and the input image for GF, respectively. Let ${\bm{q}_{GF}}\in\mathbb{R}^{HW\times D}$ be the transformed query and its each element $({\bm{q}_{GF}})_{id}$ can be optimized via solving the following problem~\cite{he2012guided}:}
% \hippo{Mathematically, let ${\bm{q}_{GF}}\in\mathbb{R}^{HW\times D}$ be the transformed query. We can estimate each element of the filtered result $({\bm{q}_{GF}})_{id}$ via solving the following problem~\cite{he2012guided}:}
\begin{equation} \label{eqn:gf}
\begin{split}
\min_{\bm{m}_{jd},\bm{n}_{jd}} &\sum_{i\in I_j} \left(\Big(({\bm{q}_{GF}})_{id} - \tilde{\bm{k}}_{id}\Big)^2+\epsilon \bm{m}_{jd}^2\right), \\
s.t.\  &\ ({\bm{q}_{GF}})_{id} = \bm{m}_{jd}\bm{q}_{id} + \bm{n}_{jd}, \forall i\in I_j,
\end{split}
 \end{equation}
where $({\bm{q}_{GF}})_{id}$, ${\bm{q}}_{id}$, and $\tilde{\bm{k}}_{id}$ are the feature elements of $\bm{q}_{GF}$, $\bm{q}$, and $\tilde{\bm{k}}$ at pixel $i$ and channel $d$, respectively; $\tilde{\bm{k}}\in\mathbb{R}^{HW \times D}$ is the bilinearly upsampled HR result of the original LR key $\bm{k}$; $d\in\{1,\ldots,D\}$; 
%$\tilde{\bm{k}}\in\mathbb{R}^{HW \times D}$ is the bilinearly upsampled HR result of the original LR key $\bm{k}$; 
$\bm{m}_{jd}$ and $\bm{n}_{jd}$ are linear coefficients for pixel $j$ at channel $d$; $I_j$ is a square window with radius $r$ centered at pixel $j\in\{1,\ldots,HW\}$; and $\epsilon$ is a small regularization weight. {In experiments, the radius $r$ is experimentally set to 8 and $\epsilon$ is set to 0.001. Please note that for element-wise minimization computation in Eq.~\eqref{eqn:gf}, we adopt the upsampled key feature $\tilde{\bm{k}}$ for keeping the same size with $\bm{q}_{GF}$, which also belongs to the semantic space.}

% The local linear transformation constraint implies that the detail-related structural information in $\bm{q}$ can be well-preserved in the transformed query ${\bm{q}_{GF}}$. The minimization objective function would promote ${\bm{q}_{GF}}$ to approach $\bm{k}$ for accomplishing the mutual-alignment in the semantic space.
 %In experiments, $r=8$.
% To keep the same size with $\bm{q}_{GF}$ for computation, the original key $\bm{k}\in\mathbb{R}^{d\times h \times w}$ is bilinearly upsampled to $\tilde{\bm{k}}\in\mathbb{R}^{d\times H \times W}$;

From~\cite{draper1998applied}, the explicit solution of Eq. \eqref{eqn:gf} can be easily derived as:
 \begin{equation} \label{eqn:gf2}
     \begin{split}
    \bm{m}_{jd} &= \frac{\frac{1}{|{I_{j}}|} \sum_{i\in I_j} \bm{q}_{id} \tilde{\bm{k}}_{id} - \bm{\mu}^q_{jd}\bm{\mu}^k_{jd}}{\bm{\sigma}^2_{{jd}} + \epsilon}, \\
    \bm{n}_{jd} & = \bm{\mu}^k_{jd} - \bm{m}_{jd} \bm{\mu}^q_{jd},
     \end{split}
 \end{equation}
where $\bm{\mu}^q_{jd}$ and $\bm{\sigma}^2_{{jd}}$ are the mean and variance of $\bm{q}_{id}$ in the local window $I_j$; $\bm{\mu}^k_{jd}$ is the mean of $\tilde{\bm{k}}_{id}$ in $I_j$, expressed as $\bm{\mu}^k_{jd}=\frac{1}{|{I_{j}}|} \sum_{i\in I_j}\tilde{\bm{k}}_{id}$; and {$|I_{j}|$ is the number of pixels contained in the local window $I_{j}$}, which is $r^2$. 

Since a pixel $i$ is involved in different overlapping window $I_j$ that covers the pixel $i$, $\left({\bm{q}_{GF}}\right)_{id}$ would change with these local windows. Following~\cite{he2012guided}, by averaging all these overlapping local windows $I_{j}$, we can get the final transformed query as:
\begin{equation} \label{eqn:gf3}
    \begin{split}
        (\bm{q}_{GF})_{id} =\bar{\bm{m}}_{id} \bm{q}_{id} + \bar{\bm{n}}_{id},
    \end{split}
\end{equation}
where $\bar{\bm{m}}_{id} = \frac{1}{|I_{i}|}\sum_{j\in{I_i}} \bm{m}_{jd}$ and $\bar{\bm{n}}_{id} = \frac{1}{|I_{i}|}\sum_{j\in{I_i}} \bm{n}_{jd}$.

% As can be seen in \cref{fig:main} (b), the visualization of the GF result $\bm{q}_{GF}$ exhibits a striking resemblance to the deep features $\bm{k}$, while effectively preserving the intricate structures in the original shallow features $\bm{q}$. 

% \TODO{whether provide visualization of real imgs here, or put in experiments section and say ``please refer to Sec 5 for more visualization"} 

As seen, the optimized query $\bm{q}_{GF}$ is explicitly derived on the basis of the original query $\bm{q}$ and key $\tilde{\bm{k}}$, achieving the controllable alignment with key in the semantic space. From Eq.~\eqref{eqn:sapa}, for the explicitly optimized query-key pair $\bm{q}_{GF}$ and $\tilde{\bm{k}}$, the similarity ${(\bm{s}_s)}_i$ for pixel $i$ is calculated as:
% between the optimized query $\bm{q}_{GF}$ and the bilinearly-upsampled key $\tilde{\bm{k}}$ is computed as:
\begin{equation} \label{eqn:semantic} \vspace{-1mm}
{(\bm{s}_s)}_i \triangleq \text{sim}(({\bm{q}_{GF}})_{i}, {\tilde{\bm{k}}}_{\mathcal{N}(i)}).
\end{equation}
We call $\bm{s}_s\in \mathbb{R}^{HW \times K^2}$ as semantic-aware mutual similarity.
% \vspace{-1mm}

\vspace{1mm}
\noindent\textit{\textbf{Remark 1:}} From the visualization in Fig. \ref{fig:main} (b), we can clearly observe that: 1) compared to the original linearly-projected query $\bm{q}$, the explicitly optimized query $\bm{q}_{GF}$ exhibits a stronger semantic resemblance to the key $\tilde{\bm{k}}$. This can promote more accurate semantic-aware similarity computation of $\bm{s}_s$ for better performance (see Sec.~\ref{sec:ablation}); 2) $\bm{q}_{GF}$ effectively preserves the original intricate structures in $\bm{q}$, which would guide the upsampling to achieve a higher structural fidelity in the upsampled feature $\tilde{\bm{x}}$ (also see Fig.~\ref{fig:vis_components}). All these advantages finely comply with our design motivation for Eq.~\eqref{eqn:gf} and substantiate its rationality. {Please note that although guided filter is the existing technique, we are the first to propose the concept of semantic-aware mutual alignment based on the essential understanding of the pipeline as analyzed in Sec.~\ref{sec:revisit}. Driven by the novel concept, the guided filter is indeed the most natural solution for efficient feature alignment in semantic space.}
\subsubsection{Detail-Aware Self-Alignment}\label{sec:self}

% \hippo{To construct a controllably aligned query-key pair in the detail space, directly transforming the original key $\bm{k}$ to the detail space is infeasible. For example, in Fig. \ref{fig:detail}, the two marked regions exhibit similar semantic information while differing substantially in the detail space. Generally, there does not exist a transformation of $\bm{k}$ that can reproduce the level of detail variation observed between the query feature regions.}

{In addition to the semantic space, we also seek to fully exploit the structural information in the detail space and generate the aligned query-key pair in this space for better guiding the upsampling process. Unfortunately, unlike the semantic space, it is hard to project the key feature into the detail space to achieve the query-key mutual-alignment. This is because that deep semantic information is more abstract and the features in the detail space generally reside in a higher-dimensional manifold than that in the semantic space. For example, in Fig. \ref{fig:detail}, the blue and green boxes highlight areas belong to the same semantic category, \ie, house, with closer distance in the semantic space. However, they possess highly diverse detail information, \ie, roof and wall, with farther distance in the detail space. In this case, it is hardly feasible to find a proper projection to finely transform the original key $\bm{k}$ to the detail space, which can reproduce the level of detail variations in the query feature.}
% Generally, the features in the detail space are deemed to reside in a higher-dimensional manifold than in the semantic space. This is because the regions with similar semantics usually contain highly diverse details, while the number of semantic categories remains limited. For example, in Fig. \ref{fig:detail}, the blue and green boxes highlight areas with different details, even though they belong to the same semantic category.

 %for the query-key pair alignment in the detail space,
%due to the more abstract characteristics of deep semantic information, it is hardly feasible to directly transform the original key $\bm{k}$ in the semantic space to the detail space. This can be easily understood from Fig. \ref{fig:detail}. As seen, the two regions marked by the two boxes exhibit similar content in the semantic space but substantially differ in the detail space.
% Generally, it is extremely difficult to find a transformation of $\bm{k}$ that can reproduce the level of detail variations observed between the two marked regions of the query feature. 
% Hence, it is impractical to execute the query-key mutual-alignment in the detail space, 

Confronted with the above issue, we propose a detail-aware self-alignment approach that exploits the query in the detail space to generate the aligned key feature for the space. One intuitive way is to directly regard $\bm{q}$ as both the query and key features in the detail space. However, we further analyze that to avoid significant fluctuations in local similarity scores, the real key feature is generally expected to have the property of smoothness \cite{vaswani2017attention,xu2023self}, like the original key $\bm{k}$ in the semantic space containing minimal semantically irrelevant noises. Motivated by this analysis, we propose to execute a simple Gaussian smoothing operator on $\bm{q}$ to suppress high-frequency noisy details to obtain a smoothed key feature $\bm{q}_{GS}\in\mathbb{R}^{HW \times D}$ in the detail space.

Then, for pixel $i$, the self-similarity 
${(\bm{s}_d)}_i$ for the aligned pair $\bm{q}$ and $\bm{q}_{GS}$ in the detail space
is computed as: \vspace{-1mm}
\begin{equation} \label{eqn:detail}
{(\bm{s}_d)}_i \triangleq \text{sim}(\bm{q}_{i},{(\bm{q}_{GS})}_{\mathcal{N}(i)}),
\end{equation}
where $\bm{q}_{GS}=\bm{f} \otimes \bm{q}$ and $\bm{f}\in \mathbb{R}^{3 \times 3}$ represents the Gaussian filter with the unit standard deviation. Here $\bm{s}_d\in \mathbb{R}^{HW \times K^2}$ is the detail-aware self-similarity.
% and its role will be validated in Sec.~\ref{sec:ablation}.

% \vspace{1mm}
% \noindent\textit{\textbf{Remark 2:}} \hippo{With such a self-alignment strategy, the calculated local similarity using these query-key feature pairs actually makes sense. This is because} within a local region of a relatively shallow feature, if two pixels exhibit similarity in detail structures, there is a high likelihood that they should possess similar semantics in the upsampled feature $\tilde{\bm{x}}$ \cite{petschnigg2004digital}. 

As seen, on the basis of the original linearly-projected query $\bm{q}$ and $\bm{k}$, we introduce explicit optimization controls to further make them better aligned from both semantic-aware and detail-aware perspectives. Correspondingly, for our method, the similarity score $\bm{s}_i$ in Eq. \eqref{eqn:attn} is designed as: \vspace{-1mm}
\begin{equation} \label{eqn:si}
\bm{s}_i=(\bm{s}_{s})_i+(\bm{s}_{d})_i.
\end{equation}
{Please note that the similarity function $\text{sim}\left(\cdot,\cdot \right)$ for computing $\bm{s}_{s}$ and $\bm{s}_{d}$ is implemented based on learnable parameterized form as formulated in Sec.~\ref{sec:sim} below. Hence, it is unnecessary to assign extra weighting coefficients on these two terms $\bm{s}_{s}$ and $\bm{s}_{d}$ in Eq.~\eqref{eqn:si}.}
% \hippo{Please note that we do not assign weights on these two terms for now, since we will soon introduce the learnable parameterization of the similarity calculation, which would automatically absorb the weights.}

% {Therefore, the final similarity $\bm{s}_i$ in \cref{eqn:attn} can be expressed by} \vspace{-1mm}
% % From \cref{eqn:semantic,eqn:detail}, the final similarity score at pixel $i$ is 
% % \hippo{}
% \begin{equation} \label{eqn:si}
% \bm{s}_i=(\bm{s}_{s})_i+(\bm{s}_{d})_i.
% \end{equation}
% By substituting it into Eq.~\eqref{eqn:attn}, we can easily get the following refreshed similarity-based feature upsampling framework \hippo{as follows:}
% , named as ReSFU:
% \begin{equation} \label{eqn:framework}
% \tilde{\bm{x}}_{i}=\text{Softmax}((\bm{s}_s)_i + (\bm{s}_d)_i)\bm{x}_{\mathcal{N}(i)},
% \end{equation}
% where \hippo{the specific similarity function $\text{sim}(\cdot , \cdot)$ for calculating} $\bm{s}_{s}$ and $\bm{s}_d$, and the neighbor selection $\mathcal{N}\left(i\right)$ will be specifically designed in \cref{sec:sim,sec:sns}%Sec.~\ref{sec:sim} and Sec.~\ref{sec:sns}
% , respectively. 

% where the similarity calculation function $\text{sim}(\cdot,\cdot)$ and the neighbor selection $\mathcal{N}\left(i\right)$ are specifically designed in \cref{sec:sim,sec:sns}%Sec.~\ref{sec:sim} and Sec.~\ref{sec:sns}
% , respectively. 

\vspace{1mm}
% \noindent{\textit{\textbf{Remark 2:}} In the aspect of query-key alignment, the key merits of our design lie in: 1) We not only optimize the mutual feature alignment in the semantic space but also propose the self-alignment from the detail-aware perspective in an explicitly controllable manner; 2) It is well-acknowledged that within a small local region of a relatively shallow feature, if two pixels exhibit similarity in detail structures, there is a high likelihood that they should possess similar semantics in the upsampled feature $\tilde{\bm{x}}$ \cite{petschnigg2004digital}. 
% Thus, it is rational to exploit the self-similarity on query $\bm{q}$ itself for further guiding the upsampling process;
% 3) From the experimental visualization in Fig.~\ref{fig:main} (b), we can clearly observe that the incorporation of the detail-aware similarity favorably encourages fine-grained detail preservation and the smoothed key generation design would indeed bring some performance improvement as validated in Sec.~\ref{sec:ablation}.} %要不要把(2)删掉, hazel：放着吧
\noindent{\textit{\textbf{Remark 2:}} {1) The rationality of the detail-aware self-alignment can be explained from another perspective. Specifically, it is well-acknowledged that within a small local region of a relatively shallow feature, if two pixels exhibit similarity in detail structures, there is a high likelihood that they should possess similar semantics in the upsampled feature $\tilde{\bm{x}}$ \cite{petschnigg2004digital}. Thus, it is reasonable to exploit the self-similarity on query $\bm{q}$ itself for guiding the upsampling process;
2) The incorporation of the detail-aware similarity favorably encourages fine-grained detail preservation and the smoothed key generation design would indeed bring some performance improvement as validated in Sec.~\ref{sec:ablation}.}} {It is worth mentioning that this proposed novel self-alignment technique can be easily transferred to the existing upsampling methods for better
performance, as verified in SM.}

%要不要把(2)删掉, hazel：放着吧

\subsection{PCDC for Flexible Similarity Calculation}\label{sec:sim}
% To introduce more flexibility to the procedure of similarity calculation, we specifically design a parameterized convolution operation, named Paired Central Difference Convolution (PCD-Conv). Different from Central Difference Convolution (CDC) \cite{yu2020searching}, PCD-Conv accepts a pair of inputs, \ie, a key $\bar{\bm{k}}$ and a query $\bar{\bm{q}}$, and the output $\bm{v}$ at any pixel $i$ and channel $g$, is computed as follows:

% Since the conventionally adopted inner product is non-parametric and inflexible, we are motivated to specifically design a parameterized convolution operation that can inherently capture the similarity information. Inspired by the idea of Central Difference Convolution (CDC) \cite{yu2020searching}, which combines the flexibility of convolution and the awareness of local gradient information, we develop Paired Central Difference Convolution (PCD-Conv) to render the local mutual-similarity between key-query pairs. 

To compute $\bm{s}_{s}$ and $\bm{s}_d$ in Eq. \eqref{eqn:si}, instead of adopting the fixed inner product-based manner for $\text{sim}(\cdot, \cdot)$ in \cite{lu2022sapa,liu2023point}, here we aim to specifically design a parameterized convolution operation to flexibly model the inherent relevance between every query-key pair, \textit{i.e.}, $\bm{q}_{GF}$ and $\tilde{\bm{k}}$, $\bm{q}$ and $\bm{q}_{GS}$, for more accurate similarity calculation.
\subsubsection{Paired Central Difference Convolution}
For a vanilla convolution layer with $G$ groups, given an arbitrary input $\bar{\bm{x}} \in \mathbb{R}^{HW\times D}$ with $D$ input channels, each element $\bm{v}_{il}$ in the convolution result $\bm{v} \in \mathbb{R}^{HW\times L}$ with $L$ output channels is calculated by:
% \begin{equation} \label{eq:vanilla}
% \bm{v}_{il} =\sum_{d=g D / G}^{(g+1)D/G-1} \sum_{j_n \in \mathcal{N}(i)} \bm{w}_{n\tilde{d}l}\  \bar{\bm{x}}_{j_nd} +\bm{b}_l,
% \end{equation}
\begin{equation} \label{eq:vanilla}
\bm{v}_{il} =\sum_{d=g D / G}^{(g+1)D/G-1} \sum_{n=1}^{K^2} \bm{w}_{n\tilde{d}l}\  \bar{\bm{x}}_{j_nd} +\bm{b}_l, \hspace{1mm}  j_n \in \mathcal{N}(i),
\end{equation}
{where $i$ is the pixel index; the output channel index $l\in \{0,\ldots,L-1\}$; $g=\lfloor \frac{lG}{L} \rfloor$ is the group index ranging from 0 to $G-1$; $d$ is the input channel index; $j_n \in \mathcal{N}(i)$ indexes a neighbor of the pixel $i$, with $n=1,\cdots,K^2$;
$\bm{w}_{n\tilde{d}l}$ is an element of $\bm{w} \in \mathbb{R}^{K^2\times D/G \times L}$ representing the grouped convolution weights; the index $\tilde{d}=d\%(\frac{D}{G})$, where $\%$ is the modulo operation; 
% $\bar{\bm{x}}_{j_nd}$ is the element of $\bar{\bm{x}}$, where $j_{n}$ indexes the neighbors in $\bar{\bm{x}}$ for the output pixel $i$, ranging from $j_{1}$ to $j_{K^{2}}$;
$\bm{b}\in \mathbb{R}^{L}$ is the bias.}

% \hippo{where $g=\lfloor \frac{lG}{L} \rfloor$ is the group index ranging from 0 to $G-1$; $\bm{w} \in \mathbb{R}^{K^2\times D/G \times L}$ denotes the convolution weights; $n$ is the spatial index of $\bm{w}$ ranging from 1 to $K^2$; $\tilde{d}=d\%(\frac{D}{G})$ is the 2nd-dimensional index of $\bm{w}$; $\bm{b}\in \mathbb{R}^{L}$ is the bias.}

% \hippo{First, let's recall the mathematical formulation of a vanilla convolution layer with $G$ groups. For any input $\bar{\bm{x}} \in \mathbb{R}^{HW\times D}$, the $i$-th pixel and $l$-th channel of the convolution output $\bm{v} \in \mathbb{R}^{HW\times L}$ can be calculated by }

\begin{figure}[!t]
  \centering
\includegraphics[width=0.99\linewidth]{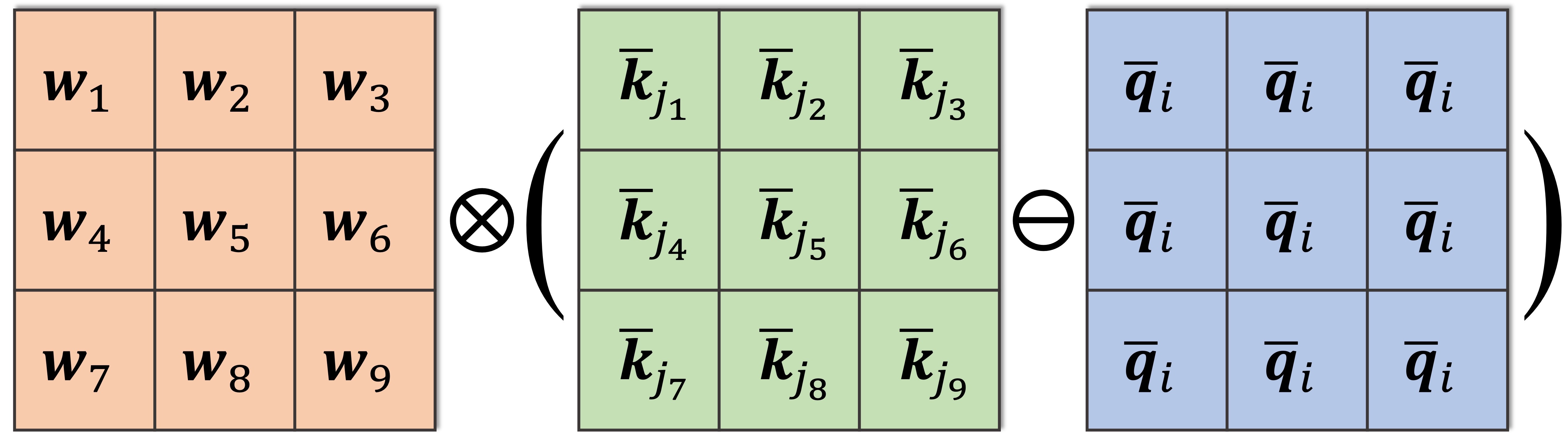}
  % \caption{{A simplistic illustration of the PCDC operation in Eq.~(\ref{eqn:pcd}) for any input channel $d$ and any output channel $g$ where the bias term $\bm{b}_{g}$ is omitted for brevity.}}
  \vspace{-2mm}
    \caption{Illustration of the PCDC operation in Eq.~(\ref{eqn:pcd}) for any input channel $d$ and any output channel $l$ where the bias term $\bm{b}_{l}$ is omitted for brevity. Here {$\otimes$ is the convolution operation and $\ominus$ is the element-wise subtraction operation.}}
  \label{fig:pcdc}
  \vspace{-3mm}
\end{figure}

Inspired by the ability of central difference convolution \cite{yu2020searching} in {combining the flexibility of learnable convolution and the awareness of local gradient information}, here we propose a paired central difference convolution (PCDC) to capture the relations between the query and key features. Specifically, following the framework of the vanilla grouped convolution in Eq. (\ref{eq:vanilla}), we replace its input $\bar{\bm{x}}$ with the ``paired central difference'' between any aligned query-key pair, $\bar{\bm{q}}$ and $\bar{\bm{k}}$, and then obtain the PCDC output $\bm{v}$ as:
%{Specifically, we calculate the central difference between any aligned query-key pair, denoted as $\bar{\bm{q}}$ and $\bar{\bm{k}}$, and use the  to replace the input the grouped convolution in Eq.~\eqref{eq:vanilla}, we can derive the PCDC output as:}
% \begin{equation} \label{eqn:pcd}
% \bm{v}_{il} =\sum_{d=g D / G}^{(g+1)D/G-1} \sum_{j_n \in \mathcal{N}(i)} \bm{w}_{n\tilde{d}l}\  (\bar{\bm{k}}_{j_nd}-\bar{\bm{q}}_{id}) +\bm{b}_l.
% \end{equation}
\begin{equation} \label{eqn:pcd}
\bm{v}_{il} =\! \! \sum_{d=g D / G}^{(g+1)D/G-1} \sum_{n=1}^{K^2} \bm{w}_{n\tilde{d}l}\  (\bar{\bm{k}}_{j_nd}-\bar{\bm{q}}_{id}) +\bm{b}_l, \hspace{1mm}  j_n \in \mathcal{N}(i).
\end{equation}
{The closer $\bar{\bm{k}}_{j_nd}-\bar{\bm{q}}_{id}$ is to 0, the more similar $\bar{\bm{q}}_{id}$ and $\bar{\bm{k}}_{j_nd}$ are. From~\cite{yu2020searching}, this difference form focuses on capturing center-oriented gradient information which carries semantic information and can contribute to generating meaningful upsampling kernels for boosting the higher-quality generation of HR features.} {For a better understanding of the working mechanism of the proposed PCDC, we provide a visual illustration of Eq.~\eqref{eqn:pcd} with the kernel size $K$ as 3. As observed from Fig. \ref{fig:pcdc}, through the learnable convolution weights, the differences between the central query element $\bar{\bm{q}}_{i}$ and its neighboring key elements $\bar{\bm{k}}_{j_n}$ can be well captured and flexibly processed. Naturally, {the PCDC output $\bm{v}$ has the capability to encode the local similarity information between $\bar{\bm{q}}$ and $\bar{\bm{k}}$.}

}%, thus generating faithful local information for assisting in the accurate similarity computation

% \hippo{Then, we formulate our core design in our parameterized similarity calculation module, \ie, the paired central difference convolution (PCDC), which takes in two inputs, $\bar{\bm{q}}$ and $\bar{\bm{k}}$, and outputs the local similarity-related information:}
% \begin{equation} \label{eqn:pcd}
% \bm{v}_{il} =\sum_{d=g D / G}^{(g+1)D/G-1} \sum_{j_n \in \mathcal{N}(i)} \bm{w}_{n\tilde{d}l}\  (\bar{\bm{k}}_{j_nd}-\bar{\bm{q}}_{id}) +\bm{b}_l,
% \end{equation}
% \hippo{where the denotations are the same as Eq. (\ref{eq:vanilla}). Please refer to Fig. \ref{fig:pcdc} for a better understanding of the working mechanism of PCDC. It can be observed that differences between $\bar{\bm{q}}_i$ and its neighboring pixels on $\bar{\bm{k}}$ can be well captured and flexibly processed with convolutional weights, thus producing faithful local similarity information in $\bm{v}$.  Reorganizing Eq.~\eqref{eqn:pcd} gives:

Equivalently, Eq.~\eqref{eqn:pcd} can be decomposed as:
% \begin{equation} \label{eqn:pcd2}
% \begin{split}
% \bm{v}_{il} =&\sum_{d=g D / G}^{(g+1)D/G-1} \sum_{j_n \in \mathcal{N}(i)}  \bm{w}_{n\tilde{d}l}\  \bar{\bm{k}}_{j_nd}\\
% &-\sum_{d=g D / G}^{(g+1)D/G-1} \bar{\bm{q}}_{id}\left(\sum_{j_n \in \mathcal{N}(i)} \bm{w}_{n\tilde{d}l}\right) +\bm{b}_l.
% \end{split}
% \end{equation}
\begin{equation} \label{eqn:pcd2}
\begin{split}
\bm{v}_{il} =&\sum_{d=g D / G}^{(g+1)D/G-1} \sum_{n=1}^{K^2}  \bm{w}_{n\tilde{d}l}\  \bar{\bm{k}}_{j_nd}\\
&-\sum_{d=g D / G}^{(g+1)D/G-1} \bar{\bm{q}}_{id}\left(\sum_{n=1}^{K^2} \bm{w}_{n\tilde{d}l}\right) +\bm{b}_l, \hspace{1mm} j_n \in \mathcal{N}(i).
\end{split}
\end{equation}
% -\sum_{d=g D / G}^{(g+1)D/G-1} \bar{\bm{q}}_{id}\sum_{j_n \in \mathcal{N}(i)} \bm{w}_{n\tilde{d}l} +\bm{b}_l,
We can find that the proposed PCDC inherently consists of a vanilla grouped convolution on $\bar{\bm{k}}$ with weight $\bm{w}$ (the first term in Eq. (\ref{eqn:pcd2})), and a $1\times 1$ convolution on $\bar{\bm{q}}$ with the weight as the aggregation of $\bm{w}$ in the spatial dimension (the second term in Eq. (\ref{eqn:pcd2})). These operations can be easily and efficiently implemented based on PyTorch \cite{paszke2017automatic} (see the pseudocode in SM).

\begin{figure}[t]
  \centering
  % \vspace{-0.2cm}
\includegraphics[width=0.98\linewidth]{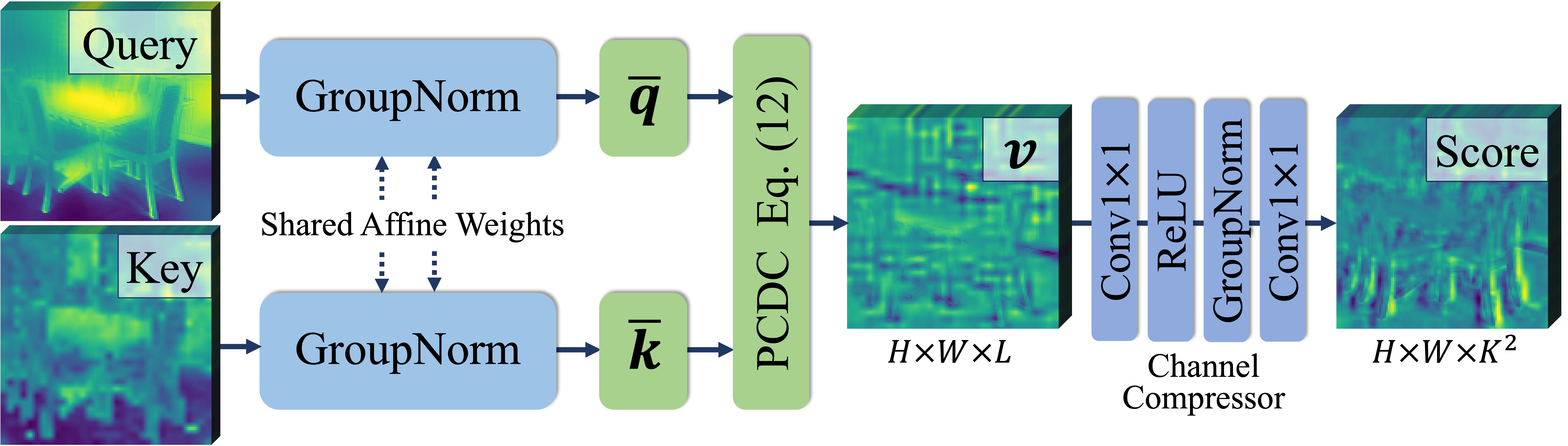}
  \vspace{-0.2cm}
  \caption{Illustration of PCDC-Block for similarity calculation between the aligned query-key pair.}
  \label{fig:pcd}
  \vspace{-0.3cm}
\end{figure}

\begin{figure}[t]
  \centering
  % \vspace{-0.2cm}
\includegraphics[width=0.95\linewidth]{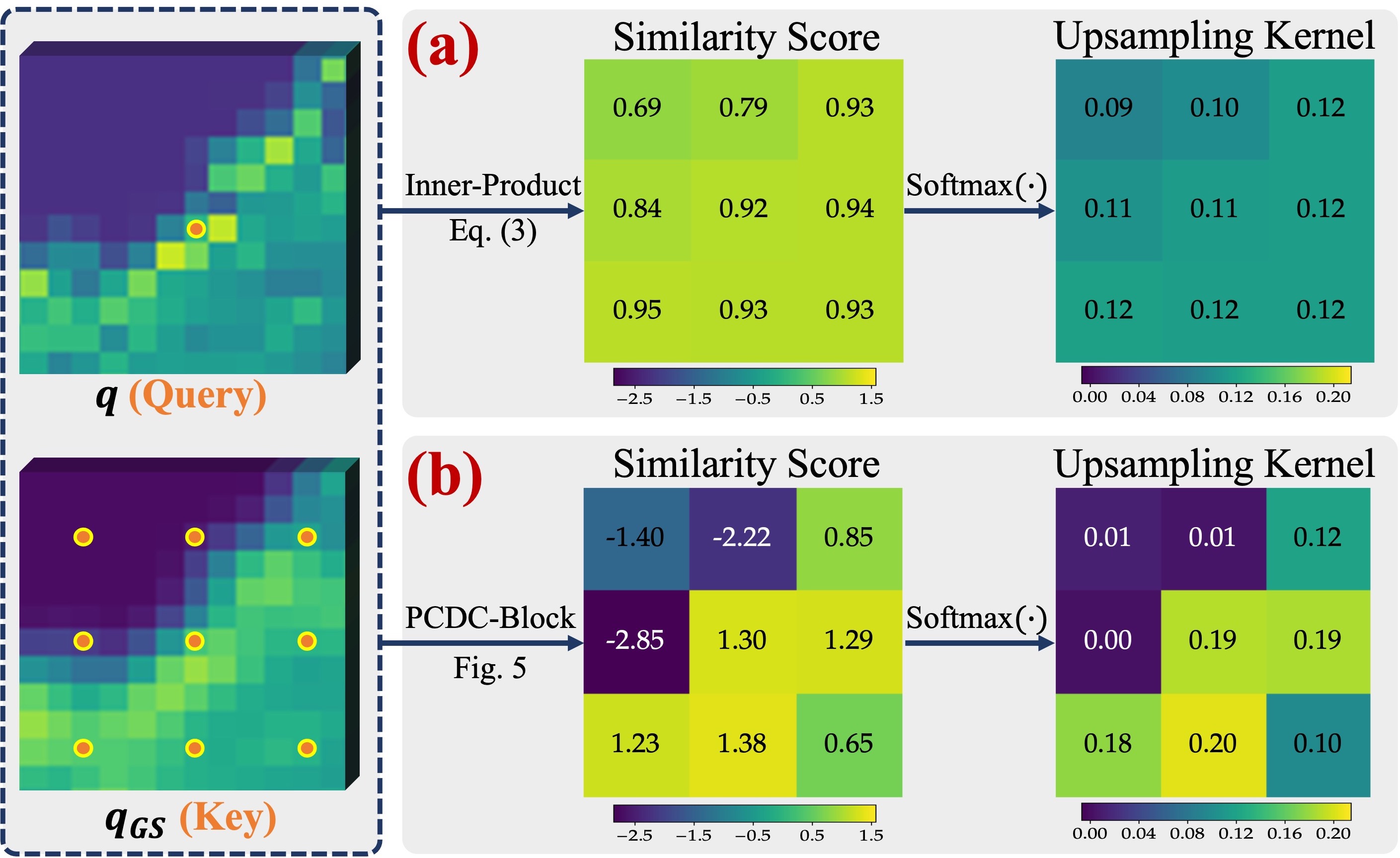}
\vspace{-0.2cm}
  \caption{Visualization of similarity scores and softmax values (\ie, upsampling kernels) computed through (a) inner product in Eq.~\eqref{eqn:sapa} and (b) PCDC-Block illustrated in Fig.~\ref{fig:pcd}. }
  \label{fig:sim}
  \vspace{-0.3cm}
\end{figure}

\subsubsection{PCDC-Block for Similarity Calculation}
Based on the capability of the proposed PCDC in capturing the local similarity between the query-key pair, here we construct a PCDC-Block to finally implement the function $\text{sim}(\cdot, \cdot)$ for similarity computation as given in Fig. \ref{fig:pcd}.

Concretely, for each aligned query-key pair, \textit{i.e.}, $\bm{q}_{GF}$ and $\tilde{\bm{k}}$, $\bm{q}$ and $\bm{q}_{GS}$, they are first separately input to a group normalization layer with shared affine parameters to get the normalized query-key pair $\bar{\bm{q}}$ and $\bar{\bm{k}}$. Then the normalized pair is passed through a PCDC computation layer, \textit{i.e.}, Eq. \eqref{eqn:pcd2}, to obtain the intermediate result $\bm{v}$. Then by feeding $\bm{v}$ to a channel compressor to transform the channel number from $L$ to $K^{2}$, we can get the corresponding similarity scores for every query-key pair, \textit{i.e.}, $\bm{s}_{s}$ and $\bm{s}_{d}$. 
% {Here the channel compressor is a ConvBlock described in Sec. \ref{sec:exp}.}
 Here the channel compressor is sequentially composed of a $1\times 1$ convolution layer, a ReLU layer, a group normalization layer, and a $1\times 1$ convolution layer, where these two convolution layers are both with 4 groups and the intermediate channel size is 128.

{Then the complete calculation procedures for semantic-aware similarity scores in Eq. (\ref{eqn:semantic}) and detail-aware similarity scores in Eq. (\ref{eqn:detail}) are:}
% \begin{equation} \label{eqn:pcdc_s}
% \bm{s}_s = \text{PCDC-Block}_{\theta_s}(\bm{q}_{\mathit{GF}},\tilde{\bm{k}}),
% \end{equation}
% \begin{equation} \label{eqn:pcdc_d}
% \bm{s}_d = \text{PCDC-Block}_{\theta_d}(\bm{q},\bm{q}_{\mathit{GS}}),
% \end{equation}
{
\begin{align}
\bm{s}_s &= \text{PCDC-Block}_{\bm{\theta}_s}(\bm{q}_{\mathit{GF}},\tilde{\bm{k}}), \label{eqn:pcdc_s}\\
\bm{s}_d &= \text{PCDC-Block}_{\bm{\theta}_d}(\bm{q},\bm{q}_{\mathit{GS}}), \label{eqn:pcdc_d}
\end{align}
where the operator PCDC-Block is illustrated by Fig. \ref{fig:pcd}; $\bm{\theta}_s$ and $\bm{\theta}_d$ represent the network parameters involved in GroupNorm, PCDC, and the channel compressor for computing $\bm{s}_{s}$ and $\bm{s}_{d}$, respectively.}

To better understand the inherent advantages of our proposed PCDC-Block over the traditional inner product-based form in Eq.~\eqref{eqn:sapa} for similarity computation, here we provide an intuitive experimental display with the kernel size $K$ as 3. Fig.~\ref{fig:sim} compares the similarity scores computed by these two methods for an image patch at pixels indicated by the orange circles in $\bm{q}$ and $\bm{q}_{GS}$. As seen, even along clear object boundaries, (a) the inner product scores exhibit minor variations within a narrow range, and then the {upsampling kernels} for these neighboring nine pixels are highly similar, weakening the discrimination of feature semantics. However, (b) our PCDC-Block generates considerably more discriminative upsampling kernels that are more accurately aligned with the boundary structures and can effectively alleviate the blurring issue.

\begin{figure}[t]
  \centering
  % \vspace{-0.2cm}
\includegraphics[width=0.98\linewidth]{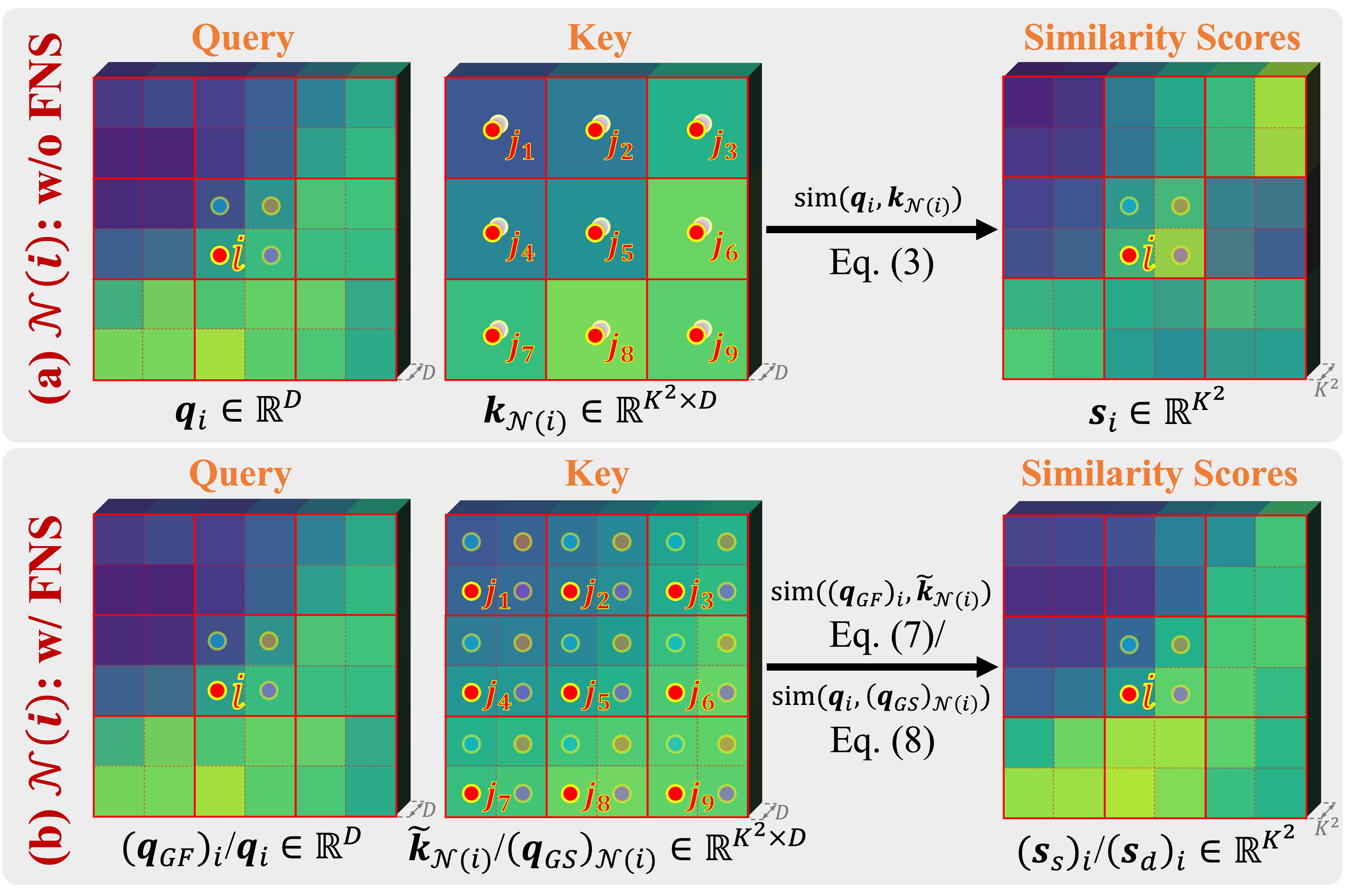}
  \vspace{-2mm}
  \caption{{Illustration of the query-key similarity calculation process under different neighbor selection strategies for $\mathcal{N}(i)$: (a) without FNS and (b) with FNS. Here the kernel size $K=3$, and the neighbors $\mathcal{N}(i)$ of the highlighted red pixel $i$ are $j_1, \cdots, j_9$. Pixels marked in circles with the same color in the query and key features are paired for computing the similarity scores.} } %The similarity score is computed on the query-key pair marked with the same color and shape. }
  \label{fig:fns}
  \vspace{-0.4cm}
\end{figure}
% elements with the same color and shape are paired for computing the similarity score and the upsampled result.

\vspace{1mm}
\noindent\textit{\textbf{Remark 3:}} 
% Compared with the conventional inner product operator in Eq.~(\ref{eqn:sapa}), the proposed PCDC-Block-based similarity has specific merits: 1) With the well-aligned query-key pair with similar local contents as inputs, such a parameterized convolution is inherently well-suited for the local similarity calculation; 2) As given in Eq.~\eqref{eqn:pcd}, the proposed PCDC layer is designed to capture the differences between $\bar{\bm{q}}_i$ and $\bar{\bm{k}}_{\mathcal{N}(i)}$, which can finely manifest their similarity; 3) The incorporation of learnable weights makes the similarity computation more flexible. The performance gains brought by these advantages will be validated in Sec.~\ref{sec:ablation}.
Compared with the conventional inner product operator in Eq.~(\ref{eqn:sapa}), the proposed PCDC-Block-based similarity calculation has specific merits: 1) With the well-aligned query-key pair containing similar local contents as inputs, such a paired central difference design is inherently well-suited for manifesting the similarity between $\bar{\bm{q}}$ and $\bar{\bm{k}}$, and is proficient in detecting intricate structural details; %such a parameterized convolution is inherently well-suited for the local similarity calculation; 2) As given in Eq.~\eqref{eqn:pcd}, the proposed PCDC layer is designed to capture the differences between $\bar{\bm{q}}_i$ and $\bar{\bm{k}}_{\mathcal{N}(i)}$, which can finely manifest their similarity;
2) The incorporation of learnable weights makes the similarity computation more flexible, which is helpful for accurate upsampling. The performance gains brought by these advantages will be validated in Sec.~\ref{sec:ablation}.

\subsection{Fine-grained Neighbor Selection}\label{sec:sns}
From Sec.~\ref{sec:revisit}, for the similarity-based feature upsampling pipeline, the neighborhood selection $\mathcal{N}\left(i\right)$ exists in two procedures, \ie, 1) the selection on key feature for query-key similarity calculation and 2) the selection on value feature for the weight-value computation in Eq.~\eqref{eqn:attn}. For the existing approaches implemented based on Eqs.~\eqref{eqn:attn}\eqref{eqn:sapa}, they select the neighbors $\mathcal{N}\left(i\right)$ directly on LR key feature $\bm{k}$ and LR value feature $\bm{x}$ in a grid-wise manner, respectively. This would possibly result in mosaic artifacts as analyzed in Sec.~\ref{sec:revisit}. To address this problem, {for our proposed refreshed pipeline mainly implemented based on Eqs.~\eqref{eqn:attn}~\eqref{eqn:si}~\eqref{eqn:pcd2}}, we develop a simple yet effective fine-grained neighbor selection (FNS) strategy and choose $\mathcal{N}\left(i\right)$ on HR key feature (\ie, $\tilde{\bm{k}}$ or $\bm{q}_{GS}$) for Eq.~\eqref{eqn:pcd2} and HR value feature (\ie, bilinearly upsampled $\bm{x}$) for Eq.~\eqref{eqn:attn}, respectively.

{To understand our FNS strategy more intuitively, Fig.~\ref{fig:fns} presents the query-key similarity calculation process with the kernel size $K=3$ for a $\times 2$ upsampling. 
As seen, for the case without FNS, the four HR query elements with different colors in a $2 \times 2$ local region share the same set of neighboring LR key pixels indexed by $j_1, \cdots, j_9$. However, for our proposed FNS, the neighbor selection is executed on the HR key feature. In this manner, the four HR query elements in a $2 \times 2$ local region would separately pair with different nine neighboring HR key elements. Such a smoother query-key pair selection manner would promote the smoothness of similarity scores for adjacent HR query pixels with different neighbors in the key feature. }
Albeit simple, this FNS strategy can be implemented efficiently without introducing any additional computational cost, and can fundamentally eliminate mosaic artifacts, which can be observed by comparing the upsampled features $\tilde{\bm{x}}$ in Figs. \ref{fig:main} (a) and (b). Please refer to Fig.~\ref{fig:ablation_fns} for more experimental validation.

In practical implementation, 1) for Eq. \eqref{eqn:pcd2}, the FNS strategy can be seamlessly combined with the PCDC layer by executing PCDC with a dilation rate equal to the upsampling ratio; 
% Note that from Sec.~\ref{sec:sim}, the input key features $\tilde{\bm{k}}$ and $\bm{q}_{GS}$ of the PCDC-Block are HR features, making the neighbor selection smooth;
2) To adopt our FNS in Eq.~\eqref{eqn:attn}, a naive implementation is to bilinearly upsample $\bm{x}$ to obtain the HR one. Fortunately, we eliminate this need by technically encapsulating the weight-value computation and the bilinear upsampling within a single CUDA package, which helps avoid explicitly computing the temporary bilinear upsampling results. These careful designs enable the proposed FNS to be easily encapsulated into different backbones. %{code.......}

% \begin{wrapfigure}{R}{0.58\textwidth}
% \vspace{-1.95cm}
% \begin{figure}[h]
% % \vspace{-4mm}
% \centering
%     \begin{minipage}{0.495\textwidth}

\begin{algorithm}[t]
\small
{
\caption{{Algorithm of ReSFU for feature upsampling}}
\label{alg:resfu}
\begin{algorithmic}[1] %每行显示行号  
\Require LR deep feature $\bm{x}\in \mathbb{R}^{hw\times C}$, HR guidance feature $\bm{y} \in \mathbb{R}^{HW\times c}$; parameters of linear projections $\bm{\theta}_{\bm{k}}$ and $\bm{\theta}_{\bm{q}}$, hyperparamters in guided filter $r$ and $\epsilon$, parameters of the PCDC-Blocks $\bm{\theta}_s$ and $\bm{\theta}_d$.
\Ensure Upsampled deep feature $\tilde{\bm{x}}\in \mathbb{R}^{HW\times C}$.
\State $\bm{k}, \bm{q} \gets \text{Proj}_{\bm{\theta}_{\bm{k}}}(\bm{x}), \text{Proj}_{\bm{\theta}_{\bm{q}}}(\bm{y})$  \Comment{linear projection}
\State $\tilde{\bm{k}} \gets \text{Bilinear}(\bm{k}, \text{size}=(H,W))$
\State $\bm{q}_{\mathit{GF}} \gets \text{Guided\_Filter}_{r,\epsilon}(\bm{q},\tilde{\bm{k}})$ \Comment{Eqs. (\ref{eqn:gf2})(\ref{eqn:gf3})}
\State $\bm{s}_s \gets \text{PCDC-Block}_{\bm{\theta}_s}(\bm{q}_{\mathit{GF}}, \tilde{\bm{k}})$ \Comment{Eq. (\ref{eqn:pcdc_s})}
% \State $\bm{q}_{\mathit{GS}} \gets \text{Gaussian\_Smoothing}(\bm{q})$
\State $\bm{q}_{\mathit{GS}} \gets \bm{f}\otimes \bm{q}$ \Comment{$\bm{f}$ is the $3\times 3$ Gaussian filter }
\State $\bm{s}_d \gets \text{PCDC-Block}_{\bm{\theta}_d}(\bm{q},\bm{q}_{\mathit{GS}})$ \Comment{Eq. (\ref{eqn:pcdc_d})}
\State $\bm{s} \gets \bm{s}_s + \bm{s}_d$ \Comment{Eq. (\ref{eqn:si})}
% \State $\tilde{\bm{x}} \gets \text{FNS\_Attn}(\bm{s}, \bm{x})$ \Comment{Eq. (\ref{eqn:attn}), Fig. \ref{fig:fns} (b)}
\State $\tilde{\bm{x}}_{i}\gets\text{Softmax}(\bm{s}_{i})\bm{x}_{\mathcal{N}(i)}; \mathcal{N}(i)$: w/ FNS \Comment{Eq. (\ref{eqn:attn}), Fig. \ref{fig:fns} (b)}
\end{algorithmic} }
\end{algorithm}
% \end{minipage}
% \vspace{-4mm}
% \end{figure}

% \vspace{1mm}
\subsection{{Overall Framework of ReSFU}}
% \noindent\textit{\textbf{Remark 4:}}
{
% Collectively, for the three key components in Eqs. (\ref{eqn:attn})(\ref{eqn:sapa}), we systematically construct the similarity-based upsampling methodology, described by Alg. \ref{alg:resfu}.
Overall, for the three key components in Eqs. (\ref{eqn:attn})(\ref{eqn:sapa}), we systematically reform the similarity-based upsampling methodology. Specifically, we explicitly optimize the query-key alignment for facilitating the computation of $\bm{s}_i$, design a flexible similarity function $\text{sim}(\cdot,\cdot)$, and develop a fine-grained neighbor selection strategy for $\mathcal{N}(i)$ as derived in Secs. \ref{sec:align}, \ref{sec:sim}, and \ref{sec:sns}, respectively. We accordingly construct a refreshed similarity-based feature upsampler, named ReSFU. The overall framework of ReSFU is illustrated in Fig.~\ref{fig:main} (b) and the complete computation procedure is summarized in Alg. \ref{alg:resfu}. Notably, the proposed three designs harmoniously form an inseparable whole with each part being indispensable (see Sec.~\ref{sec:ablation}). The specific and meticulous design of each part ultimately enables ReSFU to be applicable to different network structures, which will be validated in Sec. \ref{sec:exp} below.}

\vspace{1mm}
\noindent{\textit{\textbf{Remark 4:}} Please be aware that we only continue the similarity-based research paradigm introduced by SAPA. Nevertheless, we have made highly comprehensive and systematic fundamental enhancements to SAPA across multiple aspects, including problem formulation, essence discovery, and methodological designs. All these efforts promote our proposed ReSFU to possess wider application scenarios than SAPA. These merits will be fully validated through extensive experiments based on diverse configurations.}
 % Furthermore, our proposed novel techniques, such as, query-key alignment and PCDC-Block, can be easily transferred to the existing upsampling methods for better performance. 

\section{Experiments} \label{sec:exp} 
% \hippo{To demonstrate the superior applicability of our proposed ReSFU to different network architectures, we first conduct comprehensive experiments for the semantic segmentation task. Then, we provide detailed model verification and ablation studies to showcase the rationale of our model designs. Finally, to better showcase the generality of ReSFU, we further perform experiments on a wider range of datasets or tasks, including medical image segmentation, instance segmentation, and panoptic segmentation. }

In this section, we comprehensively substantiate the applicability of our proposed ReSFU by conducting extensive experiments on various network structures for the semantic segmentation task. Then, we provide detailed model verification and ablation studies to clearly show the working mechanism of ReSFU and validate the rationality of every involved component. Finally, in order to better demonstrate the universality of the proposed ReSFU, we extend it to more application scenarios with a variety of datasets, including medical image segmentation, instance segmentation, panoptic segmentation{, object detection, and monocular depth estimation.}

\subsection{Experimental Setup}

\vspace{1mm}
\noindent\textbf{Implementation details.} 
% {For the mutual alignment in Sec.~\ref{sec:mutual}, the radius $r$ and the regularization weight $\epsilon$ in Eq.~\eqref{eqn:gf} are empirically set to 8 and 0.001, respectively.} 
For similarity calculation in Eq.~\eqref{eqn:pcd2}, the number of output channels $L$ and groups $G$ are 32 and 4, respectively. The projection dimension $D$ is 32 and the kernel size $K$ is 3. 
%For different network structures, we adopt their default training settings, including loss function and learning rate in the off-the-shelf MMSegmentation toolkit~\cite{mmseg2020}. 
For all the involved networks, we modify the default bilinear/nearest neighbor upsampling method by substituting it with alternative upsamplers (replacement details are described in the experimental comparisons below), while keeping other training settings constant for end-to-end training on NVIDIA V100 GPUs. Please refer to SM for more details. %{Our source code is released at...} 

%\hippo{Our empirical findings indicate that increasing the learning rate of the PCDC-Block sometimes improves performance across various networks.}

% \hippo{For all the feature map visualizations throughout this section, we use Principal Component Analysis (PCA) \cite{abdi2010principal} to reduce the feature maps to 3 channels, representing RGB values. Since the feature extractors are not frozen during training with different upsamplers, there may be significant variations in color tone across different visualizations.}
% \hippo{In terms of hyperparameters, we empirically set the radius for GF-based optimization $r$ to 8 and set $\epsilon$ to 0.001. The channel size $L$ of the intermediate result $\bm{v}$ in PCDC-Block is set to 32, the group number $G=4$, and the kernel size $K=3$. The channel compressor is composed of two $1\times 1$ convolution layers of group number equal to 4, with ReLU and group normalization layers in between, with the intermediate channel sizes to be 128 and the output channel size equal to $K^2=9$. The experiments are conducted on NVIDIA V100 GPUs.}

\begin{figure*}[t]
  \centering
  % \vspace{-0.2cm}
\includegraphics[width=0.99\linewidth]{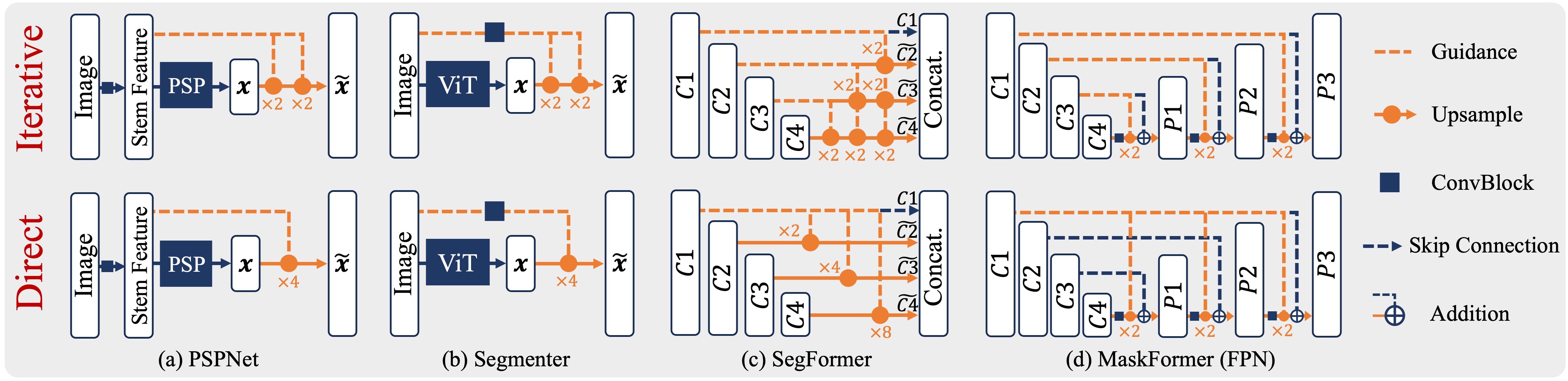}
  % \vspace{-0.25cm}
  \caption{{Illustration of different backbones, \ie, (a) PSPNet, (b) Segmenter, (c) SegFormer, and (d) MaskFormer (FPN) with iterative (upper row) and direct (lower row) upsampling manners. Please refer to SM for more details.}}
  \label{fig:heads}
  % \vspace{-0.2cm}
\end{figure*}

\vspace{1mm}
\noindent\textbf{Comparison methods.} 
% Except the network structure with the FPN neck~\cite{lin2017feature} which utilizes the nearest neighbor interpolation, the default upsampling strategy for other network backbones is bilinear. We aim to replace the default upsampler with different feature upsampling strategies (replacement details are described in the experimental comparisons below). 
Consistent to the latest works~\cite{liu2023learning,fu2024featup}, we compare the proposed ReSFU with the following feature upsamplers along this research field, including:
% \hippo{The default upsampler for the involved networks is bilinear interpolation, except for the architecture with the FPN neck~\cite{lin2017feature} where the default one is nearest neighbor interpolation. Following \cite{liu2023learning,fu2024featup}, we further compare with other feature upsampling methods:
\begin{itemize}
\item Deconvolution. It can be executed via the function `ConvTranspose2d' in PyTorch with the kernel size and stride number set to the upsampling ratio.
    \item Pixelshuffle \cite{shi2016real}. It consists of a convolution layer for channel expansion and a reshape process for the element reorganization between the channel dimension and the spatial dimension.
    % This upsampling method is composed of a convolution layer followed by a reassembling PixelShuffle layer;
    % \item Stack-JBU \cite{fu2024featup}. It is a parameterized version of Joint Bilateral Upsampler (JBU) \cite{kopf2007joint} adopted in FeatUp \cite{fu2024featup}. Since FeatUp also incorporates multi-view reconstruction loss which is orthogonal to our model design, we only compare with its adopted upsampler. It is worth noting that FeatUp also has an implicit variant that demands an arduous test-time training process. We refrain from comparing with this variant due to the excessive time requirements.
    \item Stack-JBU \cite{fu2024featup}. It is a parameterized version of joint bilateral upsampler (JBU) \cite{kopf2007joint} adopted in FeatUp \cite{fu2024featup}. As analyzed in Sec.~\ref{sec:related}, the entire multi-view framework in FeatUp is orthogonal to our upsampling design. Hence, we only compare with the upsampler module in FeatUp. %\hippo{Note that FeatUp also has an implicit variant with an MLP upsampler. However, this variant is not included in our comparisons for its significantly longer inference time.}
    \item CARAFE \cite{wang2019carafe}. We adopt its default setting with the kernel size as 3.
    \item IndexNet \cite{lu2019indices}. Following~\cite{liu2023learning}, we select its `HIN' version. Please note that IndexNet was initially conceived for strict encoder-decoder-based backbones.
    % \item A2U \cite{dai2021learning}. The `dynamic-cs-d†' version is used. Like IndexNet, it is also evaluated on partial backbones in our experiments;
    \item FADE \cite{lu2022fade}. Since its gating version requires that the guidance feature has the same channel size as the encoder feature, we choose its no-gating version.
    \item SAPA \cite{lu2022sapa}. Following \cite{liu2023learning,fu2024featup}, we use its base version for better stability where the kernel size is 5.
    \item DySample \cite{liu2023learning}. We select its `S+' version due to its better overall performance on different network structures as presented in \cite{liu2023learning}.
\end{itemize}

\begin{figure*}[t]
  \centering
  % \vspace{-0.2cm}
\includegraphics[width=0.99\linewidth]{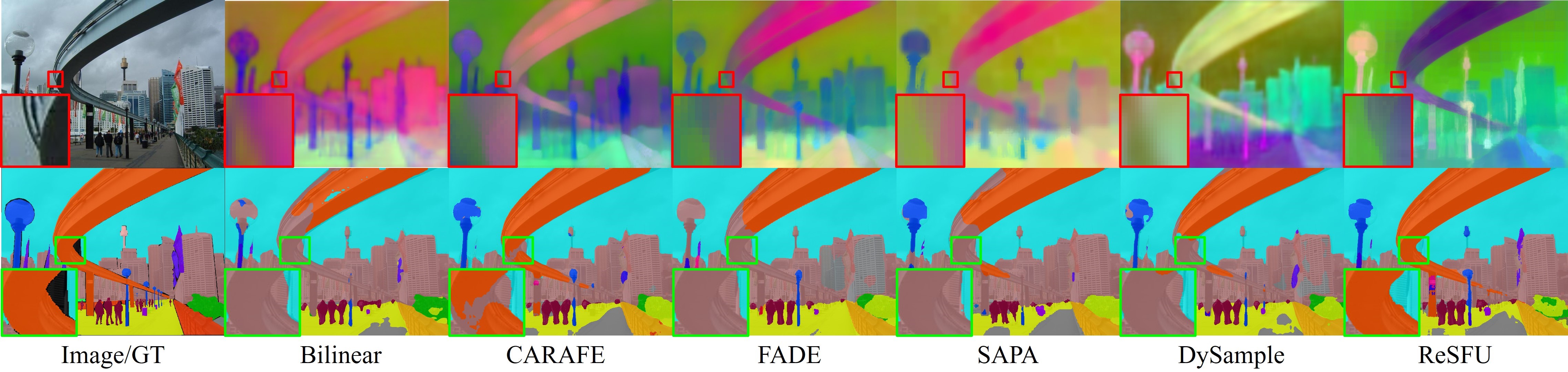}
  \vspace{-0.2cm}
  \caption{Visual comparison of the $\times 4$ upsampled features $\tilde{\bm{x}}$ (top) and predicted masks (bottom) on PSPNet-ResNet101.}
  \label{fig:vis_psp}
  \vspace{-0.3cm}
\end{figure*}

\subsection{Semantic Segmentation on Various Architectures} \label{sec:sem_seg}
\subsubsection{{Experimental Settings}}
In this section, for the semantic segmentation task, we verify the effectiveness of our proposed ReSFU by conducting comprehensive experiments on various network structures with different types of popular backbones including ResNet~\cite{he2016deep}, Vision TransFormer (ViT)~\cite{dosovitskiy2021an}, Mix Transformer (MiT)~\cite{xie2021segformer}, and  Swin Transformer~\cite{liu2021swin}. For the selection of network structures, {we consider not only the hierarchical upsampling manner adopted by the existing baselines,  \eg, SegFormer~\cite{xie2021segformer}, but also the non-hierarchical upsampling manner, \eg, PSPNet~\cite{zhao2017pyramid} and Segmenter~\cite{strudel2021segmenter}, as well as MaskFormer with the FPN head~\cite{cheng2021per}, covering most of the existing architecture types involving feature upsampling.} {Fig.~\ref{fig:heads} illustrates the corresponding network structures of different backbones, which contain two different upsampling manners, \ie, iterative one and direct one. Please note that for other baselines, we implement them in an iterative upsampling manner by default for fair comparison. 
For our ReSFU, we specifically implement it in the more difficult direct upsampling manner to validate its capability in achieving the direct high-ratio upsampling. Actually, directly inserting other comparison methods in the direct upsampling manner would impair their performance. Please refer to Sec. \ref{sec:discuss} and SM for more discussion and implementation details.}

% {We provide the illustration of the experimented network architectures under iterative and direct upsampling manners in Fig. \ref{fig:heads}. In experiments, we implement the comparison upsampling methods in the iterative manner by default to avoid their performance decay and ensure fairness of comparison. In contrast, we implement ReSFU in the direct upsampling manner to validate its ability to accomplish direct high-ratio upsampling. Please refer to Sec. \ref{sec:discuss} and the supplementary material for more discussion and implementation details.}

{Following~\cite{lu2022sapa,liu2023learning}, the semantic segmentation experiments are evaluated using the single-scale testing on the ADE20K dataset \cite{zhou2017scene}, which is a widely adopted large-scale scene parsing dataset containing over 20,000 images with pixel-level annotations for 150 object categories.} Three typical metrics are used for quantitative evaluation, including mean IoU (mIoU), mean pixel accuracy (mAcc), and boundary IoU (bIoU) \cite{cheng2021boundary}. For the bIoU computation, the pixel distance is set to $2\%$ \cite{cheng2021boundary}. Experiments on more datasets are provided in SM.

\begin{table}[t] 
\footnotesize
\caption{Evaluation on ADE20K with PSPNet-ResNet50 and PSPNet-ResNet101. The best and second-best results are highlighted in bold and underlined, respectively.} 
\vspace{-2mm}
\label{tab:psp}
\centering
\renewcommand{\arraystretch}{1.14}
\scalebox{0.84}{\setlength{\tabcolsep}{1.4mm}{
\begin{tabular}{@{}l|cccc|cccc@{}}
\toprule
\multicolumn{1}{c|}{\multirow{2}{*}{Method}} & \multicolumn{4}{c|}{ PSPNet-ResNet50} & \multicolumn{4}{c}{PSPNet-ResNet101}   \\
  & mIoU & mAcc  & bIoU& Params  & mIoU & mAcc & bIoU & Params  \\ 
\midrule
 Bilinear {\textit{(default)}}  & 41.13  & 51.41  & 26.01 & 46.68M  & 43.57  & 54.51 & 28.28 & 65.67M \\
  Deconv   &  39.11 & 48.79 & 24.09& 50.87M & 41.74 & 51.53 & 26.65 & 69.87M  \\
  PixelShuffle~\cite{shi2016real}  & 38.31  & 47.78 & 23.77& 50.88M  &  41.39 & 50.95 &  26.64& 69.87M \\
   Stack-JBU \cite{fu2024featup}  & 41.68  & \underline{52.69}  & 25.78& 46.95M &  43.93 & 54.90 &  28.34 & 65.95M  \\
   CARAFE~\cite{wang2019carafe}  & \underline{42.06}  & 52.41  & {26.96}& 46.86M   & \underline{44.46}  & \underline{55.47} & 29.09 & 65.85M \\
  FADE~\cite{lu2022fade}  & 41.80 & 52.07  & 26.77  & 46.78M  & 43.84  & 53.63 & \underline{29.36} & 65.77M \\
  SAPA~\cite{lu2022sapa}  & 41.76 & 52.32  & \underline{26.99} & 46.72M  &  44.25 & 55.24 & 29.18 & 65.71M \\
  DySample~\cite{liu2023learning}  &  41.36 & 51.94 & 26.10 & 46.68M  & 43.99  & 54.63 &  28.33& 65.67M \\
 ReSFU  & \textbf{42.64}  & \textbf{53.26} & \textbf{27.89}& 46.73M  &  \textbf{45.19} & \textbf{56.29} & \textbf{29.87} & 65.72M \\ 
\bottomrule
\end{tabular}}}
\vspace{-3mm}
\end{table}

%41.80 & 52.07  & 26.77  & 44.53  & 54.37 & 29.88

\subsubsection{{Experimental Results}}
\noindent {\textbf{Evaluation on PSPNet with ResNet backbone.}} We first select the representative PSPNet~\cite{zhao2017pyramid} with the pioneering CNN-based backbone ResNet~\cite{he2016deep} for the semantic segmentation task. {Fig.~\ref{fig:heads} (a) presents the iterative and direct upsampling manners for PSPNet.} % presents the replacement manners of different upsampling methods. 

% As seen, for the feature $\bm{x}$ extracted in the last block, to achieve the $\times$4 upsampling for the final segmentation output, previous upsamplers adopt two successive $\times$2 upsampling processes by default. In contrast, our ReSFU is simply embedded with a direct $\times$4 upsampling. For guidance-based upsamplers, {\eg, FADE, SAPA, and ReSFU}, the guidance feature is from the stem convolution block.

\begin{figure*}[!h]
  \centering
  % \vspace{-0.2cm}
\includegraphics[width=0.99\linewidth]{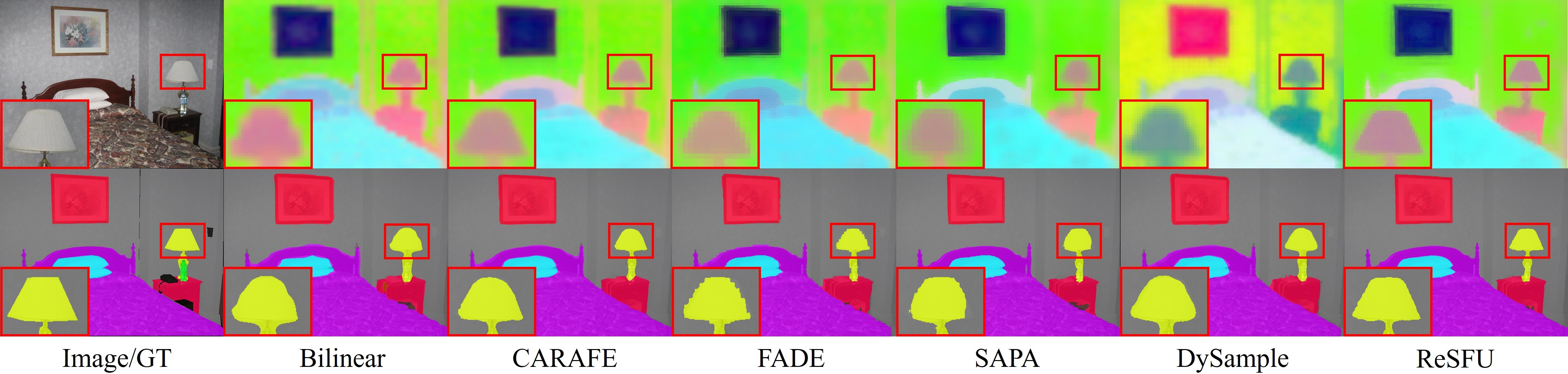}
  \vspace{-0.2cm}
  \caption{Visual comparison of the $\times 4$ upsampled features $\tilde{\bm{x}}$ (top) and predicted masks (bottom) with Segmenter-ViT-S.}
  \label{fig:vis_segmenter}
  \vspace{-0.3cm}
\end{figure*}

% \input{misc/fig_head_segmenter}
% \vspace{1mm}
% \noindent\textbf{Performance Comparison.}
Table~\ref{tab:psp} reports the quantitative results on ADE20K with two variants of the ResNet backbone, \ie, ResNet50 and ResNet101. It is clearly observed that with the comparable number of network parameters, our proposed ReSFU consistently outperforms other upsampling approaches in three metrics, \ie, mIoU/mAcc/bIoU (\%). Taking PSPNet-ResNet101 as an example, Fig. \ref{fig:vis_psp} presents the upsampled features $\widetilde{\bm{x}}$ and the corresponding segmentation results obtained with different upsamplers.\footnote{{For feature visualization in all the experiments, we follow \cite{fu2024featup} and adopt principal component analysis \cite{abdi2010principal} to reduce the features to three channels corresponding to the RGB values. Note that since the feature extractors are not frozen and are trained together with different upsamplers, there may be significant variations in color tone across different feature visualizations.}} From the feature visualizations, we can find that the object boundary upsampled by our ReSFU is much clearer, thus boosting more accurate segmentation results. In comparison with other approaches with multi-step $\times 2$ upsampling processes, even with a direct one-step high-ratio upsampling, our proposed ReSFU can still obtain superior performance, which facilitates the deployment in practical applications. 

\begin{table}[t] 
\footnotesize
\caption{Evaluation on ADE20K with Segmenter-ViT-S and Segmenter-ViT-L. The best and second-best results are highlighted in bold and underlined, respectively.} 
\vspace{-2mm}
\label{tab:segmenter}
\centering
\renewcommand{\arraystretch}{1.14}
\scalebox{0.84}{\setlength{\tabcolsep}{1.4mm}{
\begin{tabular}{@{}l|cccc|cccc@{}}
\toprule
\multicolumn{1}{c|}{\multirow{2}{*}{Method}} & \multicolumn{4}{c|}{Segmenter-ViT-S} & \multicolumn{4}{c}{Segmenter-ViT-L}   \\
 & mIoU & mAcc  & bIoU & Params & mIoU & mAcc & bIoU & Params   \\ 
\midrule
  Bilinear {\textit{(default)}}  & 45.75  & 56.88 & 27.82  & 22.04M  & 50.96  & 62.05 &  33.07& 304.17M\\
  Deconv   &  40.93 & 51.62 & 23.77& 24.40M &  50.01 & 61.14 & 31.75& 321.67M   \\
  PixelShuffle~\cite{shi2016real}   & 41.24 & 52.37 & 23.87 & 24.41M  & 50.50  & 61.49 & 31.45 & 321.69M  \\
   Stack-JBU \cite{fu2024featup} & 45.70  & 56.86  & 27.84 & 22.20M  & 51.03 & 62.39 & 32.90 & 305.96M \\
CARAFE~\cite{wang2019carafe}  & \underline{46.25} & \underline{57.63} & \underline{28.99}& 22.21M  & 51.85 & 62.96 & \underline{34.49} & 305.14M \\
    FADE~\cite{lu2022fade}  & 45.71 & 56.81 & 28.72 & 22.14M   & 50.07 & 61.79 & 33.32 & 305.08M \\
  SAPA~\cite{lu2022sapa}   & 45.79 & 57.36 & 28.97 & 22.09M   & 51.20 & 62.46 & 33.31 & 304.99M \\
  DySample~\cite{liu2023learning}    & 45.78 & 56.88 & 28.37& 22.09M  & \underline{52.04} & \underline{63.27} & 34.17 & 305.02M \\
 ReSFU   & \textbf{46.97}  & \textbf{57.91} &  \textbf{29.96} & 22.11M  & \textbf{52.56} & \textbf{63.63} &  \textbf{35.27} &  304.98M \\ 
\bottomrule
\end{tabular}}}
\vspace{-2mm}
\end{table}

\vspace{1mm}
\noindent {\textbf{Evaluation on Segmenter with ViT backbone.}} 
Here we introduce the representative ViT-based network, Segmenter \cite{strudel2021segmenter} with the FCN-head \cite{long2015fully}, and present the {iterative/direct upsampling manners as depicted in Fig.~\ref{fig:heads} (b). }

For comprehensiveness, we evaluate our ReSFU based on the ViT backbone with different sizes, \eg, ViT-S and ViT-L. As reported in Table \ref{tab:segmenter}, the proposed ReSFU evidently surpasses other methods, and achieves the best mIoU/mAcc/bIoU (\%) scores under different backbones with comparable number of parameters. The visual comparison is presented in Fig.~\ref{fig:vis_segmenter}, clearly showing that ReSFU achieves better feature upsampling effects with more accurate delineation of object boundaries, \eg, the desk lamp. These findings effectively validate the suitability of our ReSFU in the current popular Transformer-based architectures, demonstrating its strong potential for practical application.

\begin{figure*}[t]
  \centering
  % \vspace{-0.2cm}
\includegraphics[width=0.99\linewidth]{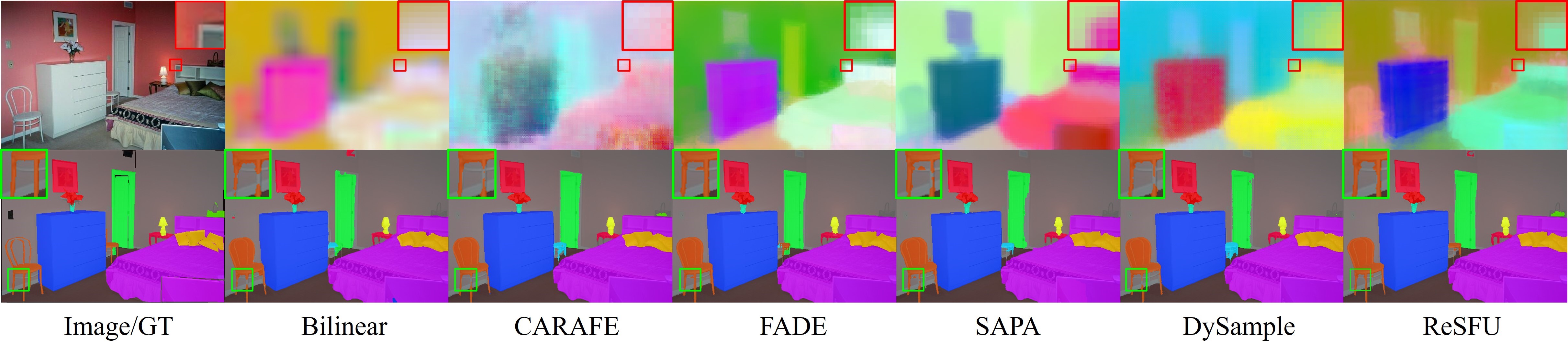}
  \vspace{-0.3cm}
  \caption{Visual comparison of the $\times 8$ upsampled features $\widetilde{C4}$ and predicted masks with SegFormer-MiT-B5.}
  \label{fig:vis_segformer}
  \vspace{-0.3cm}
\end{figure*}

% \hippo{The evaluation results are reported in Table \ref{tab:segmenter}. It can also be observed that our ReSFU outperforms all the comparison upsampling methods with few extra involved parameters. We further provide an exemplary visualization of upsampled feature maps and segmentation results.} 
% Figure~\ref{fig:vis_segmenter} presents the visual comparison of different upsampling methods based on Segmenter-S. It can be easily seen that our proposed ReSFU helps accomplish better feature upsampling effects with more accurate delineation of object boundaries, \eg, the desk lamp.

% \subsubsection{Evaluation on SegFormer with MiT}

% \noindent\textbf{Upsampling Details.} 
\vspace{1mm}
\noindent {\textbf{Evaluation on SegFormer with MiT.}} 
For the semantic segmentation task, SegFormer~\cite{xie2021segformer} is another mainstream Transformer-based network structure. Different from Segmenter with a non-hierarchical upsampling architecture, the upsampling process in SegFormer is composed of hierarchical forms among different levels of features{, as displayed in Fig.~\ref{fig:heads} (c).}

\begin{table}[t] 
\footnotesize
\caption{Evaluation on ADE20K with SegFormer-MiT-B1 and SegFormer-MiT-B5. The best and second-best results are highlighted in bold and underlined, respectively.} 
\vspace{-2mm}
\label{tab:segformer}
\centering
\renewcommand{\arraystretch}{1.14}
\scalebox{0.84}{\setlength{\tabcolsep}{1.4mm}{
\begin{tabular}{@{}l|cccc|cccc@{}}
\toprule
 \multicolumn{1}{c|}{\multirow{2}{*}{Method}} & \multicolumn{4}{c|}{ SegFormer-MiT-B1} & \multicolumn{4}{c}{SegFormer-MiT-B5}   \\
  & mIoU & mAcc  & bIoU & Params & mIoU & mAcc & bIoU & Params  \\ 
\midrule
 Bilinear  {\textit{(default)}}  & 40.97 & 51.73 & 24.89 & 13.72M & 49.13 & 60.69 & 32.32 & 82.01M  \\
  Deconv   & 40.73 & 51.25 & 23.78 & 15.29M  & 49.76 & 61.39 & 32.95 & 83.52M \\
  PixelShuffle~\cite{shi2016real} & 39.69 & 50.11 & 23.55   & 15.30M   & 50.09 & 61.69 &  33.11 & 83.59M \\
     Stack-JBU \cite{fu2024featup}  & 41.71 & 52.01  & 25.80 & 13.95M  & 49.53 & 60.72 & 32.89 & 82.24M \\ 
CARAFE~\cite{wang2019carafe}   & 42.75 & 53.96 & 27.09 & 14.16M  & \underline{50.45}  & \textbf{62.29}  & 33.78 & 82.45M \\
  IndexNet \cite{lu2019indices}  & 41.79 & 52.40 & 25.45&  26.32M  & 49.73 & 61.32 & 33.21 & 94.61M \\
  FADE~\cite{lu2022fade}  & 43.09 & \underline{53.89} & \textbf{28.33}  & 14.00M   &  {50.43} & 61.99 & \textbf{34.51} & 82.29M \\
  SAPA~\cite{lu2022sapa}   & 42.74 & 53.39 & {27.61} & 13.82M   &  49.61 &  61.11 & 34.02 & 82.11M \\
  DySample~\cite{liu2023learning}   & \underline{43.48} & 53.82 & 27.38 & 13.72M  &50.14  & 61.51 & 33.93 & 82.01M \\
   ReSFU   &\textbf{43.84}  & \textbf{54.75} & \underline{28.06}  & 13.86M  & \textbf{50.89} & \underline{62.00} &  \underline{34.40} & 82.15M  \\ 
\bottomrule
\end{tabular}}}
\vspace{-3mm}
\end{table}

\begin{figure*}[h]
  \centering
  % \vspace{-0.2cm}
\includegraphics[width=0.99\linewidth]{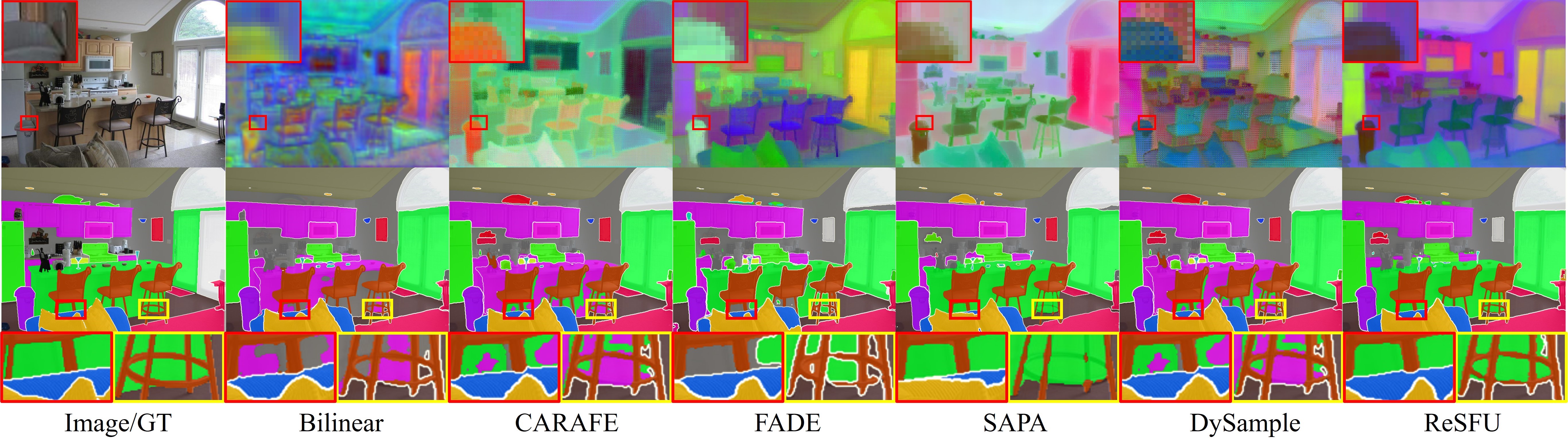}
  \vspace{-0.2cm}
  % \caption{Visual comparison of the upsampled features $P3$ (before the residual connection) and prediction masks on MaskFormer-Swin-L.}
    \caption{Visual comparison of the upsampled features $P2$ in {Fig.~\ref{fig:heads} (d)} and predicted masks with MaskFormer-Swin-L.}
  \label{fig:vis_maskformer}
  \vspace{-0.2cm}
\end{figure*}

% \input{misc/fig_head_fpn}
% \vspace{1mm}
% \noindent\textbf{Performance Comparison.}
Table \ref{tab:segformer} lists the quantitative results on two variants of the backbone MiT, \ie, MiT-B1 and MiT-B5, and the proposed ReSFU achieves better overall performance on all these metrics. Fig.~\ref{fig:vis_segformer} illustrates the $\times 8$ upsampled feature $\widetilde{C4}$ and the segmentation result obtained by different upsamplers. We can observe that for the comparing approaches with three consecutive $\times 2$ upsampling modules, \eg, CARAFE and DySample, the final upsampled features exhibit blurry effects. Nevertheless, even with a simpler $\times 8$ upsampling module, our ReSFU still accomplishes better detail preservation without mosaic artifacts. 
These favorable effects are mainly attributed to the careful designs of the controllable feature alignment and the fine-grained neighbor selection process, which enables ReSFU to be suitable for this direct high-ratio upsampling scenario. Thus, different from the current guidance-based feature upsampling methods, our ReSFU does not require features of adjacent levels as guidance, which naturally makes it possible to be extended to more network architectures not limited to hierarchical upsampling.

\begin{table}[t] 
\footnotesize
% \vspace{-0.2cm}
\caption{Evaluation on ADE20K with MaskFormer-Swin-B and MaskFormer-Swin-L. The best and second-best results are highlighted in bold and underlined, respectively.} 
\vspace{-2mm}
\label{tab:maskformer}
\centering
\renewcommand{\arraystretch}{1.14}
\scalebox{0.84}{\setlength{\tabcolsep}{1.4mm}{
\begin{tabular}{@{}l|cccc|cccc@{}}
\toprule
\multicolumn{1}{c|}{\multirow{2}{*}{Method}} & \multicolumn{4}{c|}{ MaskFormer-Swin-B} & \multicolumn{4}{c}{MaskFormer-Swin-L}   \\
  & mIoU & mAcc  & bIoU & Params & mIoU & mAcc & bIoU  & Params \\ 
\midrule 
Nearest  {\textit{(default)}}  & 52.56  & 65.76 &  37.41 & 101.83M  &  53.94 & 66.13  & 39.02 & 211.58M \\
Bilinear   & 52.99 & 66.33 &  37.19 & 101.83M  & 53.86 & 66.21 & 38.46 & 211.58M \\
Stack-JBU \cite{fu2024featup} & 53.53 &  66.57 & 37.53  & 101.91M  & 54.22  & \underline{67.26} &  38.44& 211.66M \\
CARAFE~\cite{wang2019carafe}  &  53.70  & 66.79 & 38.31 & 102.05M  & \underline{54.48}  & 67.12  &  39.06 & 211.80M \\
FADE~\cite{lu2022fade}  & 53.69  & 66.12 &  \underline{38.35}  & 101.97M & 54.47  & 66.59 & \underline{39.46}  & 211.72M \\
SAPA~\cite{lu2022sapa}   & \underline{53.87} & \underline{66.87} & 37.97 & 101.88M  &  54.41 &  66.40 & 39.20 & 211.63M \\
DySample~\cite{liu2023learning}   & 53.62  &  66.77 & 38.12 & 101.83M  & 54.17  & 67.13 & 39.22  & 211.58M \\
ReSFU   & \textbf{54.66}  & \textbf{68.25} & \textbf{38.69} & 101.97M & \textbf{55.33}  & \textbf{68.64} &  \textbf{39.58} & 211.72M  \\
\bottomrule
\end{tabular}}}
\vspace{-3mm}
\end{table}

\vspace{1mm}
\noindent{\textbf{Evaluation on MaskFormer with Swin-Transformer.}} 
 MaskFormer~\cite{cheng2021per} is also a cutting-edge Transformer-based network for the segmentation task. {As presented in Fig.~\ref{fig:heads} (d),} it contains a FPN neck-based upsampling module \cite{lin2017feature}, different from that adopted in Segmenter and SegFormer. 
 
First, we quantitatively evaluate all the comparing methods based on two versions of Swin Transformer \cite{liu2021swin} with varying sizes, including Swin-B and Swin-L. As observed from Table~\ref{tab:maskformer}, ReSFU consistently obtains state-of-the-art outcomes over all metrics. Then based on MaskFormer-Swin-L, we visualize the HR features upon upsampling $P2$. %\hippo{(after the third upsampling module, before the residual addition)}.
From the upper row in Fig.~\ref{fig:vis_maskformer}, it is clearly seen that {the comparing methods suffer from either blurriness or checkerboard artifacts \cite{sugawara2019checkerboard}.}
%the default bilinear upsampler exhibits blurry effects; CARAFE and DySample cause checkerboard artifacts \cite{sugawara2019checkerboard}; FADE suffers from mosaic effects; and SAPA leads to slight aliasing effects near object boundaries. 
In contrast, our ReSFU can generate clearer object boundaries and preserve more useful content, which in turn promotes better mask prediction as shown in the lower row.

% \hippo{The evaluation results of MaskFormer with Swin-B and Swin-L are reported in Table \ref{tab:maskformer}. The superior performance of ReSFU finely demonstrates its generality to strong segmentation backbones. We further provide a visual comparison based on MaskFormer-Swin-L in Fig. \ref{fig:vis_maskformer}. The upper row presents the visualizations of the upsampled feature before the residual connection for $P3$. It can be seen that the bilinear upsampler exhibits blurry effects; CARAFE and DySample have the checkerboard effects; FADE suffers from mosaic effects; SAPA also has some slight aliasing effects near object boundaries; and ReSFU can generate the clearest object boundary among these competing methods. The segmentation results displayed in the bottom row, with a particular focus on the zoomed-in region, clearly showcase the superior performance of ReSFU.}
\subsection{Model Analysis}
To better understand the working mechanism underlying the proposed ReSFU and validate the rationality of each methodological design, in this section, we conduct a comprehensive model analysis experiment, including model visualization and a series of ablation studies.

\subsubsection{Model Visualization} 
Based on the backbone Segmenter-ViT-S in {Fig.~\ref{fig:heads} (b)}, we first visualize each component learned by ReSFU.

\begin{figure*}[!h]
  \centering
  % \vspace{-0.3cm}
\includegraphics[width=0.99\linewidth]{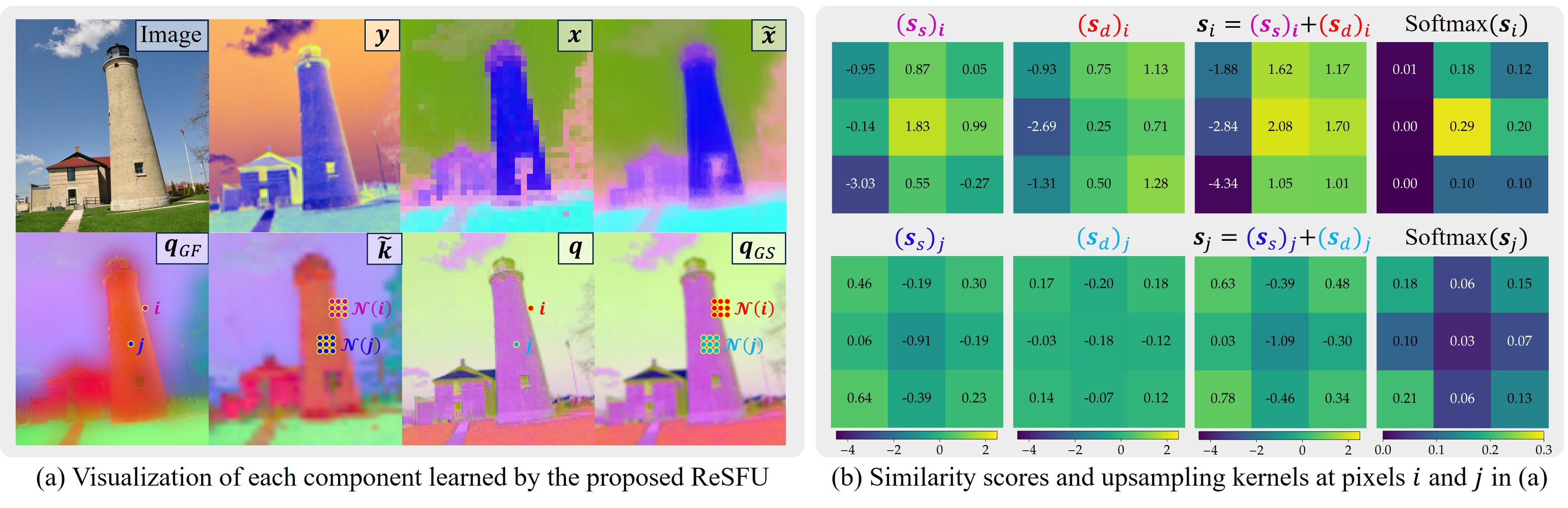}
  \vspace{-0.2cm}
    \caption{Visualization verification about the working mechanism underlying ReSFU based on Segmenter-ViT-S.}
    
  %including the \hazel{original image}, a shallow HR guidance feature $\bm{y}$, LR deep feature $\bm{x}$, the upsampled deep feature $\tilde{\bm{x}}$, the original query $\bm{q}$ extracted by the network backbone, blurred query $\bm{q}_{blur}$, bilinearly upsampled key $\tilde{\bm{k}}$, and the GF-based optimization query $\bm{q}_{GF}$.}
  \label{fig:vis_components}
  % \vspace{-0.4cm}
\end{figure*}

\begin{table*}[t] 
\footnotesize
% \vspace{-0.2cm}
\caption{Ablation study on the query-key feature alignment in Sec.~\ref{sec:align} with the similarity measurement $\text{sim}(\cdot, \cdot)$ implemented by the PCDC-Block in Sec.~\ref{sec:sim} and the neighbor selected via the FNS strategy in Sec.~\ref{sec:sns}.} 
\vspace{-2mm}
\label{tab:ab1}
\centering
\renewcommand{\arraystretch}{1.2}
\scalebox{1.02}{\setlength{\tabcolsep}{1.4mm}{
\begin{tabular}{@{}c|cccc|c|cccc|cccc@{}}
\toprule
 \multirow{2}{*}{Variant} &\multicolumn{4}{c|}{Feature Alignment} & \multirow{2}{*}{Similarity Score $\bm{s}$} & \multicolumn{4}{c|}{ Segmenter-ViT-S} & \multicolumn{4}{c}{SegFormer-MiT-B1}   \\
 & $\bm{s}_s$ & $\bm{q}_{GF}$ & $\bm{s}_d$ & $\bm{q}_{GS}$ & & mIoU & mAcc  & bIoU & Params  & mIoU & mAcc  & bIoU & Params  \\ 
\midrule
(a)  &  \grey{\xmark}  &  \grey{\xmark}&  \grey{\xmark}  & \grey{\xmark}  &Bilinear & 45.75  & 56.88 & 27.82 & 22.04M   & 40.97 & 51.73  & 24.89 & 13.72M  \\
(b)  &  \hazel{\cmark}  &  \grey{\xmark}  & \grey{\xmark} &  \grey{\xmark}  &$\bm{s} = \bm{s}_s=\text{sim}(\bm{q}, \tilde{\bm{k}})$ & 46.30 &  57.45 & 29.47 & 22.09M  & 43.26 &  54.07 & 27.76 & 13.81M  \\
% (c)  &  \hazel{\cmark}  & \hazel{\cmark} &  \grey{\xmark}  & \grey{\xmark} &$\bm{s} = \bm{s}_s=\text{sim}(\bm{q}_{GF}, \tilde{\bm{k}})$  &  46.76 & 58.26 & 29.67 & 22.09M  & 43.68 & 54.36 & 28.13 & 13.81M  \\
(c)  &  \hazel{\cmark}  & \hazel{\cmark} &  \grey{\xmark}  & \grey{\xmark} &$\bm{s} = \bm{s}_s=\text{sim}(\bm{q}_{GF}, \tilde{\bm{k}})$  &  46.51 & 57.89 & 29.66 & 22.09M  & 43.68 & 54.36 & {28.13} & 13.81M  \\
(d)  &  \grey{\xmark}  & \grey{\xmark}  & \hazel{\cmark} & \grey{\xmark} & $\bm{s} = \bm{s}_d=\text{sim}(\bm{q}, \bm{q})$  & 46.21 & 57.35 & 29.13 & 22.07M  & 42.91 & 53.60  & 27.43 & 13.79M  \\
(e) &  \grey{\xmark}  & \grey{\xmark}  & \hazel{\cmark} & \hazel{\cmark} & $\bm{s} = \bm{s}_d=\text{sim}(\bm{q}, \bm{q}_{{GS}})$  &  46.48  & 57.52 &  29.47 & 22.07M  & 43.02 & 53.83  & 27.57 & 13.79M  \\
(f)  &  \hazel{\cmark}  & \hazel{\cmark}  & \hazel{\cmark} & \hazel{\cmark} & $\bm{s} = \text{(c)}~ \bm{s}_s+ \text{(e)}~ \bm{s}_d$ & {46.97} & {57.91} & {29.96} & 22.11M   & {43.84} & {54.75} & {28.06} & 13.86M  \\
\bottomrule
\end{tabular}}}
\vspace{-2mm}
\end{table*}

% \begin{table*}[t] 
% \footnotesize
% % \vspace{-0.2cm}
% \caption{Ablation studies on similarity calculation $\text{sim}(\cdot, \cdot)$ where the feature alignment is fixed as variant (f) in Table~\ref{tab:ab1} and the neighbor is selected via the FNS strategy in Sec.~\ref{sec:sns}.} 
% \label{tab:ab2}
% \centering
% \renewcommand{\arraystretch}{1.2}
% \scalebox{1.02}{\setlength{\tabcolsep}{1.4mm}{
% \begin{tabular}{@{}c|c|ccccc|ccccc@{}}
% \toprule
%  \multirow{2}{*}{Variant} &\multirow{2}{*}{Similarity Calculation $\text{sim}(\cdot,\cdot)$} & \multicolumn{5}{c|}{ Segmenter-ViT-S} & \multicolumn{5}{c}{SegFormer-MiT-B1}   \\
%  &   & mIoU & mAcc  & bIoU & Params & FLOPs  & mIoU & mAcc  & bIoU & Params & FLOPs \\ 
% \midrule
% (a)  &  Inner Product in Eq.~\eqref{eqn:sapa} & 45.86  & 56.92 & 28.14 & 22.08M & 33.22G  & 42.35 & 53.06 & 26.97 & 13.77M & 16.50G \\
% % (b)  & Concat-Conv  & 46.59 &  57.85 & 29.69 & 22.12M & 34.15G  & 42.88 & 53.52 & 27.88 & 13.91M & 19.28G \\
% (b)  & PCDC in Fig.~\ref{fig:pcd} $\rightarrow$ Conv (concat($\bar{\bm{q}}$, $\bar{\bm{k}}$)) & 46.59 &  57.85 & 29.69 & 22.12M & 34.15G  & 42.88 & 53.52 & 27.88 & 13.91M & 19.28G \\
% (c)  &  PCDC-Block in Fig.~\ref{fig:pcd} &  46.97 & 57.91 & 29.96 & 22.11M & 33.88G   & 43.84 & 54.75 & 28.06 & 13.86M & 18.48G \\
% \bottomrule
% \end{tabular}}}
% % \vspace{-1mm}
% \end{table*}

\begin{table*}[t] 
\footnotesize
% \vspace{-0.2cm}
\caption{Ablation study on similarity measurement $\text{sim}(\cdot, \cdot)$.}   
% where the feature alignment is fixed as variant (f) in Table~\ref{tab:ab1} and the neighbor is selected via the FNS strategy in Sec.~\ref{sec:sns}
\vspace{-2mm}
\label{tab:ab2}
\centering
\renewcommand{\arraystretch}{1.2}
\scalebox{1.02}{\setlength{\tabcolsep}{1.4mm}{
\begin{tabular}{@{}c|c|cccc|cccc@{}}
\toprule
 \multirow{2}{*}{Variant} &\multirow{2}{*}{Similarity Measurement $\text{sim}(\cdot,\cdot)$} & \multicolumn{4}{c|}{ Segmenter-ViT-S} & \multicolumn{4}{c}{SegFormer-MiT-B1}   \\
 &   & mIoU & mAcc  & bIoU & Params   & mIoU & mAcc  & bIoU & Params  \\ 
\midrule
(a)  &  Inner Product in Eq.~\eqref{eqn:sapa} & 45.86  & 56.92 & 28.14 & 22.08M  & 42.35 & 53.06 & 26.97 & 13.77M  \\
% (b)  & Concat-Conv  & 46.59 &  57.85 & 29.69 & 22.12M & 34.15G  & 42.88 & 53.52 & 27.88 & 13.91M & 19.28G \\
(b)  & PCDC in Fig.~\ref{fig:pcd} $\rightarrow$ Conv (concat($\bar{\bm{q}}$, $\bar{\bm{k}}$)) & 46.59 &  57.85 & 29.69 & 22.12M  & 42.88 & 53.52 & 27.88 & 13.91M  \\
(c)  &  PCDC-Block in Fig.~\ref{fig:pcd} &  46.97 & 57.91 & 29.96 & 22.11M   & 43.84 & 54.75 & 28.06 & 13.86M  \\
\bottomrule
\end{tabular}}}
\vspace{-2mm}
\end{table*}

\vspace{1mm}
% \noindent \textbf{Visualization of intermediate results.}
\noindent\textbf{Visualization of Query-Key Features.}
Fig.~\ref{fig:vis_components} (a) presents each component, including original features $\bm{y}$ and $\bm{x}$, two aligned query-key pairs ($\bm{q}_{GF}$ and $\tilde{\bm{k}}$, $\bm{q}$ and $\bm{q}_{GS}$), and the upsampled feature $\tilde{\bm{x}}$. It is clearly observed that the GF-based optimization query $\bm{q}_{GF}$ semantically resembles $\tilde{\bm{k}}$ while retaining structural information in $\bm{q}$, \textit{e.g.}, the edges of the tower. The smoothed query $\bm{q}_{GS}$ effectively removes minor noises in the original query $\bm{q}$ within the semantically consistent regions, such as the lawn. Attributed to such rational alignment in both the detail and semantic spaces, the upsampled HR feature $\tilde{\bm{x}}$ maintains semantic fidelity to the original LR feature $\bm{x}$, while finely preserving distinguishable edges and removing mosaic artifacts. All these visualization results exhibit desirable effects and comply with our design motivations.

%\hippo{\sout{For every query-key pair,}}
\vspace{1mm}
\noindent \textbf{Visualization of Upsampling Kernels.}  Fig.~\ref{fig:vis_components} (b) further presents the PCDC-Block-based similarity scores $\bm{s}_s$ for the pair $\bm{q}_{GF}$ and $\tilde{\bm{k}}$, and $\bm{s}_d$ for the pair $\bm{q}$ and $\bm{q}_{GS}$, and the upsampling kernels $\text{Softmax}(\bm{s})$ at two representative locations, \ie, pixel $i$ near the object boundary and pixel $j$ in the smooth region. Concretely, 1) for pixel $i$ belonging to the class category of `sky', we can observe that $(\bm{s}_s)_{i}$ and $(\bm{s}_d)_{i}$ effectively collaborate to generate a rational upsampling kernel $\text{Softmax}(\bm{s}_{i})$ that assigns negligible weights to the left region representing the semantics of `tower'; {2) For pixel $j$, $(\bm{s}_s)_{j}$ and $(\bm{s}_d)_{j}$ are smoother than the scores for pixel $i$, complying with the smoothness of this region. Besides, we discover that the scores at four corners are slightly higher. This is mainly attributed to the flexible learning capability of PCDC-Block, which has the potential to adaptively detect the smoothness of the location and correspondingly produce more stable and smoother upsampling results by evenly assigning more weights to the far-away feature elements.}

% \subsection{Performance Evaluation}
% Table~\ref{tab:semantic} reports the quantitative results on the semantic segmentation task by employing different upsampling methods to five different network backbones. Specifically, 1) for the non-hierarchical architectures PSPNet and Segmenter, all the comparing methods are implemented in a direct HR upsampling manner as presented in Figs. \ref{fig:head} (a)(b). It is clearly observed that the proposed ReSFU significantly outperforms other approaches and consistently obtains higher segmentation accuracy on all the evaluation metrics; 2) For SegFormer and SegNeXt, the comparing methods are incorporated in the iterative $\times 2$ upsampling manner with an adjacent-level feature as guidance as described in Fig.~\ref{fig:first} (a). Although our proposed ReSFU directly and only adopts $C1$ as guidance as shown in Figs. \ref{fig:head} (c)(d), it can still achieve superior or at least comparable performance to others; 3) For UNet, all the comparing methods are used in the way as displayed in {Fig.~\ref{fig:head} (e)} and ReSFU achieves the best performance. These results comprehensively substantiate the superiority of our ReSFU, and demonstrate its high flexibility in selecting the guidance features and wide applicability to various backbones.

% Figure~\ref{fig:vis_seg} presents the visual comparison of different upsampling methods based on Segmenter-S. It can be easily seen that our proposed ReSFU helps accomplish better feature upsampling effects with more accurate delineation of object boundaries, \eg, the desk lamp.

\begin{figure*}[t]
  \centering
  % \vspace{-0.2cm}
\includegraphics[width=0.94\linewidth]{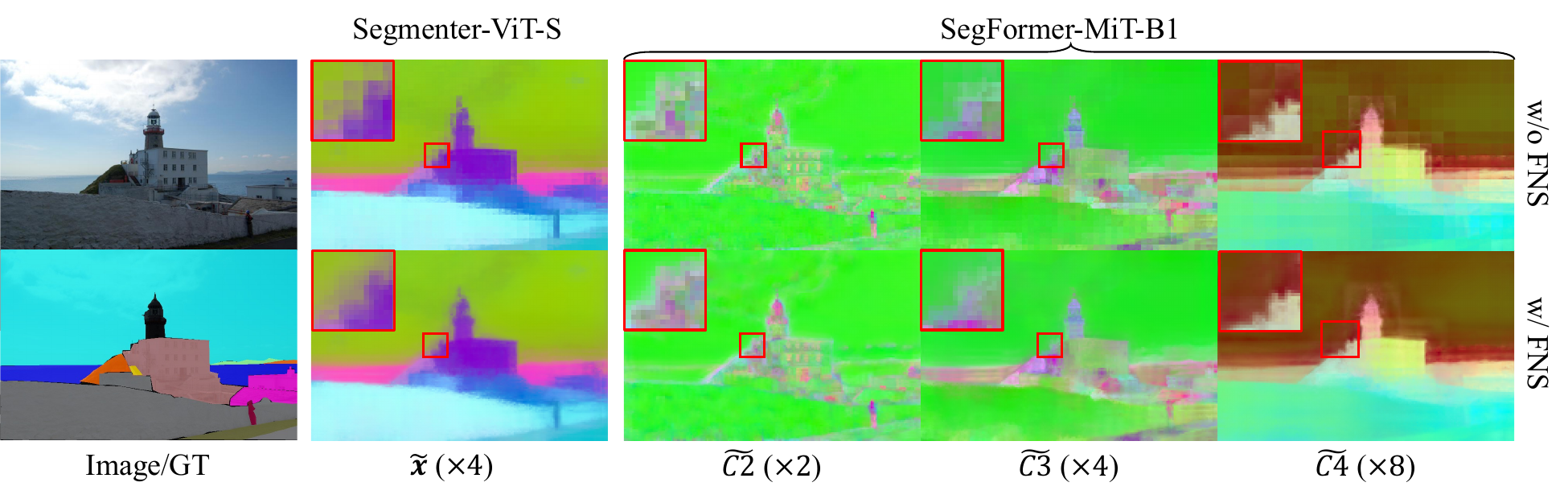}
  \vspace{-0.2cm}
  \caption{Effect of FNS on upsampled features extracted by Segmenter-ViT-S and SegFormer-MiT-B1. }
  \label{fig:ablation_fns}
  \vspace{-0.3cm}
\end{figure*}

\subsubsection{Ablation Study} \label{sec:ablation}
Based on the non-hierarchical Segmenter-ViT-S and the hierarchical SegFormer-MiT-B1, we conduct a series of ablation studies to fully verify the role of each design in ReSFU, {including feature alignment in the semantic space and the detail space, paired central difference convolution (PCDC) block for similarity calculation, and fine-grained neighbor selection (FNS) strategy.}

% , including the semantic-aware mutual-similarity $\bm{s}_s$, PCDC-Block for similarity computation, guided filter (GF) for query optimization, detail-aware self-similarity $\bm{s}_d$, and fine-grained neighbor selection (FNS). The results are listed in Table~\ref{tab:ablation}. More ablation studies are provided in \textit{SM}. 
 
\vspace{1mm}
\noindent \textbf{Ablation Study on Feature Alignment.}
Here we execute a fine-grained ablation study on the query-key feature alignment in Sec.~\ref{sec:align}, including $\bm{s}_{s}$ in Eq.~\eqref{eqn:semantic} and $\bm{s}_{d}$ in Eq.~\eqref{eqn:detail}, as well as the designs, \ie, guided filter-based optimized query $\bm{q}_{GF}$ and Gaussian smoothed detail-space key $\bm{q}_{GS}$. 

Table~\ref{tab:ab1} reports the quantitative results under different variants, where the variant (f) is exactly our proposed ReSFU. Specifically, compared with the bilinear upsampler (a), the performance gains achieved by variant (b) and variant (d) correspondingly validate that the introduction of the semantic-aware mutual similarity $\bm{s}_{s}$ and the detail-aware self-similarity $\bm{s}_{d}$ are beneficial for improving the segmentation performance, respectively. From the comparison between (b) and (c), it is easily known that the proposed guided filter-based feature alignment in the semantic space can indeed further boost $\bm{s}_{s}$ for more accurate segmentation. By comparing (d) with (e), we can find that the utilization of a smoothed key in the detail space can indeed bring slight performance improvement without any additional cost, which finely complies with the analysis in Sec.~\ref{sec:self}.

\vspace{1mm}
\noindent \textbf{Ablation Study on Similarity Measurement.}
To comprehensively validate the effectiveness of our proposed PCDC-Block, we compare three different similarity measurements $\text{sim}(\cdot, \cdot)$, including (a) traditional inner product in Eq.~\eqref{eqn:sapa}; (b) replacing the PCDC layer in Fig.~\ref{fig:pcd} with a vanilla convolution by inputting the concatenation of $\bar{\bm{q}}$ and $\bar{\bm{k}}$; (c) PCDC-Block in Fig.~\ref{fig:pcd}. From the results in Table \ref{tab:ab2}, it is observed that the proposed PCDC-Block has the better capability to capture the relations between query-key pairs for precise similarity calculation and thus promotes the segmentation accuracy. Considering the evident gains over (a), the slight computational cost brought by our method is acceptable. Please refer to SM for visual results.

% \hippo{To validate the effectiveness of our proposed PCDC-Block, we further conduct ablation studies on different similarity calculation methods. As shown in Table \ref{tab:ab2}, we experiment on three variants: (a) the traditional inner product; (b) concatenating the key and query features and then input to a convolutional block to get the output similarity results; (c) our proposed PCDC-Block. It should be noted that the only difference between variants (b) and (c) lies in the PCDC layer. The results reported in Table \ref{tab:ab2} have shown the superiority of our proposed PCDC-Block. }

% For similarity calculation, the superiority of our proposed PCDC-Block over the conventional inner product in Eq.~\eqref{eqn:sapa} can be easily observed by comparing variant (b) with (c). We further verify the role of the PCDC layer in the PCDC-Block and conduct an experiment on variant (f) which replaces the PCDC in Eq.~\eqref{eqn:pcd} with the conventional dilated convolution by taking the concatenation of the normalized query-key pair $\bar{\bm{q}}$ and $\bar{\bm{k}}$ as inputs. It is observed that compared with the plain dilated convolution, our PCDC achieves better segmentation results. 

\begin{table}[t] 
\footnotesize
% \vspace{-0.2cm}
\caption{Ablation study on the neighbor selection manner.} 
\vspace{-2mm}
\label{tab:ab3}
\centering
\renewcommand{\arraystretch}{1.2}
\scalebox{1.1}{\setlength{\tabcolsep}{1.4mm}{
\begin{tabular}{@{}c|ccc|ccc@{}}
\toprule
 Neighbor  & \multicolumn{3}{c|}{ Segmenter-ViT-S} & \multicolumn{3}{c}{SegFormer-MiT-B1}   \\
 Selection & mIoU & mAcc  & bIoU  & mIoU & mAcc  & bIoU  \\ 
\midrule
w/o FNS &  46.80 & 57.78 & 29.70 & 43.12 & 53.70  & 27.24 \\
w/ FNS  & 46.97 & 57.91 & 29.96  & 43.84 & 54.75 & 28.06 \\
\bottomrule
\end{tabular}}}
\vspace{-1mm}
\end{table}

\begin{table}[t] 
\footnotesize
% \vspace{-0.2cm}
\caption{Ablation study about hyperparameters based on Segmenter-ViT-S. `$\dagger$' denotes the default setting for ReSFU.} 
\vspace{-2mm}
\label{tab:ab_hyper}
\centering
\renewcommand{\arraystretch}{1.2}
\scalebox{1}{\setlength{\tabcolsep}{1.4mm}{
\begin{tabular}{@{}cc|ccccc@{}}
\toprule
 \multicolumn{2}{c|}{Hyperparameter} & mIoU & mAcc  & bIoU & Params & FLOPs \\  \midrule 
 % K=2 & 46.22 & 57.20 & 29.13 & 22.11M & 33.86G \\
  \multirow{3}{*}{\makecell{Kernel\\Size}} &  $\dagger$ K=3 & 46.97 & 57.91 & 29.96 & 22.11M & 33.88G \\
 & {\color{white}$\dagger$} K=5 & 46.99 & 57.90 & 30.39 & 22.14M & 34.49G \\ 
  & {\color{white}$\dagger$} K=7 & 47.02 & 58.37 & 30.41 & 22.20M & 35.39G \\ \midrule
 \multirow{3}{*}{\makecell{Projection\\Dimension}} & {\color{white}$\dagger$} D=16 & 46.58 & 58.09 & 29.88 & 22.09M & 33.63G \\
 & $\dagger$ D=32 & 46.97 & 57.91 & 29.96 & 22.11M & 33.88G \\ 
 & {\color{white}$\dagger$} D=64 & 47.00 & 58.35 & 30.18 & 22.14M & 34.39G \\ \midrule
  \multirow{4}{*}{Group} & {\color{white}$\dagger$} G=1 & 47.06 & 57.97 & 30.36 & 22.19M & 35.29G \\
 & {\color{white}$\dagger$} G=2 & 47.01 & 58.05 & 30.22 & 22.13M & 34.35G \\
 & $\dagger$ G=4  & 46.97 & 57.91 & 29.96 & 22.11M & 33.88G \\ 
 & {\color{white}$\dagger$} G=8 & 46.71 & 57.97 & 29.94 & 22.09M & 33.65G \\ 
\bottomrule
\end{tabular}}}
\vspace{-2mm}
\end{table}

\vspace{1mm}
\noindent \textbf{Ablation Study on Neighbor Selection.} 
Table~\ref{tab:ab3} reports the effect of FNS on the segmentation performance achieved by our ReSFU. As seen, without adopting FNS, the model would suffer from a large performance drop, especially for SegFormer with a higher-ratio $\times 8$ upsampling process as presented in {Fig.~\ref{fig:heads} (c) (lower)}. For better clarity, we provide visual comparisons of upsampled features extracted by these two different network structures. From Fig.~\ref{fig:ablation_fns}, we can clearly find that for different levels of features, \ie, $\bm{x}$ in Segmenter,  $C2$, $C3$, and $C4$ in SegFormer, their upsampled feature maps without FNS always suffer from severe mosaic effects, especially for larger upsampling ratios.
Fortunately, this unfavorable phenomenon can be effectively eliminated by adopting our proposed FNS strategy. It is worth mentioning that as explained in Sec.~\ref{sec:sns}, FNS is directly implemented based on bilinearly-upsampled features, and it introduces no extra parameters and computational costs.

\begin{table}[!t]
%\small
\footnotesize
% \vspace{-0.3cm}
\caption{{Comparisons on the number of parameters, FLOPs, inference time, and training time per iteration, tested on Segmenter-ViT-S with a 512$\times$512 image on a V100 GPU.}}
\label{tab:flop}
\vspace{-2mm}
\centering
\renewcommand{\arraystretch}{1.1}
\setlength{\tabcolsep}{0.9mm}{
{
\begin{tabular}{@{}l|c|cccc@{}} 
\toprule 
  \multicolumn{1}{c|}{Method}&  Type & Params & FLOPs  & Inference & Training \\  \midrule
 Bilinear & \multicolumn{1}{c|}{\multirow{5}{*}{w/o  guidance}}  & 22.04M & 32.77G  & 0.012s & 0.060s \\
   Deconv& & 24.40M & 35.16G  & 0.014s & 0.062s \\
   PixelShuffle~\cite{shi2016real} & & 24.41M & 35.16G & 0.013s & 0.061s \\
   CARAFE~\cite{wang2019carafe} & & 22.21M & 32.96G & 0.013s & 0.061s \\
    DySample~\cite{liu2023learning} & & 22.09M & 32.90G & 0.013s & 0.061s \\ \midrule
   Stack-JBU~\cite{fu2024featup} & \multicolumn{1}{c|}{\multirow{4}{*}{w/ guidance}}  & 22.20M & 35.32G & 0.019s & 0.080s \\
  FADE~\cite{lu2022fade} & & 22.14M & 33.92G & 0.034s & 0.133s \\
  SAPA~\cite{lu2022sapa} & & 22.09M & 33.28G & 0.013s & 0.062s \\
 ReSFU & &  22.11M & 33.88G  & 0.015s & 0.070s \\
\bottomrule
\end{tabular} } }
\vspace{-1mm}
\end{table}

% \begin{table}[!t]
% %\small
% \footnotesize
% % \vspace{-0.3cm}
% \caption{Comparisons on FLOPS and the number of parameters, where FLOPs are tested with images of size 512$\times$512. }
% \label{tab:flop}
% \vspace{-2mm}
% \centering
% \renewcommand{\arraystretch}{1.1}
% \setlength{\tabcolsep}{1.3mm}{
% \begin{tabular}{@{}lc|cccc@{}}
% \toprule
% \multicolumn{1}{c}{Method} & Type & FLOPs & Params & Inference & Training \\ \midrule
% Bilinear & \multicolumn{1}{c|}{\multirow{5}{*}{w/o guidance}} & 32.77G & 22.04M & 0.012s & 0.060s \\
% Deconv & & 35.16G & 24.40M & 0.014s & 0.062s \\
% PixelShuffle~\cite{shi2016real} & & 35.16G & 24.41M & 0.013s & 0.061s \\
% CARAFE~\cite{wang2019carafe} & & 32.96G & 22.21M & 0.013s & 0.061s \\
% DySample~\cite{liu2023learning} & & 32.90G & 22.09M & 0.013s & 0.061s \\ \midrule
% Stack-JBU~\cite{fu2024featup} & \multicolumn{1}{c|}{\multirow{4}{*}{w/ guidance}} & 35.32G & 22.20M & 0.019s & 0.080s \\
% FADE~\cite{lu2022fade} & & 33.92G & 22.14M & 0.034s & 0.133s \\
% SAPA~\cite{lu2022sapa} & & 33.28G & 22.09M & 0.013s & 0.062s \\
% ReSFU & & 33.88G & 22.11M & 0.015s & 0.070s \\
% \bottomrule
% \end{tabular} }
% \vspace{-1mm}
% \end{table}
\begin{figure}[t]
  \centering
  \vspace{-0.2cm}
\includegraphics[width=0.99\linewidth]{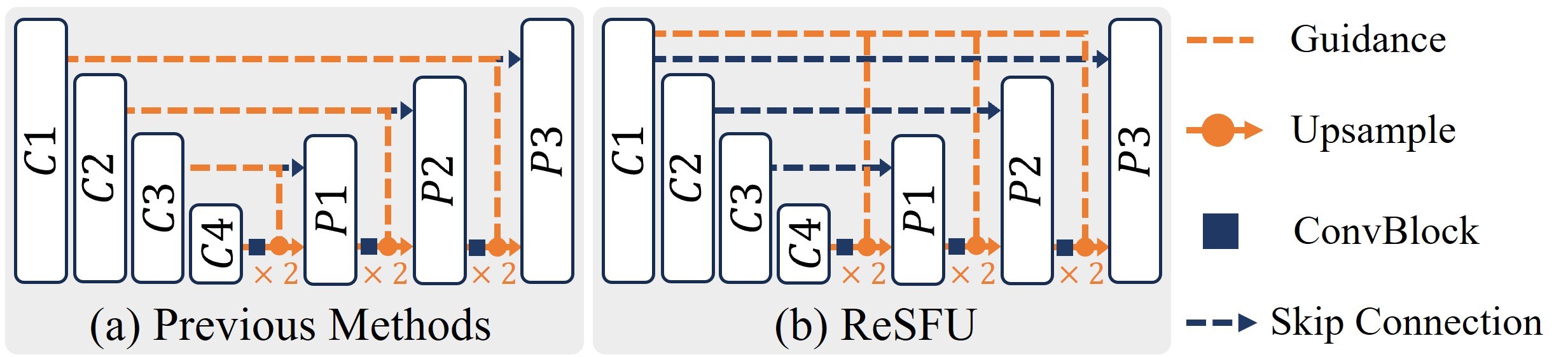}
  \vspace{-0.2cm}
  \caption{For the U-shape TransUNet, (a) previous methods sequentially use $C3$, $C2$, and $C1$ as guidance; (c) ReSFU only uses $C1$ as guidance for upsampling at different levels. }%\hippo{The skip connection is implemented by the concatenation along the channel dimension.}}
  % \caption{}
  \label{fig:head_unet}
  \vspace{-2mm}
\end{figure}

\begin{figure*}[t]
  \centering
  % \vspace{-0.2cm}
\includegraphics[width=0.99\linewidth]{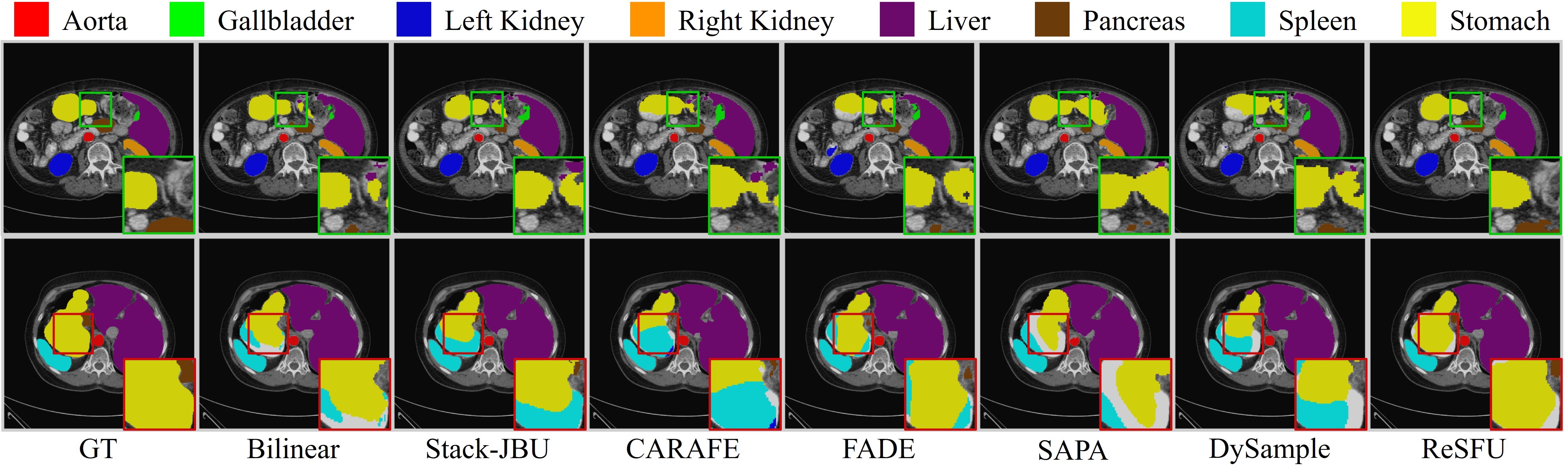}
  \vspace{-0.2cm}
  \caption{Visual comparison of different feature upsampling methods on the Synapse dataset based on TransUNet.}
  \label{fig:synapse}
  %\vspace{-0.1cm}
\end{figure*}

\begin{table*}[t] 
\footnotesize
% \small
% \vspace{-0.2cm}
% \caption{Evaluation on the Synapse dataset with TransUNet. The best and second-best results are highlighted in bold and underlined, respectively.} 
\caption{Evaluation on the Synapse dataset annotated with eight
abdominal organs based on TransUNet. The best and second-best results are highlighted in bold and underlined, respectively.} 
\vspace{-2mm}
\label{tab:synapse}
\centering
\renewcommand{\arraystretch}{1.14}
\setlength{\tabcolsep}{1.4mm}{
\begin{tabular}{@{}l|cccccccc|cc@{}}
\toprule
% \multicolumn{1}{c|}{Method}  & Aorta & Gallbladder & Kidney (L) &  Kidney (R) &  Liver & Pancreas & Spleen & Stomach & Avg. Dice (\textuparrow) & Avg. HD (\textdownarrow)  \\
% \multicolumn{1}{c|}{\multirow{2}{*}{Method}} & \multicolumn{3}{c|}{ Dice (\%)  (\textuparrow)} & \multicolumn{2}{c}{ Average} \\
\multicolumn{1}{c|}{\multirow{2}{*}{Method}} & \multicolumn{8}{c|}{ Dice (\%)  (\textuparrow)} & \multicolumn{2}{c}{ Average} \\
 &Aorta & Gallbladder & Kidney (L) &  Kidney (R) &  Liver & Pancreas & Spleen & Stomach& Dice (\%) (\textuparrow) & HD95 (\textdownarrow) \\
\midrule
Bilinear {\textit{(default)}}   & \underline{87.23} & \underline{63.13} & {81.87} & {77.02} & 94.08 & \underline{55.86} & \underline{85.08} & \underline{75.62} & \underline{77.48} & 31.69 \\
% Deconv  & & & & & & & &  &    \\
% PixelShuffle~\cite{shi2016real}  & & & & & & & &  &    \\
Stack-JBU \cite{fu2024featup}  & 87.17 & 59.13 & \textbf{85.03} & 78.24 & 94.20 & 54.73 & 83.85 & 76.36 & 77.34 & 30.88  \\
CARAFE~\cite{wang2019carafe}  &  86.34 & 62.21 & 83.32 & \textbf{79.31} & 93.69 & 52.03 & 82.06 & 73.96 & 76.61 & 30.89  \\
FADE~\cite{lu2022fade} & 86.22 & 62.56 & 83.44 & 77.67 & 93.41 & 53.84 & 83.39 & 73.65 & 76.77 & 30.55  \\
SAPA~\cite{lu2022sapa} & 84.22 & 56.91 & 75.30 & 71.79 & 91.41 & 47.92 & 81.50 & 67.25 & 72.03 & 35.46 \\
DySample~\cite{liu2023learning}  & \textbf{87.47} & 61.50 & 83.00 & 77.28 & \textbf{94.63} & 53.63 & \textbf{85.23 }& 74.20 & 77.12 & \underline{28.58}  \\
ReSFU & 86.93 & \textbf{65.30} & \underline{83.66} & \underline{78.42} & \underline{94.61} & \textbf{60.98} & 84.95 & \textbf{77.31} & \textbf{79.02} & \textbf{25.64} \\
\bottomrule
\end{tabular}}
\vspace{-4mm}
\end{table*}

% \begin{table}[t] 
% \footnotesize
% \caption{Evaluation on the ACDC dataset with TransUNet. The best and second-best results are highlighted in bold and underlined, respectively.} 
% \label{tab:acdc}
% \centering
% \renewcommand{\arraystretch}{1.14}
% \setlength{\tabcolsep}{1.75mm}{
% \begin{tabular}{@{}l|c|ccc|c@{}}
% \toprule
% \multicolumn{1}{c|}{Method}&Params & RV & Myo & LV & Average \\
% \midrule
% Bilinear  & 105.28M& \textbf{88.86} & \underline{84.53} & 95.73  & \underline{89.71}  \\
% Stack-JBU \cite{fu2024featup}& 105.33M & 88.67 & 82.83 & \underline{95.69} & 89.06 \\
% CARAFE~\cite{wang2019carafe} & 105.51M & 87.57 & {83.16}  & 95.38  & 88.70  \\
% FADE~\cite{lu2022fade}& 105.40M & 88.28 & 84.43  & 95.96 & 89.56 \\
% SAPA-B~\cite{lu2022sapa} & 105.34M & 87.37 & 84.01 & 95.00 & 88.79  \\
% DySample-S+~\cite{liu2023learning}& 105.28M & \underline{88.85} & 82.96 & 95.37 & {89.06} \\
% ReSFU & 105.40M & 88.70 & \textbf{85.16} & \textbf{96.26} & \textbf{90.04} \\
% \bottomrule
% \end{tabular}}
% \end{table}

\begin{table}[t] 
\footnotesize
\caption{Evaluation on the ACDC dataset with TransUNet. The best and second-best results are highlighted in bold and underlined, respectively.}
\vspace{-2mm}
%The best and second-best results are highlighted in bold and underlined, respectively.} 
\label{tab:acdc}
\centering
\renewcommand{\arraystretch}{1.14}
\setlength{\tabcolsep}{1.5mm}{
\begin{tabular}{@{}l|ccc|cc@{}}
\toprule
% \multicolumn{1}{c|}{Method} & RV & Myo & LV & Avg. Dice (\%) (\textuparrow) & Avg. HD (\textdownarrow) \\
\multicolumn{1}{c|}{\multirow{2}{*}{Method}} & \multicolumn{3}{c|}{ Dice (\%)  (\textuparrow)} & \multicolumn{2}{c}{ Average} \\
 &RV & Myo & LV & Dice (\%) (\textuparrow) & HD95 (\textdownarrow) \\
\midrule
Bilinear {\textit{(default)}}   &  \textbf{88.86} & \underline{84.53} & 95.73  & \underline{89.71} & 2.18 \\
Stack-JBU \cite{fu2024featup} & 88.67 & 82.83 & \underline{95.69} & 89.06 & 2.09 \\
CARAFE~\cite{wang2019carafe}& 87.57 & {83.16}  & 95.38  & 88.70 & 2.26 \\
FADE~\cite{lu2022fade}& 88.28 & 84.43  & 95.96 & 89.56 & 2.24 \\
SAPA~\cite{lu2022sapa}  & 87.37 & 84.01 & 95.00 & 88.79 & \underline{1.90} \\
DySample~\cite{liu2023learning}& \underline{88.85} & 82.96 & 95.37 & {89.06} & 2.75 \\
ReSFU & 88.70 & \textbf{85.16} & \textbf{96.26} & \textbf{90.04} & \textbf{1.82} \\
\bottomrule
\end{tabular}}
\vspace{-2mm}
\end{table}

\vspace{1mm}
\noindent \textbf{Ablation Study on Hyperparameters.}
Besides, based on Segmenter-ViT-S, we analyze the influence of the key hyperparameters on the segementation performance, including the kernel size $K$ for neighbor selection, the projection dimension $D$ for generating query-key features, and the number of group $G$ in Eq.~\eqref{eqn:pcd2} for similarity measurement.

The results are reported in Table~\ref{tab:ab_hyper}. As observed, a larger kernel size $K$ means that more neighboring pixels would be selected for the more accurate similarity measurement, thereby boosting the performance. However, it would decrease the computation efficiency and incur additional parameters. We select the default kernel size $K=3$ to balance the performance and cost. 
For the query-key features, we find that a larger $D$ would generate better feature representation, and then benefit the information propagation for the final segmentation. We choose $D$ as 32 by default. 
The smaller $G$ is, the better it is to capture the relations between query-key pairs, but it brings more computational overhead. In experiments, we set the default group number $G=4$. 

% Under the default configurations with $K=3$, $D=32$, and $G=4$, the overhead of our ReSFU is comparable or even lower than other baselines (see Table~\ref{tab:flop}). Taking into account performance, universality, and ease of deployment, our method has a better potential for real applications.
\vspace{1mm}
\noindent {\textbf{Analysis about Model Complexity.}}
{
Table \ref{tab:flop} reports the model complexities of different comparison methods, including the number of parameters, FLOPs, average inference time and training time per iteration based on Segmenter-ViT-S. We can find that 1) Compared to the guidance-free methods, due to the introduction of the HR feature for extra computation, the guidance-based upsampling methods generally cause a certain increase in model complexity and inference/training time; 2) Among the guidance-based feature upsampling methods, our ReSFU still demonstrates stronger competitiveness, which is comparable to SAPA and obviously outperforms other methods, including Stack-JBU and FADE. As seen, our ReSFU performs very competitively in terms of both effectiveness and efficiency. More results are provided in SM.}

\subsection{Application to More Tasks}
In this section, we extend ReSFU to more application scenarios including medical image segmentation, instance segmentation, and panoptic segmentation. 
% \hippo{To further demonstrate the generality of ReSFU, we additionally conduct experiments on medical datasets for the multi-organ segmentation task, and finally we explore the application of ReSFU to instance segmentation and panoptic segmentation tasks.}
% \noindent \textbf{Instance Segmentation and Panoptic Segmentation.}

\subsubsection{Medical Image Segmentation}
\noindent\textbf{Upsampling Details.}
For the medical image segmentation task, we choose the prevalent network, TransUNet~\cite{chen2021transunet}, with a U-shape architecture different from those presented in Sec. \ref{sec:sem_seg}. As illustrated in Fig.~\ref{fig:head_unet}, this network contains three $\times 2$ upsampling procedures. For the guidance, other upsamplers sequentially adopt $C3$, $C2$, and $C1$, which is a more suitable setting for them, while our ReSFU only uses the shallow $C1$ for guiding the upsampling at different levels of features. %{Tool?}

% \hippo{To further demonstrate the generality of ReSFU across different types of datasets, we additionally conduct experiments on the medical image segmentation task. Here, we adopt the popular network for medical image segmentation, TransUNet \cite{chen2021transunet}, which is also a different architecture from those in Sec. \ref{sec:sem_seg} to incorporate upsampling methods. The schematic illustration of the incorporation manner of the upsampling methods is shown in Fig. \ref{fig:head_unet}. There are three involved $\times 2$ upsampling processes in the decoder. Our ReSFU regards the shallow encoder feature $C1$ as the guidance feature for the three upsampling processes, while previous methods (which require guidance) achieve relatively better performance when using $C3$, $C2$, and $C1$ sequentially as the guidance features.}

\begin{figure*}[!t]
  \centering
  % \vspace{-0.7cm}
  % \vspace{-2mm}
\includegraphics[width=0.99\linewidth]{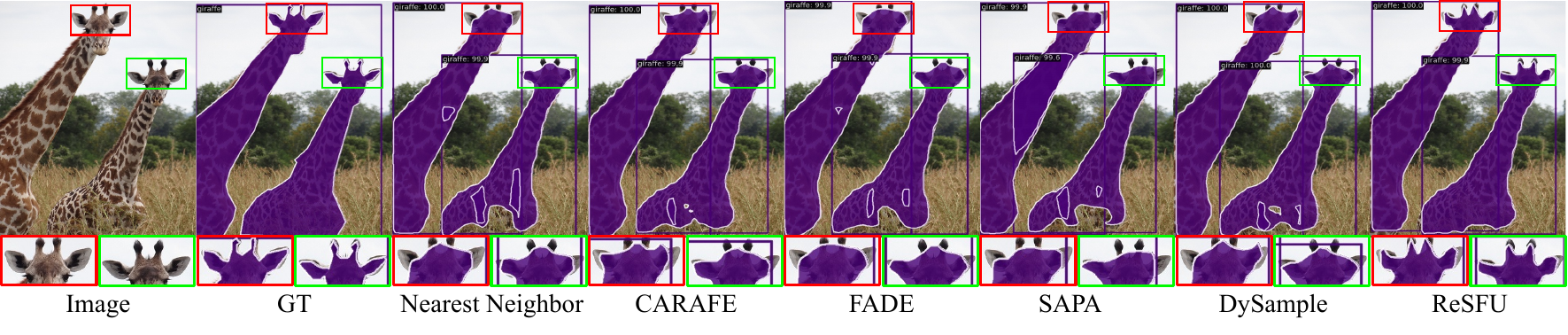}
  \vspace{-0.2cm}
  \caption{Visualization results of different upsampling methods based on Mask R-CNN for instance segmentation.}
  \label{fig:vis_ins}
  \vspace{-0.1cm}
\end{figure*}

\begin{figure*}[!t]
  \centering
  % \vspace{-0.7cm}
  % \vspace{-2mm}
\includegraphics[width=0.99\linewidth]{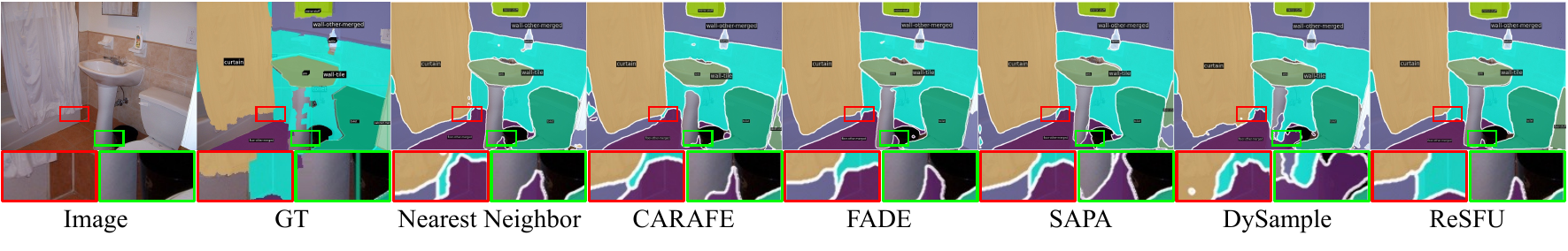}
  \vspace{-0.2cm}
  \caption{Visualization results of different upsampling methods based on Panoptic FPN for panoptic segmentation.}
  \label{fig:vis_panoptic}
  % \vspace{-0.1cm}
\end{figure*}

\begin{table*}[t] 
% \footnotesize
\small
% \vspace{-0.2cm}
\caption{Performance evaluation on two additional segmentation tasks, including instance segmentation and panoptic segmentation. The best and second-best results are highlighted in bold and underlined, respectively. }%\vspace{-3mm} ``NA'' in Stack-JBU for instance segmentation indicates that it suffers from unstable performance leading to infinite loss value during training.
\vspace{-2mm}
\label{tab:instance}
\centering
\renewcommand{\arraystretch}{1.14}
\setlength{\tabcolsep}{1.8mm}{
\begin{tabular}{@{}l|cccccc|ccccc@{}}
\toprule
\multicolumn{1}{c|}{\multirow{2}{*}{Method}} & \multicolumn{6}{c|}{Instance Segmentation} & \multicolumn{5}{c}{Panoptic Segmentation} \\
 & AP & AP$_{50}$ & AP$_{75}$ & AP$_{S}$ & AP$_{M}$ & AP$_{L}$ & PQ & PQ$^{th}$ & PQ$^{st}$ & SQ & RQ \\  \midrule
Nearest {\textit{(default)}}  & 34.7 & 55.8 & 37.2 & 16.1 & 37.3 & 50.8 & 40.2 & 47.8 & 28.9 & 77.8 & 49.3 \\
Bilinear  & 34.4 & 55.4 & 36.7 & 15.6 & 36.9 & 51.2 & 40.1 & 47.5 & 28.9 & 77.8 & 49.1 \\
Deconv & 34.5  & 55.5  & 36.8  & 16.4  & 37.0  & 49.5  & 39.6  & 47.0  & 28.4  & 77.1  & 48.5 \\
PixelShuffle~\cite{shi2016real} & 34.8  & 56.0  & 37.3  & 16.3  & 37.5  & 50.4  & 40.0  & 47.4  & 28.8  & 77.1  & 49.1 \\
% Stack-JBU \cite{fu2024featup} & NA & NA  & NA & NA  & NA & NA  & 40.4  & 47.7  & 29.3  & 77.8  & 49.4 \\
CARAFE~\cite{wang2019carafe} &  35.4 &  56.7 & 37.6 & \underline{16.9}  & \underline{38.1}  & 51.3  & 40.8  & 47.7  & 30.4  & 78.2  & 50.0 \\
IndexNet \cite{lu2019indices} & 34.7 & 55.9  & 37.1 & 16.0 & 37.0  & 51.1 & 40.2 &  47.6 & 28.9 &  77.1 & 49.3 \\
% A2U \cite{dai2021learning} & 34.6 & 56.0 & 36.8  &  16.1 &  37.4 & 50.3  &  40.1 & 47.6  & 28.7  &  77.3 & 48.0 \\
FADE \cite{lu2022fade} & 35.1 & 56.7 &  37.2 &  16.7 &  37.5 & 51.4  &  {40.9} &  48.0 & 30.3  & 78.1  & {50.1} \\
SAPA \cite{lu2022sapa} & 35.1 & 56.5  & 37.4  & 16.7  & 37.6  & 50.6  & 40.6  & 47.7  & 29.8  & 78.0  & 49.6 \\
% DySample+ \cite{liu2023learning} & \underline{35.7} & \underline{57.3}  &  \underline{38.2} &  \textbf{17.3} & \underline{38.2}  & \underline{51.8}  & \textbf{41.5}  & \textbf{48.5}  & \underline{30.8}  & \underline{78.3}  & \textbf{50.7} \\
DySample \cite{liu2023learning} & \underline{35.5} & \underline{56.8} & \underline{37.8}  & \textbf{17.0}  & 37.9  & \underline{51.9}  & \underline{41.1}  & \underline{48.1}  & \underline{30.5}  &  \underline{78.2} & \underline{50.2} \\
ReSFU &  \textbf{36.1} & \textbf{57.4} &  \textbf{38.4} & 16.8 & \textbf{38.8}  & \textbf{52.5}  & \textbf{41.5}  & \textbf{48.4}   & \textbf{31.1}  & \textbf{78.4}  & \textbf{50.7} \\
\bottomrule
\end{tabular}}
\vspace{-4mm}
\end{table*}

\vspace{1mm}
\noindent\textbf{Performance Comparison.}
The experiments are executed based on two widely-adopted datasets, \ie, Synapse for multi-organ segmentation,\footnote{\url{https://www.synapse.org/\#!Synapse:syn3193805/wiki/217789}} and the automated cardiac diagnosis challenge (ACDC).\footnote{\url{https://www.creatis.insa-lyon.fr/Challenge/acdc/}} 
{Specifically, Synapse consists of computed tomography (CT) scans annotated with eight abdominal organs, and ACDC includes magnetic resonance imaging (MRI) scans with three types of annotations. We follow~\cite{chen2021transunet} for constructing the training and evaluation datasets.}
% Specifically, Synapse is annotated with eight abdominal organs, \ie, aorta, gallbladder, spleen, left kidney, right kidney, liver, pancreas, spleen, and stomach. Following~\cite{chen2021transunet}, we employ 18 computed tomography (CT) scans with 2,212 axial slices for training and 12 CT scans for validation. For ACDC, the involved magnetic resonance imaging (MRI) scans are annotated with left ventricle (LV), right ventricle (RV), and myocardium (MYO). Consistent to~\cite{chen2021transunet}, we utilize 70 scans with 1,930 axial slices for training and 40 scans for validation.
Quantitative evaluation is conducted based on Dice (the higher the better) and the 95-th percentile of the Hausdorff Distance (HD95, the lower the better). %\hippo{The Dice score of each organ is reported as well as the average Dice and the average HD scores.}

From the results reported in Table~\ref{tab:synapse} for Synapse and Table~\ref{tab:acdc} for ACDC, we can easily conclude that most of the comparing baselines are inferior to the default bilinear upsampler. However, our proposed ReSFU consistently achieves competitive segmentation performance on different organs, and then obtains the best average accuracy with higher Dice scores and lower HD scores. In Fig. \ref{fig:synapse}, we present the segmentation results on two cases randomly selected from Synapse. As seen, our proposed ReSFU shows the remarkable ability to capture object boundaries with higher segmentation accuracy. These results further demonstrate the satisfactory application potential of ReSFU. 

% {visual analysis}

% \hippo{Firstly, we experiment with the Synapse multi-organ segmentation dataset\footnote{\url{https://www.synapse.org/\#!Synapse:syn3193805/wiki/217789}}, which covers 8 abdominal organs, \ie, aorta, gallbladder, spleen, left kidney, right kidney, liver,
% pancreas, spleen, and stomach. We use the split of 18 CT scans for training (2212 axial
% slices) and 12 CT scans for validation following \cite{chen2021transunet}. We report the average Dice of each organ, along with the average Dice and Hausdorff Distance (HD) 95 metrics of all organs over the validation set in Table \ref{tab:synapse}. It can be seen that the comparison upsampling methods can hardly surpass the default bilinear upsampler, while our ReSFU achieves overall superior performance, showing its good generality. Fig. \ref{fig:synapse} shows the segmentation results obtained with different feature upsampling methods, showing the superiority of ReSFU.
% }

% \hippo{We further adopt the Automated cardiac diagnosis challenge (ACDC) dataset\footnote{\url{https://www.creatis.insa-lyon.fr/Challenge/acdc/}}, which contains MRI images annotated with left ventricle (LV), right ventricle (RV) and myocardium (MYO). We take the split of 70 scans (1930 axial slices) for training and 40 scans for validation following \cite{chen2021transunet}. The average Dice score results are reported in Table \ref{tab:acdc}. Although different upsampling methods do not lead to much variation in the average Dice scores, it can be seen that our proposed ReSFU still outperforms all the comparison techniques.}

% \vspace{-2mm}
\subsubsection{Instance Segmentation and Panoptic Segmentation}

%To further verify the universality of our proposed ReSFU, we execute experiments on two additional tasks, including instance segmentation and panoptic segmentation.

Here we further verify the universality of our proposed ReSFU through two other segmentation tasks, including instance segmentation and panoptic segmentation. 

Specifically,  for instance segmentation, we employ the classic Mask R-CNN \cite{he2017mask} with ResNet50 as the backbone and replace the upsampling stages in FPN with ReSFU, as shown in {Fig. \ref{fig:heads} (d) (lower)}. All the results are quantitatively evaluated based on metrics of the average precision (AP) series, including mask AP, AP$_{50}$, AP$_{75}$, AP$_{S}$, AP$_{M}$, and AP$_{L}$. For panoptic segmentation, following~\cite{liu2023learning}, we choose Panoptic FPN \cite{kirillov2019panoptic} with ResNet50 as the backbone and adjust the three upsamplers in FPN, as presented in {Fig. \ref{fig:heads} (d) (lower)}. We report panoptic quality (PQ), PQ on things (PQ$^{th}$), PQ on stuff (PQ$^{st}$), segmentation quality (SQ), and recognition quality (RQ) \cite{kirillov2019panoptic}. The MS COCO dataset \cite{lin2014microsoft} is used for training and evaluating the two tasks. %The experiments are conducted based on MMDetection \cite{mmdetection}.

%where the default upsampler is nearest neighbor interpolation. % for a fair comparison

% The quantitative results are shown in \cref{tab:instance}, where the `HIN' version of IndexNet \cite{lu2019indices}, the 'dynamic-cs-d†' version of A2U \cite{dai2021learning} are used, and the `+' version of DySample is chosen because it achieves the best performance for the two tasks. It can be observed that ReSFU achieves better or comparable results than the competing methods. The qualitative results are provided in \cref{fig:vis_ins}, showing that ReSFU can exhibit better details (see the horns and ears of the giraffes in the upper image) and more consistent segmentation results.

Table~\ref{tab:instance} reports the quantitative results on the two additional segmentation tasks.
As seen, our proposed ReSFU performs more competitively over almost all the evaluation metrics, showing favorable universality.
Fig.~\ref{fig:vis_ins} and Fig. \ref{fig:vis_panoptic} provide the visual results achieved by different upsampling methods for instance segmentation and panoptic segmentation, respectively. It is easily observed that the proposed ReSFU attains better segmentation results with more details and more consistent semantics, such as the horns and ears of the giraffes in Fig.~\ref{fig:vis_ins}, and the floor in Fig. \ref{fig:vis_panoptic}.

\begin{table*}[t] 
\footnotesize
% \vspace{-0.2cm}
\caption{{Object detection results under Faster R-CNN-ResNet50 with the default nearest upsampler on MS COCO dataset, and monocular depth estimation results under DepthFormer-Swin-T with the default bilinear upsampler on NYU Depth V2. The best and second-best results are highlighted in bold and underlined, respectively.} }
% With regard to indicators like ``Abs Rel'', ``RMS'', ``log10'', ``RMS(log)'', and ``Sq Rel'', a smaller number indicates better performance, whereas the opposite holds true for the other indicators. 
\vspace{-2mm}
\label{tab:depth}
\centering
{
\renewcommand{\arraystretch}{1.2}
\scalebox{1.0}{\setlength{\tabcolsep}{0.7mm}{
\begin{tabular}{@{}l|cccccc|cccccccc@{}}
\toprule
\multicolumn{1}{c|}{\multirow{2}{*}{Method}} & \multicolumn{6}{c|}{Object Detection} & \multicolumn{8}{c}{Monocular Depth Estimation} \\ 
   % & $\delta<1.25 (\uparrow)$ & $\delta<1.25^2(\uparrow)$ &  $\delta<1.25^3(\uparrow)$ & Abs Rel$(\downarrow)$ & RMS$(\downarrow)$ & log10$(\downarrow)$ & RMS(log)$(\downarrow)$ & Sq Rel$(\downarrow)$  & $AP$ & $AP_{50}$ & $AP_{75}$ & $AP_{S}$ & $AP_{M}$ & $AP_{L}$ \\ 
     & $AP$ & $AP_{50}$ & $AP_{75}$ & $AP_{S}$ & $AP_{M}$ & $AP_{L}$ & $\delta<1.25$\tiny{$(\uparrow)$}\!\! & $\delta<1.25^2$\tiny{$(\uparrow)$}\!\! &  $\delta<1.25^3$\tiny{$(\uparrow)$}\!\! & Abs Rel\tiny{$(\downarrow)$}\!\! & RMS\tiny{$(\downarrow)$}\!\! & log10\tiny{$(\downarrow)$}\!\! & RMS(log)\tiny{$(\downarrow)$}\!\! & Sq Rel\tiny{$(\downarrow)$} \\ 
\midrule 
Default & 37.5 & 58.2 & 40.8 & 21.3 & 41.1 & 48.9 & 0.873 & \underline{0.978} & \underline{0.994} & 0.120 & 0.402 & \underline{0.050} & 0.148 & 0.071  \\
Stack-JBU \cite{fu2024featup} & 37.5 & 58.5 & 40.5 & 22.1 & 41.5 & 47.8 & 0.874 & \textbf{0.979} & \textbf{0.995} & 0.118 & 0.401 & \underline{0.050} & 0.148 & 0.071  \\
CARAFE \cite{wang2019carafe} & \underline{38.6} & \textbf{59.9} & \textbf{42.2} & \textbf{23.3} & \underline{42.2} & 49.7 & \underline{0.877} & \underline{0.978} & \textbf{0.995} & \textbf{0.116} & \underline{0.397} & \textbf{0.049} & \textbf{0.146} & \textbf{0.069}  \\
FADE \cite{lu2022fade} & 38.5 & 59.6 & 41.8 & \underline{23.1} & \underline{42.2} & 49.3 &  0.874 & \underline{0.978} & \underline{0.994} & 0.118 & 0.399 & \textbf{0.049} & \underline{0.147} & 0.071  \\
SAPA \cite{lu2022sapa} & 37.8 & 59.2 & 40.6 & 22.4 & 41.4 & 49.1 & 0.870 & \underline{0.978} & \textbf{0.995} & \underline{0.117} & 0.406 & \underline{0.050} & 0.149 & \textbf{0.069}  \\
DySample \cite{liu2023learning} & \underline{38.6} & \underline{59.8} & \underline{42.1} & 22.5 & 42.1 & \underline{50.0} & 0.872 & \underline{0.978} & \underline{0.994} & 0.119 & {0.398} & \underline{0.050} & 0.148 & \underline{0.070}  \\
ReSFU &\textbf{ 38.7} & \textbf{59.9} & \underline{42.1} & \underline{23.1} & \textbf{42.3} & \textbf{50.2}  &  \textbf{0.878} & \textbf{0.979} & \textbf{0.995} & \underline{0.117} & \textbf{0.395} & \textbf{0.049} & \textbf{0.146} & \underline{0.070} 
 \\
\bottomrule
\end{tabular}}}}
\end{table*}

\subsubsection{{Object Detection and Monocular Depth Estimation}}
{In addition to segmentation tasks, we further evaluate ReSFU's potential based on two additional application areas, including object detection and monocular depth estimation. Following the experimental settings adopted in DySample, we conduct experiments of object detection with Faster R-CNN \cite{ren2016faster} on the MS COCO dataset \cite{lin2014microsoft}, and conduct experiments of monocular depth estimation with DepthFormer \cite{li2023depthformer} on the NYU Depth V2 dataset \cite{silberman2012indoor}. Both Faster R-CNN and DepthFormer contain an FPN neck-based upsampling module as shown in Fig. \ref{fig:heads} (d).
The corresponding quantitative results are reported in Table \ref{tab:depth}. As seen, for these two challenging tasks, almost all methods have very limited room for improvement. However, the overall performance of our ReSFU still surpasses almost all comparison methods in these two tasks, which highlights the favorable universality of our methodological designs. 
}

\begin{table*}[t] 
\footnotesize
\caption{{Semantic segmentation results on ADE20K based on Segmenter and SegFormer with adopting different upsampling manners. The ``iterative'' upsampling manners for Segmenter and SegFormer are illustrated in Fig. \ref{fig:heads} (b)(upper) and Fig. \ref{fig:heads} (c)(upper), respectively, while the ``direct'' upsampler manner is illustrated in the lower row of Fig. \ref{fig:heads} (b) and Fig. \ref{fig:heads} (c), respectively. The best and second-best results are highlighted in bold and underlined, respectively.} }
\vspace{-2mm}
\label{tab:manner}
\centering
{
\renewcommand{\arraystretch}{1.14}
\scalebox{1}{\setlength{\tabcolsep}{1.8mm}{
\begin{tabular}{@{}c|l|ccc|ccc|ccc|ccc@{}}
\toprule
\multirow{2}{*}{Manner} & \multicolumn{1}{c|}{\multirow{2}{*}{Method}} & \multicolumn{3}{c|}{ Segmenter-ViT-S} & \multicolumn{3}{c|}{Segmenter-ViT-L} & \multicolumn{3}{c|}{ SegFormer-B1} & \multicolumn{3}{c}{SegFormer-B5}   \\
 & & mIoU & mAcc  & bIoU & mIoU & mAcc & bIoU & mIoU & mAcc  & bIoU & mIoU & mAcc & bIoU   \\ 
\midrule
\multirow{5}{*}{Iterative}  & CARAFE \cite{wang2019carafe} & 46.25 & 57.63 & 28.99 & 51.85 & 62.96 & 34.49 &  42.75 & 53.96 & 27.09 & 50.45  & \textbf{62.29}  & 33.78  \\
 & FADE \cite{lu2022fade}  & 45.71 & 56.81 & 28.72 & 50.07 & 61.79 & 33.32 &  43.09 & {53.89} & \underline{28.33}  &  50.43 & {61.99} & \underline{34.51} \\
 & SAPA \cite{lu2022sapa}  & 45.79 & 57.36 & 28.97 & 51.20 & 62.46 & 33.31 & 42.74 & 53.39 & {27.61}  &   49.61 &  61.11 & 34.02 \\
 & DySample \cite{liu2023learning}  & 45.78 & 56.88 & 28.37 & 52.04 & 63.27 & 34.17 & \underline{43.48} & 53.82 & 27.38 & 50.14  & 61.51 & 33.93 \\
% & Stack-JBU  & 45.70  & 56.86  & 27.84 &  51.03 & 62.39 & 32.90  \\
 & ReSFU  & \textbf{47.02}  & \textbf{57.99}  & \textbf{30.04}  & \underline{52.53}  &  \underline{63.58}  &  \underline{35.19} & \textbf{43.84} &  \textbf{55.08} &  \textbf{28.38} & \underline{50.55} & \underline{62.00} & \textbf{34.82} \\ \midrule
 \multirow{6}{*}{Direct} & Bilinear  & 45.75  & 56.88 & 27.82 & 50.96  & 62.05 &  33.07 & 40.97 & 51.73 & 24.89 & 49.13 & 60.69 & 32.32 \\
 % & Deconv  &  40.93 & 51.62 & 23.77 &  50.01 & 61.14 & 31.75  \\
 % & PixelShuffle & 41.24 & 52.37 & 23.87   & 50.50  & 61.49 & 31.45  \\
 & CARAFE \cite{wang2019carafe} &  {46.10}  & {57.39} & {28.78} & {{51.73}}  & {62.80}  &  {34.25}  &  42.65& 53.67 & 27.04 & 50.22 & 61.98 & 33.62 \\
 & FADE \cite{lu2022fade} & 45.52  & 56.81 & 28.13  & 50.93  & 62.13 & 32.60  &  42.54  & 53.38 & 28.11 & 50.12 & 61.75 & 34.10 \\
 & SAPA \cite{lu2022sapa} & 45.59 & 57.01 & 28.70  & 51.04  & 62.36  & 33.18 & 42.41 & 53.35 & 27.42 & 49.55 & 61.06 & 33.88 \\
 & DySample \cite{liu2023learning} & 45.09  &  55.98 & 27.73 & 51.55  & 63.35 & 33.66 & 43.11 & 53.88 & 26.53 & 49.94  & 61.32 & 33.62 \\
 & ReSFU  & \underline{46.97}  & \underline{57.91} &  \underline{29.96}  &  \textbf{{52.56}} & \textbf{{63.63}} &  \textbf{{35.27}} & \textbf{43.84}  & \underline{54.75} & {28.06} &\textbf{ {50.89}} & \underline{62.00} &  34.40 
 \\
\bottomrule
\end{tabular}}}}
\vspace{-1mm}
\end{table*}

\begin{figure*}[!h]
  \centering
  % \vspace{-0.2cm}
\includegraphics[width=0.99\linewidth]{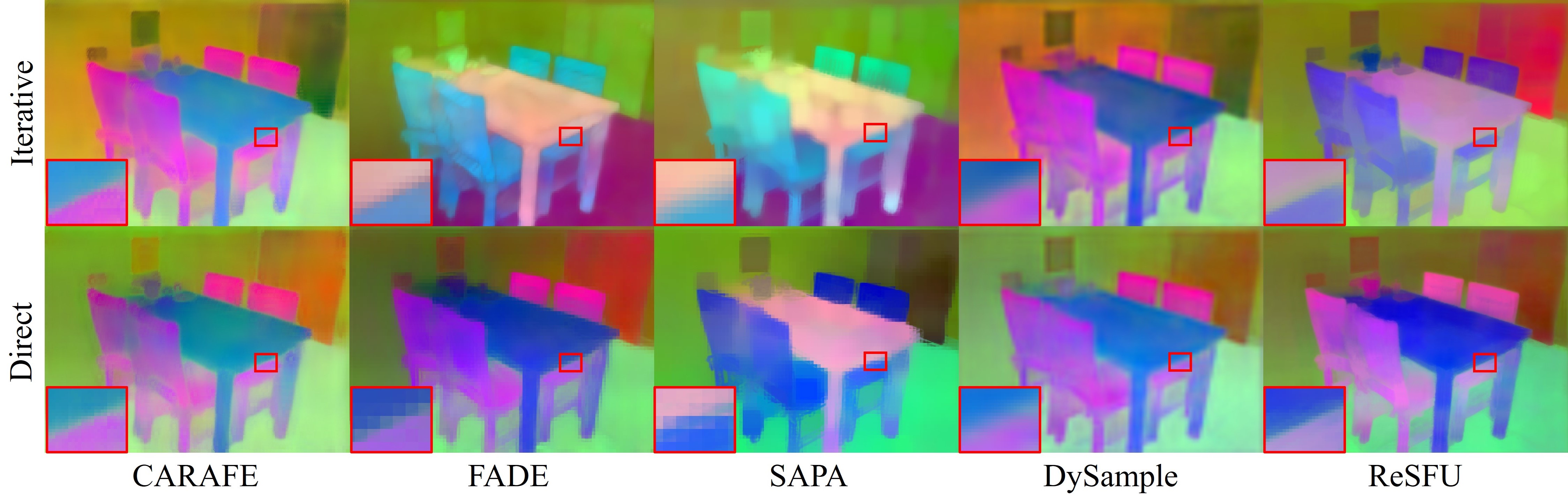}
  \vspace{-0.3cm}
  \caption{Visual comparison of upsampled features $\bm{\tilde{x}}$ obtained in an iterative upsampling ({Fig. \ref{fig:heads} (a) (upper)}) and a direct $\times 4$ upsampling manners  ({Fig. \ref{fig:heads} (a) (lower)}) based on PSPNet-ResNet101.}
  \label{fig:vis_up_iter}
  \vspace{-0.2cm}
\end{figure*}

\subsection{More Discussions on Direct/Iterative Upsampling} \label{sec:discuss}
As explained previously, other comparing upsamplers are inserted in an iterative $\times 2$ upsampling manner by default, while our proposed ReSFU is always simply executed in a direct upsampling process without iterative guidance features. Here we provide more discussions to further evaluate different methods under different upsampling manners. %we add more discussions and demonstrate the effect of other comparing approaches in a direct high-ratio upsampling form like our ReSFU.

{Based on the non-hierarchical Segmenter (Fig. \ref{fig:heads} (b)) and the hierarchical SegFormer (Fig. \ref{fig:heads} (c)), we provide quantitative segmentation results of different comparison methods with adopting different upsampling manners. As reported in Table \ref{tab:manner}, for other upsampling methods, they all achieve inferior performance in the direct upsampling manner. In contrast, ReSFU is capable of handling both of these upsampling manners with slight performance fluctuations. Besides, we use PSPNet (Fig. \ref{fig:heads} (a)) as an exemplar architecture for a visual demonstration.} For the comparing methods, \eg, CARAFE, FADE, SAPA, DySample, and ReSFU, we visually compare their upsampled features $\bm{\tilde{x}}$ obtained in the iterative $\times 2$ upsampling manner ({Fig. \ref{fig:heads} (a) (upper)}) and the direct $\times 4$ upsampling manner ({Fig. \ref{fig:heads} (a) (lower)}). As displayed in Fig.~\ref{fig:vis_up_iter}, with the high-ratio direct upsampling, CARAFE, FADE, SAPA, and DySample generate varying degrees of artifacts with more mosaics or blurrier boundaries compared with the iterative upsampling manner. {However, our proposed ReSFU achieves better visual effects with richer semantics and clearer boundaries under both upsampling manners. These advantages are  essentially attributed to our methodological designs.} 

{As validated, ReSFU achieves similar performance in the direct and iterative upsampling manners, while other comparison methods suffer from performance degradation in the direct manner both numerically and visually. Therefore, in all the experimental comparisons, for previous upsampling methods, we follow their default configuration and adopt them in an iterative manner for fair comparison. Nevertheless, for our ReSFU, we select the direct high-ratio upsampling manner as the default setting in experiments, which can eliminate the requirement for iterative guidance features and facilitate the deployment.}

\section{Conclusion and Future Work}\label{sec:con}

In this paper, for the fundamental upsampling design included in almost all current network architectures, 
we carefully reformulated the current similarity-based 
feature upsampling framework and thoroughly analyzed the limitations in methodological designs together with experimental visualizations. For each component involved in this pipeline, 
we meticulously proposed specific optimization designs and correspondingly constructed a refreshed similarity-based feature upsampling framework, called ReSFU. Through comprehensive model verification and ablation studies, we fully validated the role of each component and clearly shown the working mechanism underlying our proposed ReSFU. Based on different types of network architectures, extensive experiments substantiated the superiority of our proposed ReSFU beyond the existing baselines and demonstrated that our ReSFU can be adapted not only to different upsampling structures with flexible guidance manners but also to different applications, including medical image segmentation, instance segmentation, panoptic segmentation, {objection detection, and monocular depth estimation. However, for the applications without HR reference images, \eg, image super-resoultion~\cite{wang2020deep} and image generation~\cite{rombach2022high}, these guidance-based feature upsampling methods, \eg, FADE, SAPA, and our ReSFU, are difficult to apply. This is beyond our focus in this paper and deserves further investigation in our future work.}

\ifCLASSOPTIONcaptionsoff
  \newpage
\fi

\bibliographystyle{IEEEtran}
\normalem
%\bibliography{IEEEabrv,reference}
\bibliography{reference}

\begin{thebibliography}{10}
\providecommand{\url}[1]{#1}
\csname url@samestyle\endcsname
\providecommand{\newblock}{\relax}
\providecommand{\bibinfo}[2]{#2}
\providecommand{\BIBentrySTDinterwordspacing}{\spaceskip=0pt\relax}
\providecommand{\BIBentryALTinterwordstretchfactor}{4}
\providecommand{\BIBentryALTinterwordspacing}{\spaceskip=\fontdimen2\font plus
\BIBentryALTinterwordstretchfactor\fontdimen3\font minus \fontdimen4\font\relax}
\providecommand{\BIBforeignlanguage}[2]{{%
\expandafter\ifx\csname l@#1\endcsname\relax
\typeout{** WARNING: IEEEtran.bst: No hyphenation pattern has been}%
\typeout{** loaded for the language `#1'. Using the pattern for}%
\typeout{** the default language instead.}%
\else
\language=\csname l@#1\endcsname
\fi
#2}}
\providecommand{\BIBdecl}{\relax}
\BIBdecl

\bibitem{xie2021segformer}
E.~Xie, W.~Wang, Z.~Yu, A.~Anandkumar, J.~M. Alvarez, and P.~Luo, ``Seg{F}ormer: {S}imple and efficient design for semantic segmentation with {T}ransformers,'' in \emph{Advances in Neural Information Processing Systems}, vol.~34, 2021, pp. 12\,077--12\,090.

\bibitem{lu2022sapa}
H.~Lu, W.~Liu, Z.~Ye, H.~Fu, Y.~Liu, and Z.~Cao, ``{SAPA}: {S}imilarity-aware point affiliation for feature upsampling,'' in \emph{Advances in Neural Information Processing Systems}, vol.~35, 2022, pp. 20\,889--20\,901.

\bibitem{strudel2021segmenter}
R.~Strudel, R.~Garcia, I.~Laptev, and C.~Schmid, ``Segmenter: {T}ransformer for semantic segmentation,'' in \emph{Proceedings of the IEEE/CVF International Conference on Computer Vision}, 2021, pp. 7262--7272.

\bibitem{long2015fully}
J.~Long, E.~Shelhamer, and T.~Darrell, ``Fully convolutional networks for semantic segmentation,'' in \emph{Proceedings of the IEEE/CVF Conference on Computer Vision and Pattern Recognition}, 2015, pp. 3431--3440.

\bibitem{hafiz2020survey}
A.~M. Hafiz and G.~M. Bhat, ``A survey on instance segmentation: state of the art,'' \emph{International Journal of Multimedia Information Retrieval}, vol.~9, no.~3, pp. 171--189, 2020.

\bibitem{kirillov2019panoptic}
A.~Kirillov, K.~He, R.~Girshick, C.~Rother, and P.~Doll{\'a}r, ``Panoptic segmentation,'' in \emph{Proceedings of the IEEE/CVF Conference on Computer Vision and Pattern Recognition}, 2019, pp. 9404--9413.

\bibitem{ren2016faster}
S.~Ren, K.~He, R.~Girshick, and J.~Sun, ``Faster {R-CNN}: {T}owards real-time object detection with region proposal networks,'' \emph{IEEE Transactions on Pattern Analysis and Machine Intelligence}, vol.~39, no.~6, pp. 1137--1149, 2016.

\bibitem{li2023depthformer}
Z.~Li, Z.~Chen, X.~Liu, and J.~Jiang, ``Depthformer: {E}xploiting long-range correlation and local information for accurate monocular depth estimation,'' \emph{Machine Intelligence Research}, vol.~20, no.~6, pp. 837--854, 2023.

\bibitem{wang2019carafe}
J.~Wang, K.~Chen, R.~Xu, Z.~Liu, C.~C. Loy, and D.~Lin, ``{CARAFE}: {C}ontent-aware reassembly of features,'' in \emph{Proceedings of the IEEE/CVF International Conference on Computer Vision}, 2019, pp. 3007--3016.

\bibitem{shi2016real}
W.~Shi, J.~Caballero, F.~Husz{\'a}r, J.~Totz, A.~P. Aitken, R.~Bishop, D.~Rueckert, and Z.~Wang, ``Real-time single image and video super-resolution using an efficient sub-pixel convolutional neural network,'' in \emph{Proceedings of the IEEE/CVF Conference on Computer Vision and Pattern Recognition}, 2016, pp. 1874--1883.

\bibitem{gao2019pixel}
H.~Gao, H.~Yuan, Z.~Wang, and S.~Ji, ``Pixel transposed convolutional networks,'' \emph{IEEE Transactions on Pattern Analysis and Machine Intelligence}, vol.~42, no.~5, pp. 1218--1227, 2019.

\bibitem{wojna2019devil}
Z.~Wojna, V.~Ferrari, S.~Guadarrama, N.~Silberman, L.-C. Chen, A.~Fathi, and J.~Uijlings, ``The devil is in the decoder: {C}lassification, regression and {GAN}s,'' \emph{International Journal of Computer Vision}, vol. 127, pp. 1694--1706, 2019.

\bibitem{tian2019decoders}
Z.~Tian, T.~He, C.~Shen, and Y.~Yan, ``Decoders matter for semantic segmentation: Data-dependent decoding enables flexible feature aggregation,'' in \emph{Proceedings of the IEEE/CVF Conference on Computer Vision and Pattern Recognition}, 2019, pp. 3126--3135.

\bibitem{liu2023point}
W.~Liu, H.~Lu, Y.~Liu, and Z.~Cao, ``On point affiliation in feature upsampling,'' \emph{arXiv preprint arXiv:2307.08198}, 2023.

\bibitem{lu2022fade}
H.~Lu, W.~Liu, H.~Fu, and Z.~Cao, ``{FADE}: {F}using the assets of decoder and encoder for task-agnostic upsampling,'' in \emph{Proceedings of the European Conference on Computer Vision}.\hskip 1em plus 0.5em minus 0.4em\relax Springer, 2022, pp. 231--247.

\bibitem{liu2023learning}
W.~Liu, H.~Lu, H.~Fu, and Z.~Cao, ``Learning to upsample by learning to sample,'' in \emph{Proceedings of the IEEE/CVF International Conference on Computer Vision}, 2023, pp. 6027--6037.

\bibitem{he2012guided}
K.~He, J.~Sun, and X.~Tang, ``Guided image filtering,'' \emph{IEEE Transactions on Pattern Analysis and Machine Intelligence}, vol.~35, no.~6, pp. 1397--1409, 2012.

\bibitem{noh2015learning}
H.~Noh, S.~Hong, and B.~Han, ``Learning deconvolution network for semantic segmentation,'' in \emph{Proceedings of the IEEE/CVF International Conference on Computer Vision}, 2015, pp. 1520--1528.

\bibitem{wang2021carafe++}
J.~Wang, K.~Chen, R.~Xu, Z.~Liu, C.~C. Loy, and D.~Lin, ``{CARAFE}++: {U}nified content-aware reassembly of features,'' \emph{IEEE Transactions on Pattern Analysis and Machine Intelligence}, vol.~44, no.~9, pp. 4674--4687, 2021.

\bibitem{badrinarayanan2017segnet}
V.~Badrinarayanan, A.~Kendall, and R.~Cipolla, ``Seg{N}et: {A} deep convolutional encoder-decoder architecture for image segmentation,'' \emph{IEEE Transactions on Pattern Analysis and Machine Intelligence}, vol.~39, no.~12, pp. 2481--2495, 2017.

\bibitem{mazzini2018guided}
D.~Mazzini, ``Guided upsampling network for real-time semantic segmentation,'' in \emph{Proceedings of the British Machine Vision Conference}, vol. 117, 2018, pp. 1--12.

\bibitem{lu2019indices}
H.~Lu, Y.~Dai, C.~Shen, and S.~Xu, ``Indices matter: {L}earning to index for deep image matting,'' in \emph{Proceedings of the IEEE/CVF International Conference on Computer Vision}, 2019, pp. 3266--3275.

\bibitem{dai2021learning}
Y.~Dai, H.~Lu, and C.~Shen, ``Learning affinity-aware upsampling for deep image matting,'' in \emph{Proceedings of the IEEE/CVF Conference on Computer Vision and Pattern Recognition}, 2021, pp. 6841--6850.

\bibitem{fu2024featup}
S.~Fu, M.~Hamilton, L.~E. Brandt, A.~Feldmann, Z.~Zhang, and W.~T. Freeman, ``Feat{U}p: {A} model-agnostic framework for features at any resolution,'' in \emph{International Conference on Learning Representations}, 2024.

\bibitem{kopf2007joint}
J.~Kopf, M.~F. Cohen, D.~Lischinski, and M.~Uyttendaele, ``Joint bilateral upsampling,'' \emph{ACM Transactions on Graphics}, vol.~26, no.~3, pp. 96--100, 2007.

\bibitem{zeiler2014visualizing}
M.~D. Zeiler and R.~Fergus, ``Visualizing and understanding convolutional networks,'' in \emph{Proceedings of the European Conference on Computer Vision}.\hskip 1em plus 0.5em minus 0.4em\relax Springer, 2014, pp. 818--833.

\bibitem{dosovitskiy2021an}
A.~Dosovitskiy, L.~Beyer, A.~Kolesnikov, D.~Weissenborn, X.~Zhai, T.~Unterthiner, M.~Dehghani, M.~Minderer, G.~Heigold, S.~Gelly, J.~Uszkoreit, and N.~Houlsby, ``An image is worth 16x16 words: {T}ransformers for image recognition at scale,'' in \emph{International Conference on Learning Representations}, 2021.

\bibitem{he2015fast}
K.~He and J.~Sun, ``Fast guided filter,'' \emph{arXiv preprint arXiv:1505.00996}, 2015.

\bibitem{draper1998applied}
N.~R. Draper and H.~Smith, \emph{Applied Regression Analysis}.\hskip 1em plus 0.5em minus 0.4em\relax John Wiley \& Sons, 1998, vol. 326.

\bibitem{vaswani2017attention}
A.~Vaswani, N.~Shazeer, N.~Parmar, J.~Uszkoreit, L.~Jones, A.~N. Gomez, {\L}.~Kaiser, and I.~Polosukhin, ``Attention is all you need,'' \emph{Advances in Neural Information Processing Systems}, vol.~30, pp. 5998--6008, 2017.

\bibitem{xu2023self}
Q.~Xu, W.~Zhao, G.~Lin, and C.~Long, ``Self-calibrated cross attention network for few-shot segmentation,'' in \emph{Proceedings of the IEEE/CVF International Conference on Computer Vision}, 2023, pp. 655--665.

\bibitem{petschnigg2004digital}
G.~Petschnigg, R.~Szeliski, M.~Agrawala, M.~Cohen, H.~Hoppe, and K.~Toyama, ``Digital photography with flash and no-flash image pairs,'' \emph{ACM Transactions on Graphics}, vol.~23, no.~3, pp. 664--672, 2004.

\bibitem{yu2020searching}
Z.~Yu, C.~Zhao, Z.~Wang, Y.~Qin, Z.~Su, X.~Li, F.~Zhou, and G.~Zhao, ``Searching central difference convolutional networks for face anti-spoofing,'' in \emph{Proceedings of the IEEE/CVF Conference on Computer Vision and Pattern Recognition}, 2020, pp. 5295--5305.

\bibitem{paszke2017automatic}
A.~Paszke, S.~Gross, S.~Chintala, G.~Chanan, E.~Yang, Z.~DeVito, Z.~Lin, A.~Desmaison, L.~Antiga, and A.~Lerer, ``Automatic differentiation in {PyT}orch,'' in \emph{NIPS 2017 Workshop on Autodiff}, 2017.

\bibitem{he2016deep}
K.~He, X.~Zhang, S.~Ren, and J.~Sun, ``Deep residual learning for image recognition,'' in \emph{Proceedings of the IEEE/CVF Conference on Computer Vision and Pattern Recognition}, 2016, pp. 770--778.

\bibitem{liu2021swin}
Z.~Liu, Y.~Lin, Y.~Cao, H.~Hu, Y.~Wei, Z.~Zhang, S.~Lin, and B.~Guo, ``Swin {T}ransformer: {H}ierarchical vision transformer using shifted windows,'' in \emph{Proceedings of the IEEE/CVF International Conference on Computer Vision}, 2021, pp. 10\,012--10\,022.

\bibitem{zhao2017pyramid}
H.~Zhao, J.~Shi, X.~Qi, X.~Wang, and J.~Jia, ``Pyramid scene parsing network,'' in \emph{Proceedings of the IEEE/CVF Conference on Computer Vision and Pattern Recognition}, 2017, pp. 2881--2890.

\bibitem{cheng2021per}
B.~Cheng, A.~Schwing, and A.~Kirillov, ``Per-pixel classification is not all you need for semantic segmentation,'' \emph{Advances in Neural Information Processing Systems}, vol.~34, pp. 17\,864--17\,875, 2021.

\bibitem{zhou2017scene}
B.~Zhou, H.~Zhao, X.~Puig, S.~Fidler, A.~Barriuso, and A.~Torralba, ``Scene parsing through {ADE}20k dataset,'' in \emph{Proceedings of the IEEE/CVF Conference on Computer Vision and Pattern Recognition}, 2017, pp. 633--641.

\bibitem{cheng2021boundary}
B.~Cheng, R.~Girshick, P.~Doll{\'a}r, A.~C. Berg, and A.~Kirillov, ``Boundary {IoU}: {I}mproving object-centric image segmentation evaluation,'' in \emph{Proceedings of the IEEE/CVF Conference on Computer Vision and Pattern Recognition}, 2021, pp. 15\,334--15\,342.

\bibitem{abdi2010principal}
H.~Abdi and L.~J. Williams, ``Principal component analysis,'' \emph{Wiley Interdisciplinary Reviews: Computational Statistics}, vol.~2, no.~4, pp. 433--459, 2010.

\bibitem{lin2017feature}
T.-Y. Lin, P.~Doll{\'a}r, R.~Girshick, K.~He, B.~Hariharan, and S.~Belongie, ``Feature pyramid networks for object detection,'' in \emph{Proceedings of the IEEE/CVF Conference on Computer Vision and Pattern Recognition}, 2017, pp. 2117--2125.

\bibitem{sugawara2019checkerboard}
Y.~Sugawara, S.~Shiota, and H.~Kiya, ``Checkerboard artifacts free convolutional neural networks,'' \emph{APSIPA Transactions on Signal and Information Processing}, vol.~8, p.~e9, 2019.

\bibitem{chen2021transunet}
J.~Chen, Y.~Lu, Q.~Yu, X.~Luo, E.~Adeli, Y.~Wang, L.~Lu, A.~L. Yuille, and Y.~Zhou, ``Trans{UN}et: {T}ransformers make strong encoders for medical image segmentation,'' \emph{arXiv preprint arXiv:2102.04306}, 2021.

\bibitem{he2017mask}
K.~He, G.~Gkioxari, P.~Doll{\'a}r, and R.~Girshick, ``Mask {R-CNN},'' in \emph{Proceedings of the IEEE/CVF International Conference on Computer Vision}, 2017, pp. 2961--2969.

\bibitem{lin2014microsoft}
T.-Y. Lin, M.~Maire, S.~Belongie, J.~Hays, P.~Perona, D.~Ramanan, P.~Doll{\'a}r, and C.~L. Zitnick, ``Microsoft {COCO}: {C}ommon objects in context,'' in \emph{Proceedings of the European Conference on Computer Vision}.\hskip 1em plus 0.5em minus 0.4em\relax Springer, 2014, pp. 740--755.

\bibitem{silberman2012indoor}
N.~Silberman, D.~Hoiem, P.~Kohli, and R.~Fergus, ``Indoor segmentation and support inference from {RGBD} images,'' in \emph{Proceedings of the European Conference on Computer Vision}.\hskip 1em plus 0.5em minus 0.4em\relax Springer, 2012, pp. 746--760.

\bibitem{wang2020deep}
Z.~Wang, J.~Chen, and S.~C. Hoi, ``Deep learning for image super-resolution: A survey,'' \emph{IEEE Transactions on Pattern Analysis and Machine Intelligence}, vol.~43, no.~10, pp. 3365--3387, 2020.

\bibitem{rombach2022high}
R.~Rombach, A.~Blattmann, D.~Lorenz, P.~Esser, and B.~Ommer, ``High-resolution image synthesis with latent diffusion models,'' in \emph{Proceedings of the IEEE/CVF Conference on Computer Vision and Pattern Recognition}, 2022, pp. 10\,684--10\,695.

\end{thebibliography}

\end{document}